\documentclass{article}

\PassOptionsToPackage{numbers}{natbib}
\usepackage[final]{neurips_2021}

\usepackage[utf8]{inputenc} 
\usepackage[T1]{fontenc}    
\usepackage{hyperref}       
\usepackage{url}            
\usepackage{booktabs}       
\usepackage{amsfonts}       
\usepackage{nicefrac}       
\usepackage{microtype}      
\usepackage{xcolor}         
\usepackage{float}
\usepackage{amsmath}
\usepackage{amssymb}
\usepackage{graphicx}
\usepackage{subcaption}
\usepackage{xcolor}
\usepackage{url}
\usepackage{bm}
\usepackage{multirow}
\usepackage{listings}
\usepackage{tikz}
\usetikzlibrary{shapes,arrows}
\usetikzlibrary{decorations.pathreplacing, decorations.pathmorphing, calligraphy}

\tikzstyle{rectblock} = [rectangle, draw, thick, align=center]
\tikzstyle{block} = [rectblock, rounded corners]
\tikzstyle{boundingbox} = [very thick, gray]
\tikzstyle{dashblock} = [rectangle, draw, thick, align=center, dashed]
\tikzstyle{conc} = [ellipse, draw, thick, dashed, align=center]
\tikzstyle{netnode} = [circle, draw, very thick, inner sep=0pt, minimum size=0.5cm]
\tikzstyle{opnode} = [circle, draw, thick, inner sep=0pt, minimum size=0.5cm, align=center]
\tikzstyle{relunode} = [rectangle, draw, very thick, inner sep=0pt, minimum size=0.5cm]
\tikzstyle{line} = [draw, very thick, -latex']
\tikzstyle{arrow} = [draw, ->, thick]
\tikzstyle{attention} = [arrow, bend right]
\tikzstyle{mapsto} = [draw, |->, thick]
\tikzstyle{annobrace} = [draw, thick, decorate, decoration={brace, mirror, amplitude=0.5em}]
\tikzstyle{annotext} = [pos=0.5, text width=1.5cm, anchor=north, yshift=-0.25em, align=center]

\definecolor{bpurp}{HTML}{984ea3}
\definecolor{bblue}{HTML}{377eb8}
\definecolor{bgreen}{HTML}{4daf4a}
\definecolor{borange}{HTML}{ff7f00}
\definecolor{bred}{HTML}{a50f15}

\newfloat{lstfloat}{htbp}{lop}
\floatname{lstfloat}{Listing}

\newif\ifanonymized
\anonymizedfalse

\makeatletter
\if@submission
  \anonymizedtrue
\fi
\makeatother

\title{Towards mental time travel: a hierarchical memory for reinforcement learning agents}
\date{}

\author{%
  Andrew K. Lampinen\\
  DeepMind\\
  London, UK\\
  \texttt{lampinen@deepmind.com} \\
  \And
  Stephanie C. Y. Chan\\
  DeepMind\\
  London, UK\\
  \texttt{scychan@deepmind.com} \\
  \And
  Andrea Banino\\
  DeepMind\\
  London, UK\\
  \texttt{abanino@deepmind.com} \\
  \And
  Felix Hill\\
  DeepMind\\
  London, UK\\
  \texttt{felixhill@deepmind.com} \\
}

\begin{document}

\maketitle

\everypar{\looseness=-1}
\linepenalty=500
\begin{abstract}
Reinforcement learning agents often forget details of the past, especially after delays or distractor tasks. Agents with common memory architectures struggle to recall and integrate across multiple timesteps of a past event, or even to recall the details of a single timestep that is followed by distractor tasks. To address these limitations, we propose a Hierarchical Chunk Attention Memory (HCAM), which helps agents to remember the past in detail. HCAM stores memories by dividing the past into chunks, and recalls by first performing high-level attention over coarse summaries of the chunks, and then performing detailed attention within only the most relevant chunks. An agent with HCAM can therefore ``mentally time-travel''---remember past events in detail without attending to all intervening events. We show that agents with HCAM substantially outperform agents with other memory architectures at tasks requiring long-term recall, retention, or reasoning over memory. These include recalling where an object is hidden in a 3D environment, rapidly learning to navigate efficiently in a new neighborhood, and rapidly learning and retaining new object names. Agents with HCAM can extrapolate to task sequences an order of magnitude longer than they were trained on, and can even generalize zero-shot from a meta-learning setting to maintaining knowledge across episodes. HCAM improves agent sample efficiency, generalization, and generality (by solving tasks that previously required specialized architectures). Our work is a step towards agents that can learn, interact, and adapt in complex and temporally-extended environments.
\end{abstract}

\section{Introduction}

Human learning and generalization relies on our detailed memory of the past \citep{tulving1985memory,suddendorf2009mental, hasson2015hierarchical,kumaran2016learning,NGO2021}.  If we watch a show, we can generally recall its scenes in some detail afterward. If we explore a new neighborhood, we can recall our paths through it in order to plan new routes. If we are exposed to a new noun, we can find that object by name later. We experience relatively little interference from delays or intervening events. Indeed, human episodic memory has been compared to ``mental time travel'' \citep{tulving1985memory,suddendorf2009mental,manning2021episodic}---we are able to \emph{transport} ourselves into a past event and re-live it in sequential detail, without attending to everything that has happened since. That is, our recall is both sparse (we attend to a small chunk of the past, or a few chunks) \emph{and} detailed (we recover a large amount of information from each chunk). This combination gives humans the flexibility necessary to recover information from memory that is specifically relevant to the task at hand, even after long periods of time.

If we want Reinforcement Learning (RL) agents to meaningfully interact with complex environments over time, our agents will need to achieve this type of memory. They will need to remember event details. They will need to retain information they have acquired, despite unrelated intervening tasks. They will need to rapidly learn many pieces of new knowledge without discarding previous ones. They will need to reason over their memories to generalize beyond their training experience. 

However, current models struggle to achieve rapid encoding combined with detailed recall, even over relatively short timescales. Meta-learning approaches \citep{wang2021meta} can slowly learn global task knowledge in order to rapidly learn new information, but they generally discard the details of that new information immediately to solve the next task. While LSTMs \citep{hochreiter1997long}, Transformers \citep{vaswani2017attention}, and variants like the TransformerXL \citep{dai2019transformer} can serve as RL agent memories in short tasks \citep{parisotto2020stabilizing}, we show that they are ineffective at detailed recall even after a few minutes. Transformer attention can be ineffective at long ranges even in supervised tasks \citep{tay2021long}, and this problem is likely exacerbated by the sparse learning signals in RL. By contrast, prior episodic memory architectures \citep{hung2019optimizing,fortunato2019generalization} can maintain information over slightly longer timescales, but they struggle with tasks that require recalling a temporally-structured event, rather than simply recalling a single past state. That is, they are particularly poor at the type of event recall that is fundamental to human memory. We suggest that this is because prior methods lack 1) memory chunks longer than a single timestep and 2) sparsity of attention. This means that these models cannot effectively ``time-travel'' to an event, and relive that specific event in detail, without interference from all other events. This also limits their ability to reason over their memories to adapt to new tasks---for example, planning a new path by combining pieces of previous ones \citep{ritter2020rapid}.

Here, we propose a new type of memory that begins to address these challenges, by leveraging sparsity, chunking, and hierarchical attention. We refer to our architecture as a Hierarchical Chunk Attention Memory (HCAM). The fundamental insight of HCAM is to divide the past into distinct chunks before storing it in memory, and to recall hierarchically. To recall, HCAM first uses coarse chunk summaries to identify relevant chunks, and then mentally travels back to attend to each relevant chunk in further detail. This approach combines benefits of the sparse and relatively long-term recall ability of prior episodic memories \citep[]{wayne2018unsupervised,ritter2020rapid} with the short-term sequential and reasoning power of transformers \citep{vaswani2017attention,dai2019transformer,parisotto2020stabilizing} in order to achieve better recall and reasoning over memory than either prior approach. While we do not address the problem of optimally chunking the past here, instead employing arbitrary chunks of fixed length, our results show that even fixed chunking can substantially improve memory. (Relatedly, \citet{ke2018sparse} have shown that sparse, top-\(k\) retrieval of memory chunks can improve credit assignment in LSTMs with attention.)

We show that HCAM allows RL agents to recall events over longer intervals and with greater sample efficiency than prior memories. Agents with HCAM can learn new words and then maintain them over distractor phases. They can extrapolate far beyond the training distribution, to maintain and recall memories after 5\(\times\) more distractors than during training, or after multiple episodes when trained only on single ones. Agents with HCAM can reason over their memories to plan paths near-optimally as they explore a new neighborhood, comparable to an agent architecture designed specifically for that task. HCAM is robust to hyperparameters, and agents with HCAM consistently outperform agents with Gated TransformerXL memories \citep{parisotto2020stabilizing} that are twice as wide or twice as deep. Furthermore, HCAM is more robust to varying hyperparameters than other architectures (App.\ \ref{app:supp_exp:response}). Thus HCAM provides a substantial improvement in the memory capabilities of RL agents.

\subsection{Background}

\paragraph{Memory in RL} Learning signals are sparse in RL, which can make it challenging to learn tasks requiring long memory. Thus, a variety of sophisticated memory architectures have been proposed for RL agents. Most memories are based on LSTMs \citep{hochreiter1997long}, often with additional key-value episodic memories \citep{wayne2018unsupervised,fortunato2019generalization}. Often, RL agents require self-supervised auxiliary training because sparse task rewards do not provide enough signal for the agent to learn what to store in memory \citep{wayne2018unsupervised,fortunato2019generalization,hill2020grounded}.

\paragraph{Transformers} Transformers \citep{vaswani2017attention} are sequence models that use attention rather than recurrence. These models---and their variants \citep{dai2019transformer,kitaev2020reformer,wang2020linformer,press2021train}---have come to dominate natural language processing. This success of transformers over LSTMs on sequential language tasks suggests that they might be effective agent memories, and a recent work succesfully used transformers as RL agent memories \citep{parisotto2020stabilizing}---specifically, a gated version of TransformerXL \citep{dai2019transformer}. However, using TransformerXL memories on challenging memory tasks still requires auxiliary self-supervision \citep{hill2020grounded}, and might even benefit from new unsupervised learning mechanisms \citep{banino2021coberl}. Furthermore, even in supervised tasks Transformers can struggle to recall details from long sequences \citep{tay2021long}. In the next section, we propose a new hierarchical attention memory architecture for RL agents that can help overcome these challenges.

\section{Hierarchical Chunk Attention Memory}
\begin{figure}[tbh]
    \vspace{-2.4em}
    \centering
    \begin{subfigure}[b]{0.5\textwidth}
    \centering
    \resizebox{0.9\textwidth}{!}{
    \begin{tikzpicture}
    \node[rectblock, fill=bpurp, minimum width=1cm, minimum height=0.5cm] at (-1, 0) (c00) {}; 
    \node[rectblock, fill=bpurp!20!white, minimum width=1cm, minimum height=0.5cm] at (0, 0) (c01) {}; 
    \node[rectblock, fill=bpurp!60!white, minimum width=1cm, minimum height=0.5cm] at (1, 0) (c02) {}; 
    \node[rectblock, fill=bpurp!30!white, minimum width=1cm, minimum height=0.5cm] at (2, 0) (c03) {}; 
    \node[minimum width=1cm, minimum height=0.5cm] at (3, 0) (space) {\LARGE \(\dots\)}; 
    \node[rectblock, fill=bblue!50!white, minimum width=1cm, minimum height=0.5cm] at (4, 0) (c10) {}; 
    \node[rectblock, fill=bblue!20!white, minimum width=1cm, minimum height=0.5cm] at (5, 0) (c11) {}; 
    \node[rectblock, fill=bblue, minimum width=1cm, minimum height=0.5cm] at (6, 0) (c12) {}; 
    \node[rectblock, fill=bblue!30!white, minimum width=1cm, minimum height=0.5cm] at (7, 0) (c13) {}; 
    \node[rectblock, fill=bgreen!20!white, minimum width=1cm, minimum height=0.5cm] at (8, 0) (c20) {}; 
    \node[rectblock, fill=bgreen, minimum width=1cm, minimum height=0.5cm] at (9, 0) (c21) {}; 
    \node[rectblock, fill=bgreen!60!white, minimum width=1cm, minimum height=0.5cm] at (10, 0) (c22) {}; 
    \node[rectblock, fill=bgreen!30!white, minimum width=1cm, minimum height=0.5cm] at (11, 0) (c23) {}; 
    \node[rectblock, fill=borange, minimum width=1cm, minimum height=0.5cm] at (12, 0) (c30) {}; 
    
    \path[attention] (c30.north) to (c00.north);
    \path[attention] (c30.north) to (c01.north);
    \path[attention] (c30.north) to (c02.north);
    \path[attention] (c30.north) to (c03.north);
    \path[attention] (c30.north) to (c10.north);
    \path[attention] (c30.north) to (c11.north);
    \path[attention] (c30.north) to (c12.north);
    \path[attention] (c30.north) to (c13.north);
    \path[attention] (c30.north) to (c20.north);
    \path[attention] (c30.north) to (c21.north);
    \path[attention] (c30.north) to (c22.north);
    \path[attention] (c30.north) to (c23.north);
    
    \end{tikzpicture}
    }
    \caption{Standard attention.} \label{fig:HM_attention:standard}
    \end{subfigure}%
    \begin{subfigure}[b]{0.5\textwidth}
    \centering
    \resizebox{0.9\textwidth}{!}{
    \begin{tikzpicture}
    \node[rectblock, fill=bpurp, minimum width=1cm, minimum height=0.5cm] at (-1, 0) (c00) {}; 
    \node[rectblock, fill=bpurp!20!white, minimum width=1cm, minimum height=0.5cm] at (0, 0) (c01) {}; 
    \node[rectblock, fill=bpurp!60!white, minimum width=1cm, minimum height=0.5cm] at (1, 0) (c02) {}; 
    \node[rectblock, fill=bpurp!30!white, minimum width=1cm, minimum height=0.5cm] at (2, 0) (c03) {}; 
    \node[minimum width=1cm, minimum height=0.5cm] at (3, 0) (space) {\LARGE \(\dots\)}; 
    \node[rectblock, fill=bblue!50!white, minimum width=1cm, minimum height=0.5cm] at (4, 0) (c10) {}; 
    \node[rectblock, fill=bblue!20!white, minimum width=1cm, minimum height=0.5cm] at (5, 0) (c11) {}; 
    \node[rectblock, fill=bblue, minimum width=1cm, minimum height=0.5cm] at (6, 0) (c12) {}; 
    \node[rectblock, fill=bblue!30!white, minimum width=1cm, minimum height=0.5cm] at (7, 0) (c13) {}; 
    \node[rectblock, fill=bgreen!20!white, minimum width=1cm, minimum height=0.5cm] at (8, 0) (c20) {}; 
    \node[rectblock, fill=bgreen, minimum width=1cm, minimum height=0.5cm] at (9, 0) (c21) {}; 
    \node[rectblock, fill=bgreen!60!white, minimum width=1cm, minimum height=0.5cm] at (10, 0) (c22) {}; 
    \node[rectblock, fill=bgreen!30!white, minimum width=1cm, minimum height=0.5cm] at (11, 0) (c23) {}; 
    \node[rectblock, fill=borange, minimum width=1cm, minimum height=0.5cm] at (12, 0) (c30) {}; 
    
    \node[rectblock, fill=bpurp!70!white, minimum width=1cm, minimum height=0.5cm] at (0.5, 1.5) (sum0) {}; 
    \node[rectblock, fill=bblue!50!white, minimum width=1cm, minimum height=0.5cm] at (5.5, 1.5) (sum1) {}; 
    \node[rectblock, fill=bgreen!70!white, minimum width=1cm, minimum height=0.5cm] at (9.5, 1.5) (sum2) {};

    \path[attention] (c30.north) to [out=-50, in=-155] (sum0.north);
    \path[attention] (c30.north) to [out=-48, in=-145] (sum1.north);
    \path[attention] (c30.north) to [out=-25, in=-140] (sum2.north);
    
    \path[arrow] (sum0.south) to (c00.north);
    \path[arrow] (sum0.south) to (c01.north);
    \path[arrow] (sum0.south) to (c02.north);
    \path[arrow] (sum0.south) to (c03.north);
    
    \node[align=center] at (5.5, 0.75) {\Large No attention};
    
    \path[arrow] (sum2.south) to (c20.north);
    \path[arrow] (sum2.south) to (c21.north);
    \path[arrow] (sum2.south) to (c22.north);
    \path[arrow] (sum2.south) to (c23.north);
    
    \end{tikzpicture}
    }
    \caption{Hierarchical attention.} \label{fig:HM_attention:hierarchical}
    \end{subfigure}\\
    \begin{subfigure}[b]{0.66\textwidth}
    \centering
    \resizebox{\textwidth}{!}{
    \begin{tikzpicture}
    
    \draw[boundingbox, draw=gray] (-2.5, 1.25) rectangle (4, 5);
    \node[text=gray] at (-1.6, 4.7) {\bf Memory};
    
    \node[rectblock, fill=borange, minimum width=1cm, minimum height=0.5cm] at (-3.5, 0) (input) {}; 
    \node at (-3.5, -0.6) {\bf Input};
    
    \node[block, minimum width=1cm, minimum height=0.5cm] at (-3.5, 0.8) (inputnorm) {LayerNorm}; 
    \path[arrow] (input.north) to (inputnorm.south);
    
    \node[opnode] at (-3.5, 3.25) (dot) {\LARGE \(\bm \cdot\)};
    \path[arrow] (inputnorm.north) to (dot.south);
    \node[text width=1cm] at (-4.5, 3.25) {Attn. across chunks};
    
    \node[text=gray] at (-1.5, 1.6) {\bf Keys};
    \node[rectblock, fill=bpurp!70!white, minimum width=1cm, minimum height=0.5cm] at (-1.5, 4) (sum0) {}; 
    \node[minimum width=1cm, minimum height=0.5cm] at (-1.5, 3.6) (spacekeys) {\large \(\bm \vdots\)}; 
    \node[rectblock, fill=bblue!50!white, minimum width=1cm, minimum height=0.5cm] at (-1.5, 3) (sum1) {}; 
    \node[rectblock, fill=bgreen!70!white, minimum width=1cm, minimum height=0.5cm] at (-1.5, 2.5) (sum2) {}; 
    
    \path[arrow] (dot) to (sum0.west);
    \path[arrow] (dot) to (sum1.west);
    \path[arrow] (dot) to (sum2.west);
    
    \node[text=gray] at (1.5, 1.6) {\bf Chunks};
    \node[rectblock, fill=bpurp, minimum width=1cm, minimum height=0.5cm] at (0, 4) (c00) {}; 
    \node[rectblock, fill=bpurp!20!white, minimum width=1cm, minimum height=0.5cm] at (1, 4) (c01) {}; 
    \node[rectblock, fill=bpurp!60!white, minimum width=1cm, minimum height=0.5cm] at (2, 4) (c02) {}; 
    \node[rectblock, fill=bpurp!30!white, minimum width=1cm, minimum height=0.5cm] at (3, 4) (c03) {}; 
    \node[minimum width=1cm, minimum height=0.5cm] at (1.5, 3.6) (space) {\large \(\bm \vdots\)}; 
    \node[rectblock, fill=bblue!50!white, minimum width=1cm, minimum height=0.5cm] at (0, 3) (c10) {}; 
    \node[rectblock, fill=bblue!20!white, minimum width=1cm, minimum height=0.5cm] at (1, 3) (c11) {}; 
    \node[rectblock, fill=bblue, minimum width=1cm, minimum height=0.5cm] at (2, 3) (c12) {}; 
    \node[rectblock, fill=bblue!30!white, minimum width=1cm, minimum height=0.5cm] at (3, 3) (c13) {}; 
    \node[rectblock, fill=bgreen!20!white, minimum width=1cm, minimum height=0.5cm] at (0, 2.5) (c20) {}; 
    \node[rectblock, fill=bgreen, minimum width=1cm, minimum height=0.5cm] at (1, 2.5) (c21) {}; 
    \node[rectblock, fill=bgreen!60!white, minimum width=1cm, minimum height=0.5cm] at (2, 2.5) (c22) {}; 
    \node[rectblock, fill=bgreen!30!white, minimum width=1cm, minimum height=0.5cm] at (3, 2.5) (c23) {};

    \node[rectblock, fill=borange!30!white, minimum width=1cm, minimum height=0.5cm] at (5, 4) (inputc0) {}; 
    \path[attention] (inputc0.north) to [out=-15, in=-165] (c00.north);
    \path[attention] (inputc0.north) to [out=-15, in=-165] (c01.north);
    \path[attention] (inputc0.north) to [out=-15, in=-165] (c02.north);
    \path[attention] (inputc0.north) to [out=-15, in=-165] (c03.north);
    
    \node[rectblock, fill=borange!70!white, minimum width=1cm, minimum height=0.5cm] at (5, 2.5) (inputc2) {}; 
    \path[attention] (inputc2.south) to [out=15, in=165] (c20.south);
    \path[attention] (inputc2.south) to [out=15, in=165] (c21.south);
    \path[attention] (inputc2.south) to [out=15, in=165] (c22.south);
    \path[attention] (inputc2.south) to [out=15, in=165] (c23.south);
    
    \node[text width=1.9cm] at (5.1, 4.85) {Attn.\! within top-\(k\)\! chunks};
    
    \node[opnode] at (6.5, 3.25) (combine) {\large \(\bm +\)};
    \node[text width=1.5cm] at (7.75, 3.25) {Weighted sum of results};
    \path[arrow] (inputc0.east) to (combine);
    \path[arrow] (inputc2.east) to (combine);

    \path[arrow, rounded corners] (dot) -- (-3.5, 5.5) -- (6.5, 5.5) -- (combine);
    
    \node[rectblock, fill=borange!80!white, minimum width=1cm, minimum height=0.5cm] at (6.5, 0) (output) {}; 
    \node at (6.5, -0.6) {\bf Output};
    
    \path[arrow] (combine) to (output);
    
    \path[arrow, dashed] ([yshift=-3]input.east) to ([yshift=-3]output.west);
    \path[arrow] (inputnorm.east) -- ([xshift=3]5, 0.8) -- ([xshift=3]inputc2.south);
    \node at (1.5, -0.45) {Skip connection};
    
    \end{tikzpicture}
    }
    \vskip-0.33em
    \caption{HCAM attention block.} \label{fig:HM_attention:HCAM_block}
    \end{subfigure}%
    \hfill
    \begin{subfigure}[b]{0.3\textwidth}
    \centering
    \resizebox{\textwidth}{!}{
    \begin{tikzpicture}
    \draw[boundingbox, dashed, draw=borange] (-2.9, 0.4) rectangle (4, 1.1);
    \node[text=borange] at (3.25, 0.75) {Encoders};
    \node at (-1, 0) {Image};
    \node at (1, 0) {Language};
    
    \node[block, minimum width=1.5cm] at (-1, 0.75) (venc) {ResNet};
    \node[block, minimum width=1.5cm] at (1, 0.75) (lenc) {LSTM};
    \draw[boundingbox, dashed, draw=bpurp] (-2.9, 1.3) rectangle (4, 5.7);
    \node[text=bpurp, text width=1.75cm] at (3.4, 5.25) {Memory\phantom{ah} (4 layers)};
    \draw[boundingbox, dashed, draw=bgreen] (-2.8, 1.4) rectangle (1.2, 3.1);
    \node[text=bgreen] at (-1.9, 2.83) {Mem. layer};
    \node[block, minimum width=1.5cm] at (0, 1.75) (att1) {Attention};
    \path[arrow, very thick] (lenc.north) to ([xshift=1em]att1.south);
    \path[arrow, very thick] (venc.north) to ([xshift=-1em]att1.south);
    \node[block, minimum width=1.5cm] at (0, 2.25) (htm1) {HCAM Attn.};
    \node[block] at (-2, 2.25) (htm1mem) {Memory};
    \path[arrow, very thick, <->] (htm1.west) to (htm1mem.east);
    \node[block, minimum width=1.5cm] at (0, 2.75) (ff1) {MLP};
    
    \node[scale=2] at (0, 3.75) {\large \(\vdots\)};
    
    \draw[boundingbox, dashed, draw=bgreen] (-2.8, 3.9) rectangle (1.2, 5.6);
    \node[block, minimum width=1.5cm] at (0, 4.25) (att2) {Attention};
    \node[block, minimum width=1.5cm] at (0, 4.75) (htm2) {HCAM Attn.};
    \node[block] at (-2, 4.75) (htm2mem) {Memory};
    \path[arrow, very thick, <->] (htm2.west) to (htm2mem.east);
    \node[block, minimum width=1.5cm] at (0, 5.25) (ff2) {MLP};
    \draw[boundingbox, dashed, draw=borange] (-2.9, 5.9) rectangle (4, 6.6);
    \node[text=borange] at (3.25, 6.25) {Decoders};
    
    \node[block, minimum width=0.66cm] at (-2, 6.25) (irec) {ResN};
    \node[block, minimum width=0.66cm] at (-0.66, 6.25) (lrec) {LSTM};
    \node[block, minimum width=0.66cm] at (0.66, 6.25) (pol) {MLP};
    \node[block, minimum width=0.66cm] at (2, 6.25) (val) {MLP};
    \path[arrow, very thick] ([xshift=-1em]ff2.north) to (irec.south);
    \path[arrow, very thick] ([xshift=-1em]ff2.north) to (lrec.south);
    \path[arrow, very thick] ([xshift=1em]ff2.north) to (pol.south);
    \path[arrow, very thick] ([xshift=1em]ff2.north) to (val.south);
    \node at (-2, 6.9) (irec) {Image};
    \node at (-0.66, 6.9) (lrec) {Lang.};
    \draw[boundingbox, dashed, draw=bred] (-2.6, 6.7) rectangle (-0.1, 7.5);
    \node[text=bred] at (-1.35, 7.25) {Self-supervised};
    \draw[boundingbox, dashed, draw=bblue] (0.1, 6.7) rectangle (2.6, 7.5);
    \node[text=bblue] at (1.35, 7.25) {\(V\)-trace};
    \node at (0.66, 6.9) (pol) {\(\pi\)};
    \node at (2, 6.9) (val) {\(V\)};
    
    
    
    \end{tikzpicture}}
    \vskip-0.33em
    \caption{HCAM in an agent.} \label{fig:HM_attention:agent}
    \end{subfigure}
    \caption{HCAM motivation and implementation. (\subref{fig:HM_attention:standard}) Standard transformer attention attends to every time step in a fixed window. (\subref{fig:HM_attention:hierarchical}) HCAM divides the past into chunks. It first attends to summaries of each chunk, and only attends in greater detail within relevant chunks. (\subref{fig:HM_attention:HCAM_block}) HCAM is implemented using a key-value memory, where the keys are chunk summaries, and the values are chunk sequences. Top-level attention is performed over all summaries (left), and then standard attention is performed within the top-\(k\) chunks. The results are relevance-weighted and summed (right), and then added to a residual input to produce the block output. (\subref{fig:HM_attention:agent}) Our agent encodes inputs, processes them through a 4-layer memory with an HCAM block in each layer, and decodes outputs for self-supervised and RL losses.}
    \label{fig:HM_attention}
\end{figure}
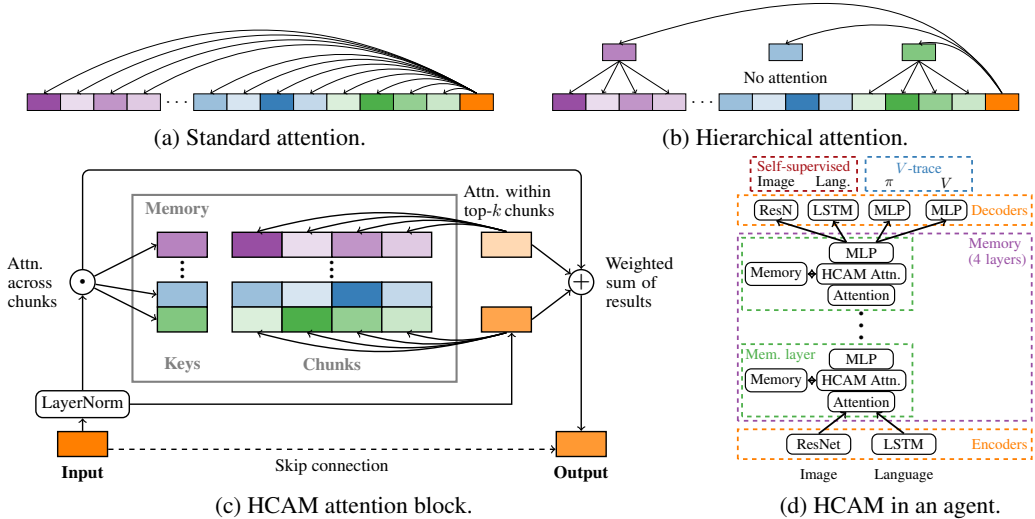

The HCAM uses a hierarchical strategy to attend over memory (Fig.\ \ref{fig:HM_attention}). Specifically, it divides the past into chunks. It stores each chunk in detail---as a sequence---together with a single summary key for that chunk. It can then recall hierarchically, by first attending to chunk summaries to choose the top-\(k\) relevant chunks, and then attending within those most relevant chunks in further detail.
In particular, let \(S\) denote the memory summary keys and \(Q\) be a linear projection (``query'') layer. Then the input is normalized by a LayerNorm, and the chunk relevance (\(R\)) is computed as:
\[
R = \text{softmax}(Q(\text{normed input}) \cdot S)
\]
Then the model attends in detail within the chunks that have the top-\(k\) relevance scores. If the memory chunks are denoted by \(\mathcal{C}\) and MHA(query inputs, key/value inputs) denotes multi-head attention:
\[
\text{memory query results} = \sum_{i\, \in \, \text{top-}k \text{ from } R} R_i \cdot \text{MHA}(\text{normed input}, \mathcal{C}_i)
\]
That is, HCAM selects the most relevant memory chunks, then time travels back into each of those chunks to query it in further detail, and finally aggregates the results weighted by relevance. 
This result is added to the input to produce the output of the block. An HCAM block can be \emph{added} to a transformer layer, to supplement standard attention. (For an open-source HCAM implementation, see App.\ \ref{app:methods:opensource})

HCAM can attend to any point in its memory, but at each step it only attends to a few chunks, which is much more compute- and memory-efficient than full attention. With \(N\) chunks in memory, each of size \(C\), using HCAM over the top-\(k\) chunks requires \(O(N + kC)\) attention computations, whereas full attention requires \(O(NC)\). HCAM succeeds even with \(k=1\) in many cases (App.\ \ref{app:supp_exp:varying_k}), reducing computation \(10\times\) compared to TrXL. Even with \(k \geq 1\), HCAM often runs faster (App.\ \ref{app:supp_exp:fps}).

\paragraph{Agents}
 Our agents (Fig.\ \ref{fig:HM_attention:agent}) consist of input encoders (a CNN or ResNet for visual input, and an LSTM for language), a 4-layer memory (one of HCAM, a Gated TransformerXL \citep{parisotto2020stabilizing}, or an LSTM \citep{hochreiter1997long}), followed by output MLPs for policy, value, and auxiliary losses. Our agents are trained using IMPALA \citep{espeholt2018impala}. To use HCAM in the memory, we adapt the Gated TransformerXL agent memory \citep{parisotto2020stabilizing}. We replace the XL memory with an HCAM block between the local attention (which provides short-term memory) and feed-forward blocks of each layer. We find that gating each layer \citep{parisotto2020stabilizing} is not beneficial to HCAM on our tasks (App.\ \ref{app:supp_exp:gating}), so we removed it. See App.\ \ref{app:methods:agents} for further agent details. 
 
 We store the \emph{initial layer inputs} in memory---that is, the inputs to the local attention---which performs better than storing the output of the local attention, and matches TransformerXL (TrXL) more closely. We accumulate these inputs into fixed-length memory chunks. When enough steps have accumulated to fill the current chunk, we add the current chunk to the memory. We average-pool across the chunk to generate a summary. We reset the memory between episodes, except where cross-episode memory is tested (Section \ref{sec:exp:fastbind}). We stop gradients from flowing into the memory, just as TrXL \citep{dai2019transformer} stops gradients flowing before the current window. This reduces challenges with the non-differentiability of top-\(k\) selection (see App.\ \ref{app:supp_exp:varying_k} for discussion). It also means that we only need to store past activations in memory, and not the preceding computations, so we can efficiently store a relatively long time period.
 
 However, the model must therefore learn to encode over shorter timescales, since it cannot update the encoders through the memory. Thus, as in prior work referenced above, we show that \emph{self-supervised learning} \citep{liu2020self}, is necessary for successful learning on our tasks (App.\ \ref{app:supp_exp:selfsupervised}). Specifically, we use an auxiliary \emph{reconstruction} loss---at each step we trained the agents (of all memory types) to reconstruct the input image and language as outputs \citep{hill2020grounded}. This forces the agents to propagate these task-relevant features through memory, which ensures that they  be encoded \citep[cf.][]{hung2019optimizing,fortunato2019generalization}.

\section{Experiments}
\begin{figure}[tbh]
    \centering
    \begin{subfigure}[b]{0.33\textwidth}
    \centering
    \includegraphics[width=0.45\textwidth]{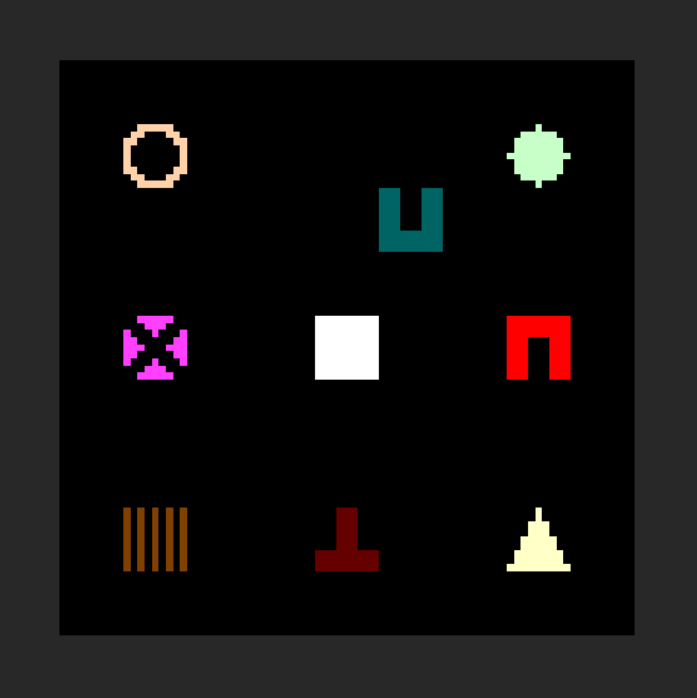}
    \vskip-0.1em
    \caption{Ballet.}
    \label{fig:exp:task_overview:ballet}
    \end{subfigure}%
    \begin{subfigure}[b]{0.33\textwidth}
    \centering
    \includegraphics[width=0.63\textwidth]{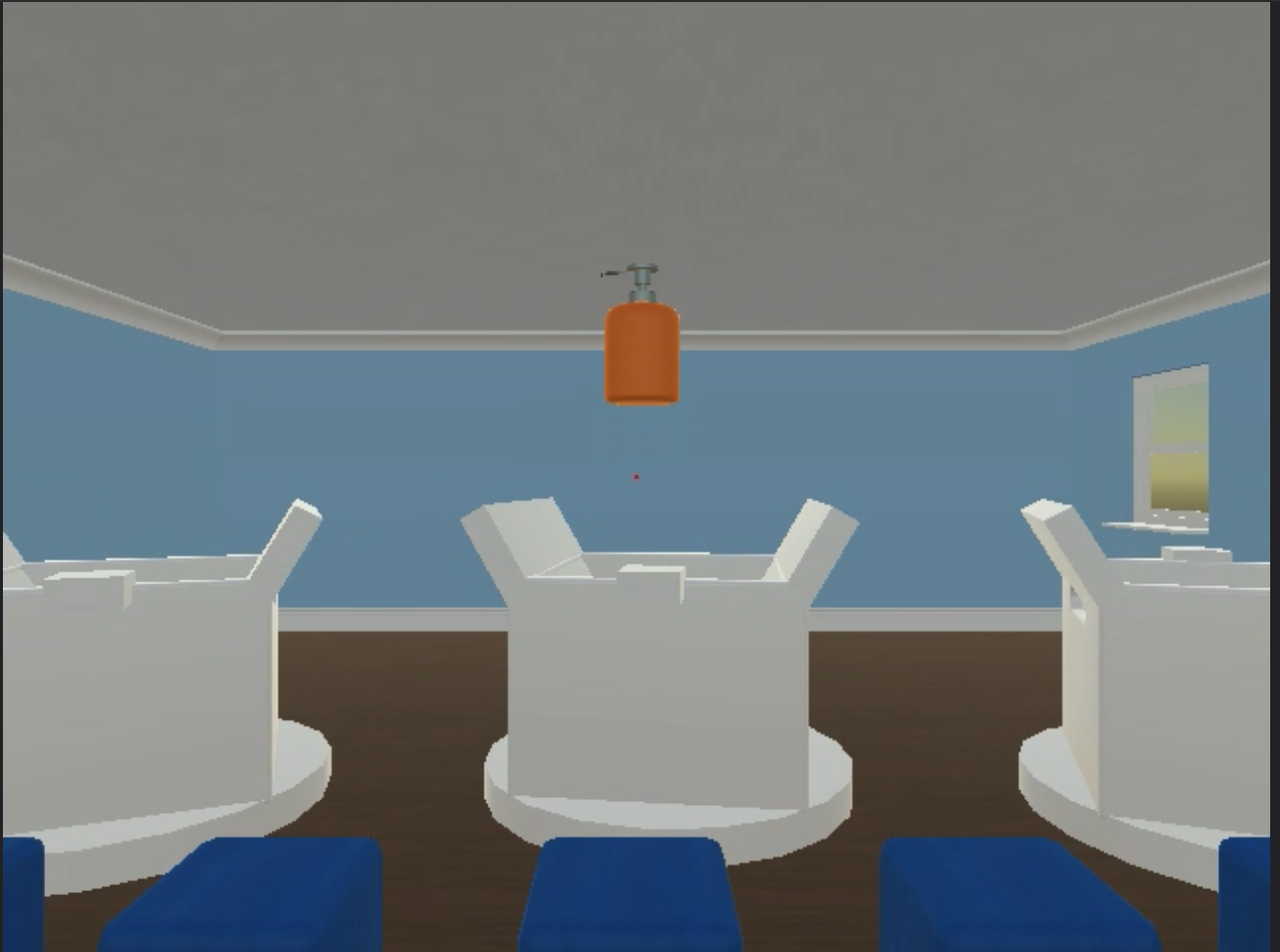}
    \vskip-0.1em
    \caption{Object permanence.}
    \label{fig:exp:task_overview:object_permanence}
    \end{subfigure}%
    \begin{subfigure}[b]{0.33\textwidth}
    \centering
    \includegraphics[width=0.63\textwidth]{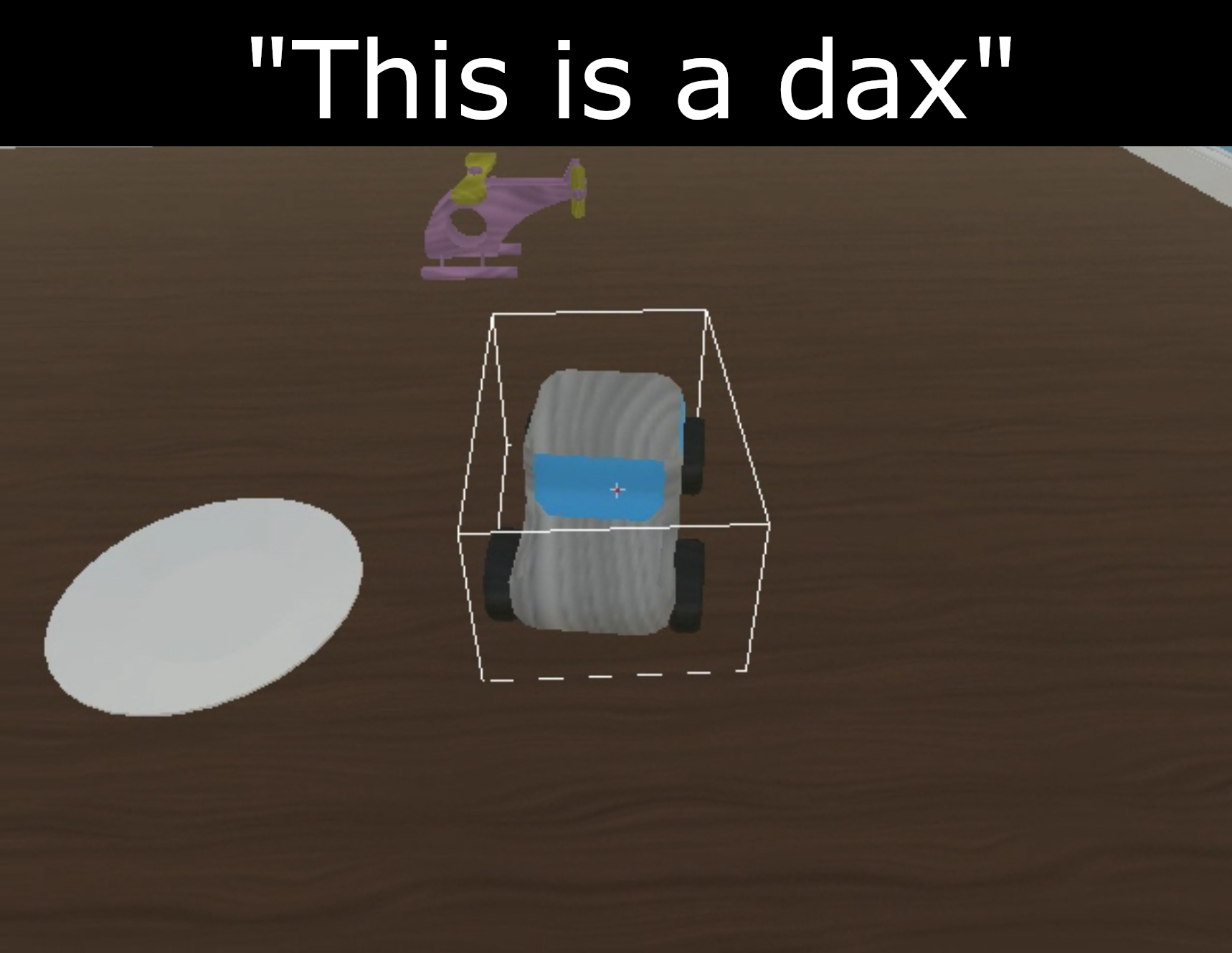}
    \vskip-0.1em
    \caption{Rapid word learning.}
    \label{fig:exp:task_overview:fast_binding}
    \end{subfigure}\\
    \begin{subfigure}[b]{0.33\textwidth}
    \includegraphics[width=\textwidth]{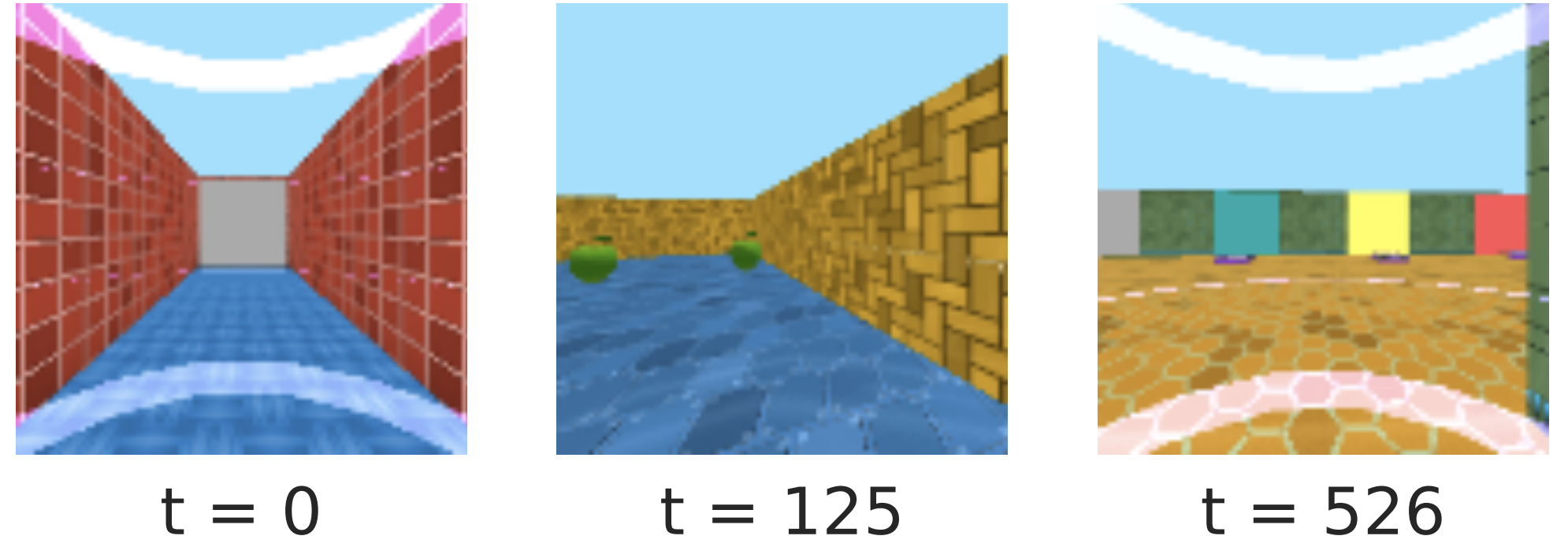}
    \vskip-0.2em
    \caption{Visual match.}
    \label{fig:exp:task_overview:tvt}
    \end{subfigure}%
    \begin{subfigure}[b]{0.33\textwidth}
    \centering
    \includegraphics[width=0.9\textwidth]{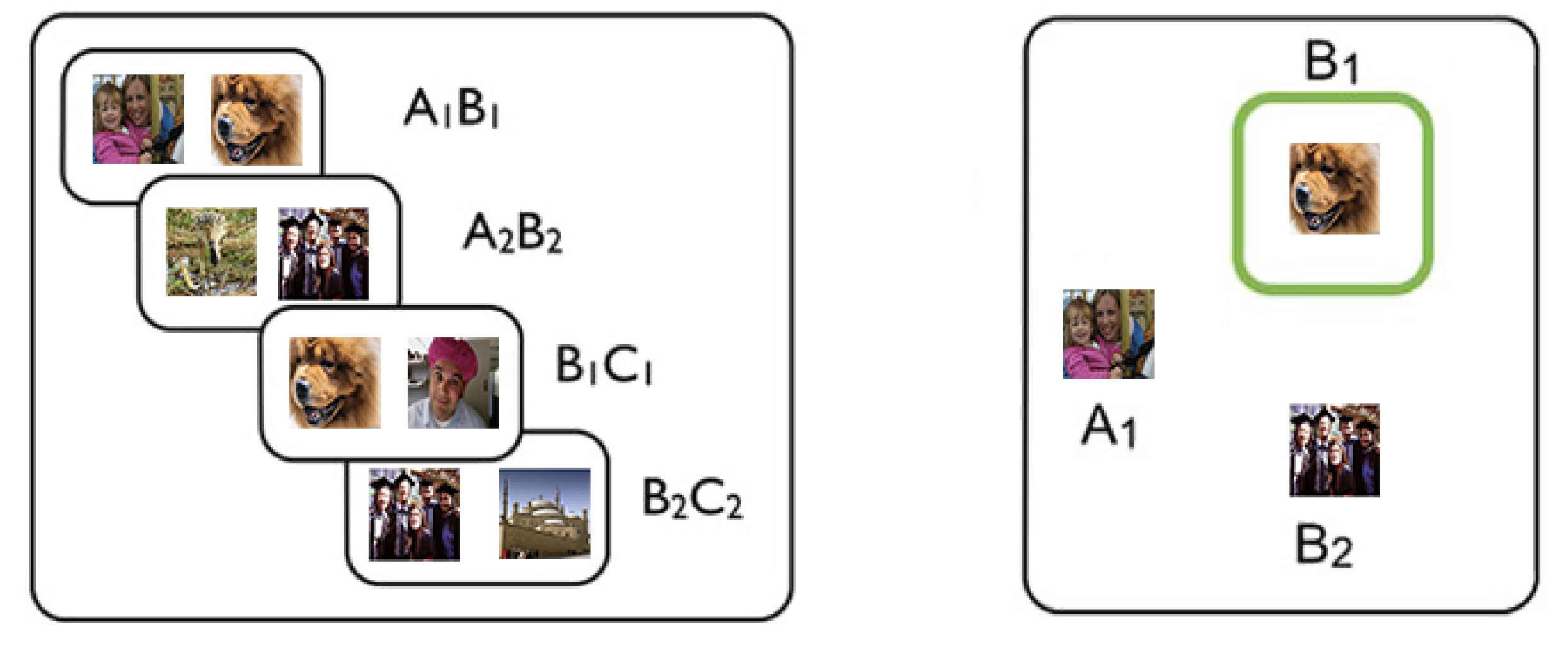}
    \vskip-0.2em
    \caption{Paired associative inference.}
    \label{fig:exp:task_overview:pai}
    \end{subfigure}%
    \begin{subfigure}[b]{0.33\textwidth}
    \centering
    \includegraphics[width=0.7\textwidth]{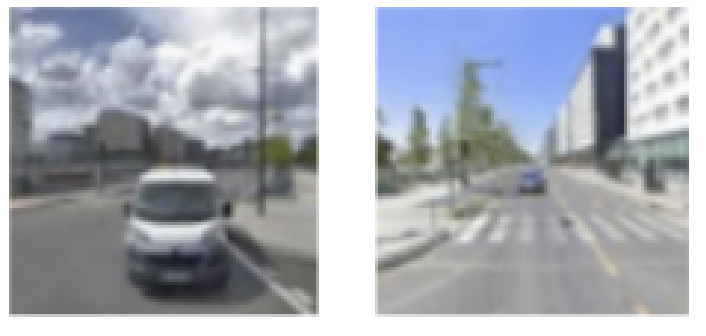}
    \vskip-0.2em
    \caption{One-shot StreetLearn navigation.}
    \label{fig:exp:task_overview:ossl}
    \end{subfigure}%
    \caption{We evaluated our model across a wide variety of task domains, some of which are modified or used from other recent memory papers \citep{hill2020grounded,hung2019optimizing,banino2020memo,ritter2020rapid}. These include a variety of environments (2D, 3D, real images) and tasks (finding hidden objects, language learning, and navigation).} 
    \label{fig:exp:task_overview}
    \vspace{-0.8em}
\end{figure}

We evaluated HCAM in a wide variety of domains (Fig.\ \ref{fig:exp:task_overview}). Our tasks test the ability of HCAM to recall event details (Sec. \ref{sec:exp:ballet}); to maintain object permanence (Sec. \ref{sec:exp:objects}); to rapidly learn and recall new words, and extrapolate far beyond the training distribution (Sec. \ref{sec:exp:fastbind}); and to reason over multiple memories to generalize, even in settings that previously required task-specific architectures (Sec. \ref{sec:exp:other}).

\subsection{Remembering the ballet} \label{sec:exp:ballet}
Our first experiment tests the ability to recall spatiotemporal detail, which is important in real-world events. We place the agent in a 2D world, surrounded by ``dancers'' with different shapes and colors (Fig.\ \ref{fig:exp:task_overview:ballet}). One at a time, each dancer performs a 16-step solo dance (moves in a distinctive pattern), with a 16- or 48-step delay between dances. After all dances, the agent is cued to e.g.\ ``go to the dancer who danced a square,'' and is rewarded if it goes to the correct dancer. This task requires recalling dances in detail, because no single step of a dance is sufficient to distinguish it from other dances. The task difficulty depends on the number of dances the agent must watch in an episode (between 2 and 8) and the delay length. We varied both factors across episodes during training---this likely increases the memory challenges, since the agent must adapt in different episodes to recalling different amounts of information for varying periods. See App.\ \ref{app:tasks:ballet} for details.

\begin{figure}[tbh]
    \centering
    \begin{subfigure}{0.33\textwidth}
    {\includegraphics[width=\textwidth]{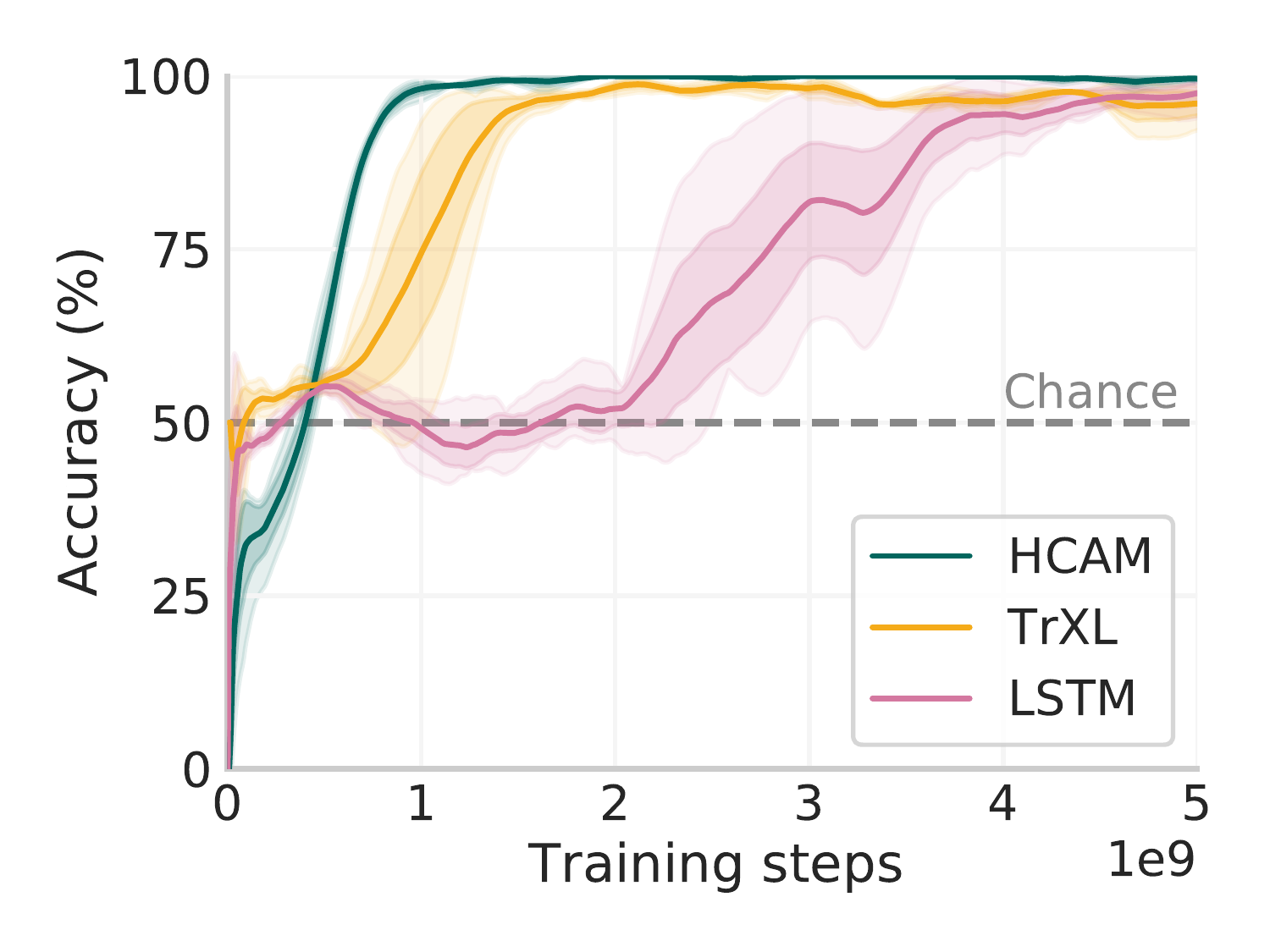}}
    \vskip-0.5em
    \caption{2 dances.}
    \label{fig:exp:ballet:2}
    \end{subfigure}%
    \begin{subfigure}{0.33\textwidth}
    {\includegraphics[width=\textwidth]{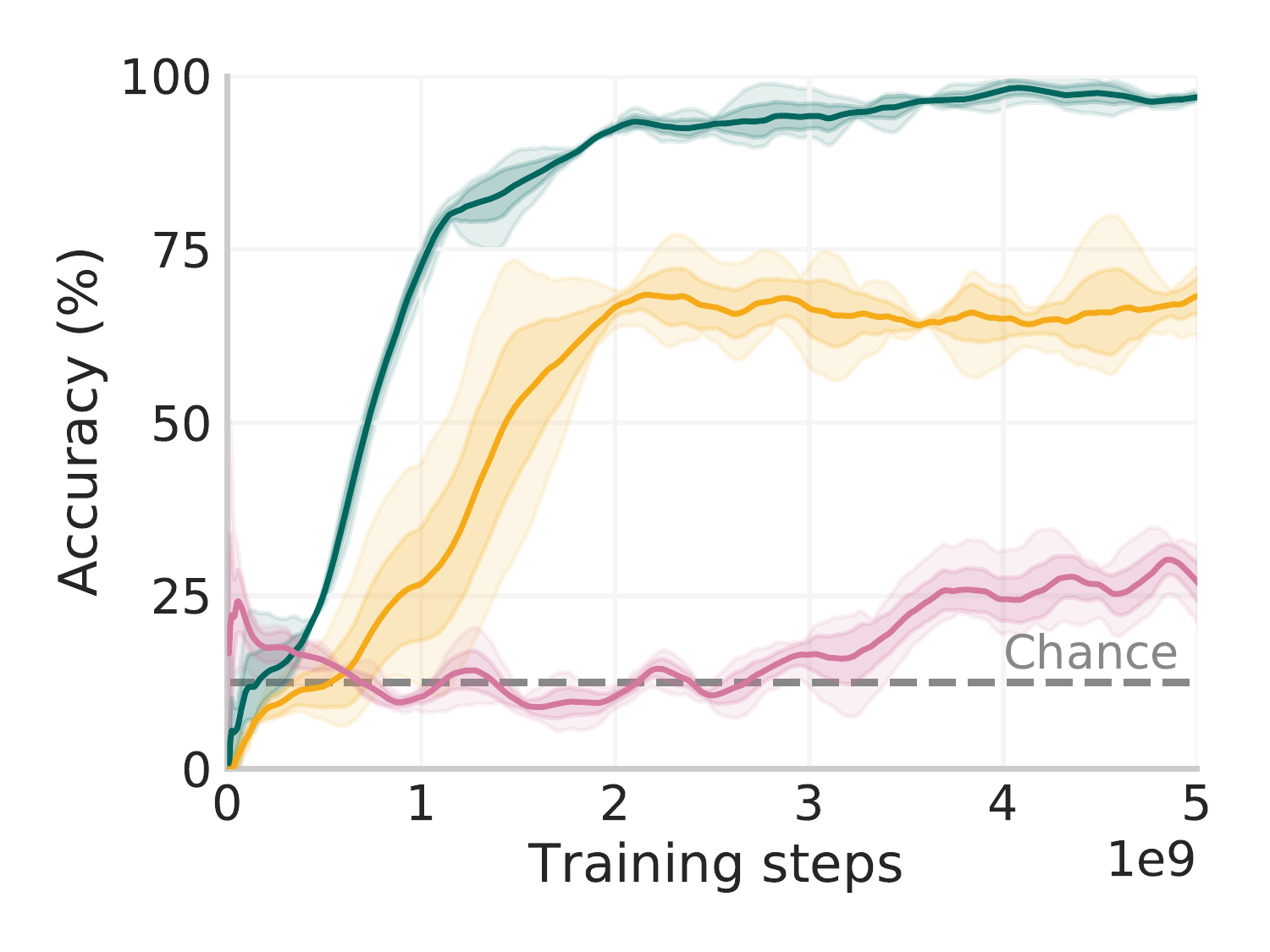}}
    \vskip-0.5em
    \caption{8 dances.}
    \label{fig:exp:ballet:8}
    \end{subfigure}%
    \begin{subfigure}{0.33\textwidth}
    {\includegraphics[width=\textwidth]{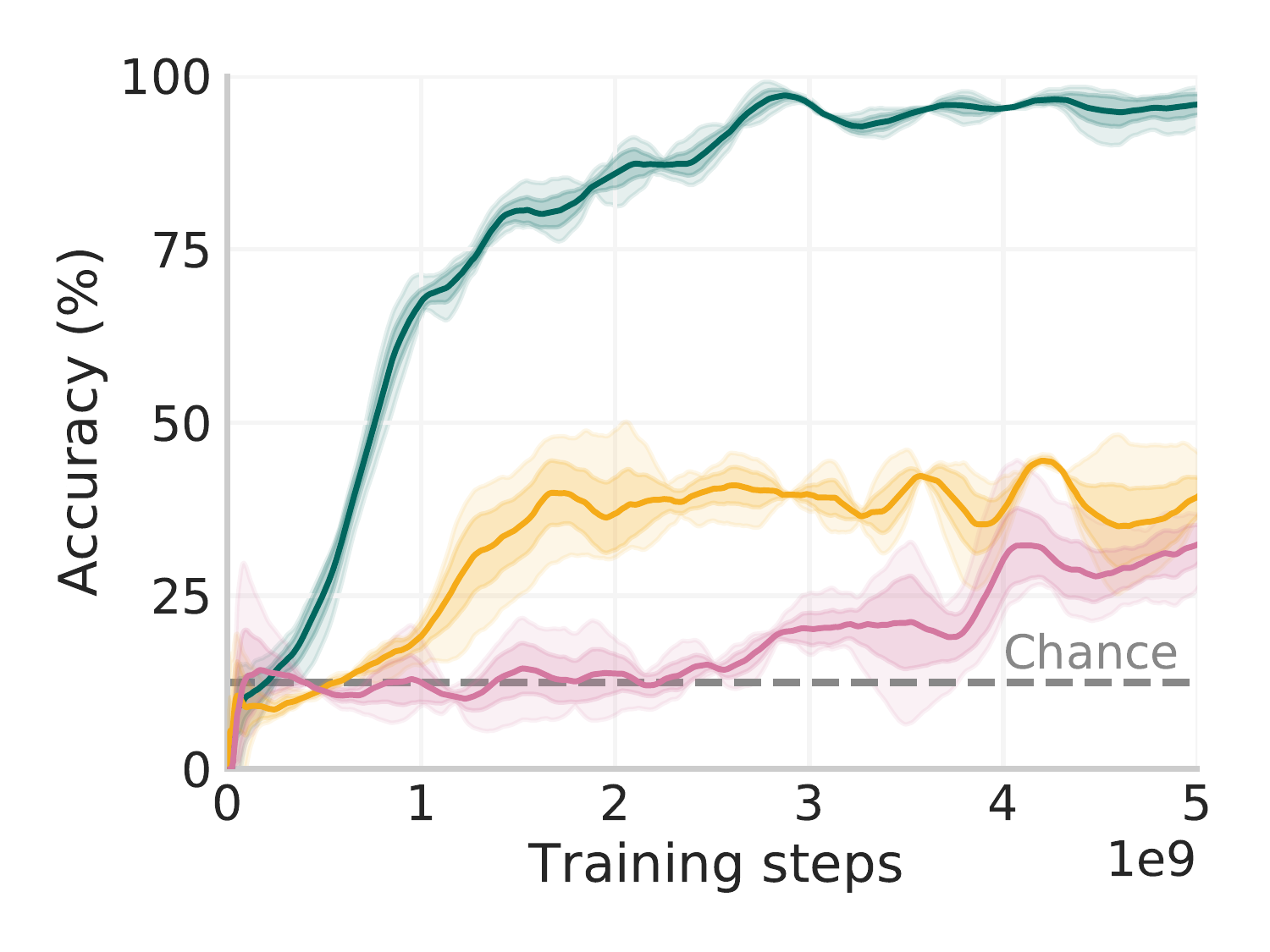}}
    \vskip-0.5em
    \caption{8 dances, long delays.}
    \label{fig:exp:ballet:8long}
    \end{subfigure}
    \caption{The ballet task---HCAM (teal) outperforms TrXL (yellow) and LSTM (red) memories at difficult ballets with many dances, especially when there are long delays between. (\subref{fig:exp:ballet:2}) In the shortest ballets, TrXL performs nearly as well as HCAM, while LSTMs require much more training. (\subref{fig:exp:ballet:8}) With 8 dances, HCAM substantially outperforms TrXL, even though the entire task fits within one TrXL attention window. LSTMs do not get far above chance at choosing among the dancers. (\subref{fig:exp:ballet:8long}) With longer delays between dances, HCAM performs similarly but TrXL degrades further. (3 runs per condition, lines=means, light regions=range, dark=\(\pm\)SD. Training steps are agent/environment steps, not the number of parameter updates. Chance denotes random choice among the dancers, not random actions.)}
    \label{fig:exp:ballet}
    \vspace{-0.5em}
\end{figure}

Fig.\ \ref{fig:exp:ballet} shows the results. Agents with an HCAM or a Gated TransformerXL (TrXL) memory perform well at the shortest task, with only 2 dances per episode (Fig.\ \ref{fig:exp:ballet:2}). LSTM agents learn much more slowly, showing the benefits of attention-based memories. With 8 dances per episode (Fig.\ \ref{fig:exp:ballet:8}), HCAM continues to perform well. However, TrXL's performance degrades, \textbf{even though the entire task is short enough to fit within a single TrXL attention window}. (The dances are largely in the XL region, where gradients are stopped for TrXL, but HCAM has the same limitation.) LSTM performance also degrades substantially. The advantage of HCAM is even more apparent with longer delays between the dances (Fig.\ \ref{fig:exp:ballet:8long}). HCAM can also generalize well (\(\geq 90\)\%) from training on up to 6 dances to testing on 8 (App.\ \ref{app:supp_exp:ballet_gen}). Furthermore, HCAM is robust to varying memory chunk sizes (App.\ \ref{app:supp_exp:varying_chunk_size}), even if the chunk segmentation is not task aligned. Finally, HCAM can perform well on this task even if it is only allowed to select a single memory chunk rather than the top-\(k\) (App.\ \ref{app:supp_exp:varying_k}), thereby attending to \(8\times\) fewer time-points than TrXL (while a sparse TrXL is not advantageous, see App.\ \ref{app:supp_exp:topk_trxl}). Thus HCAM can increase computational efficiency. In summary, both attention-based memories recall events better than LSTMs, and HCAM robustly outperforms TrXL at all but the easiest tasks.

\subsection{Object permanence} \label{sec:exp:objects}
\begin{figure}[tbh]
    \centering
    \begin{subfigure}{\textwidth}
    \centering
    \resizebox{0.75\textwidth}{!}{
    \input{figures/task_images/object_permanence/object_permanence_diagram}}\vspace{-0.5em}
    \caption{The object permanence task.} \label{fig:exp:objects:task}
    \end{subfigure} \\
    \begin{subfigure}[t]{0.33\textwidth}
    \includegraphics[width=\textwidth]{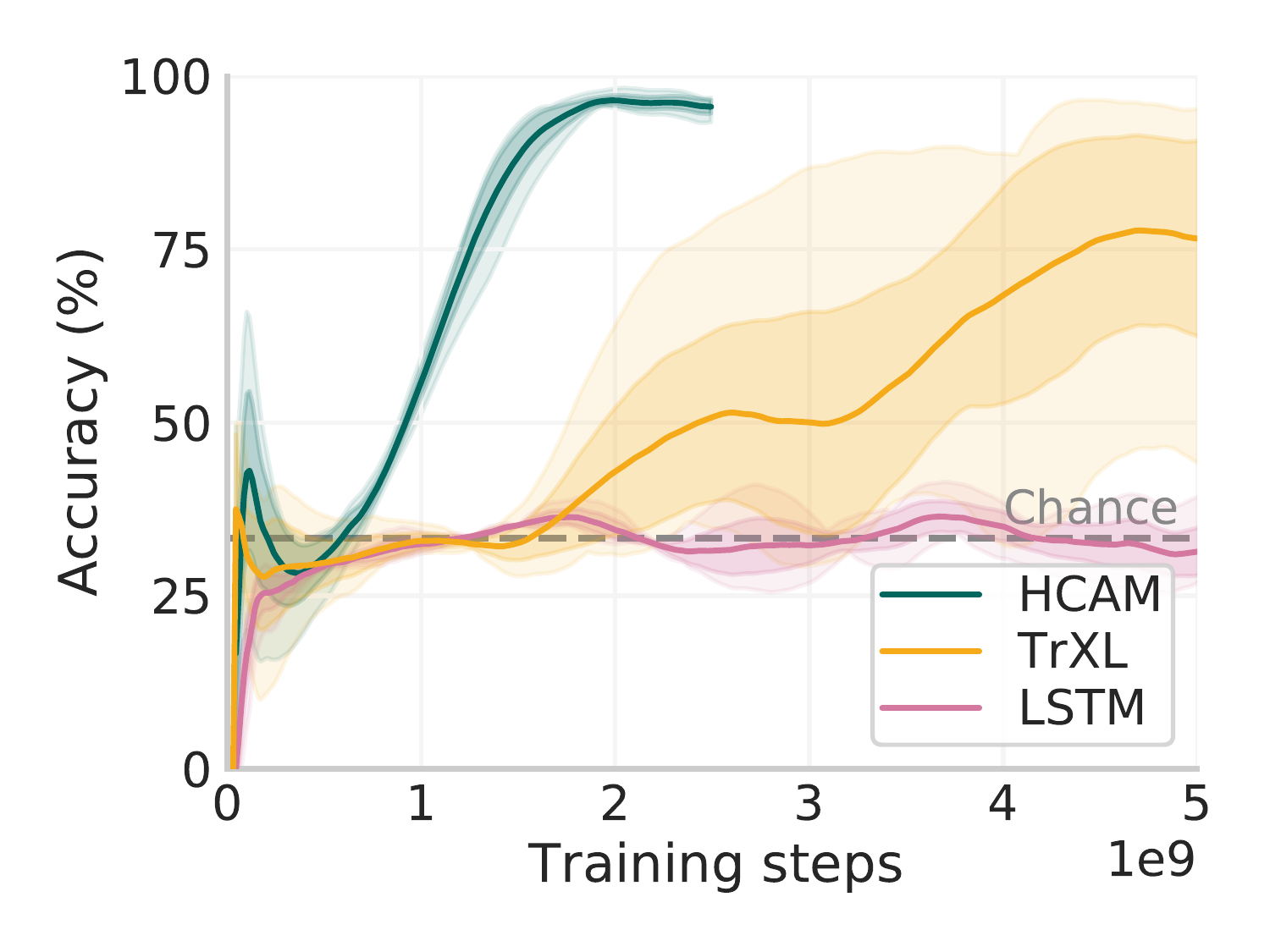}
    \vskip-0.5em
    \caption{No delays, varying training.}
    \label{fig:exp:objects:curr_short}
    \end{subfigure}%
    \begin{subfigure}[t]{0.33\textwidth}
    \includegraphics[width=\textwidth]{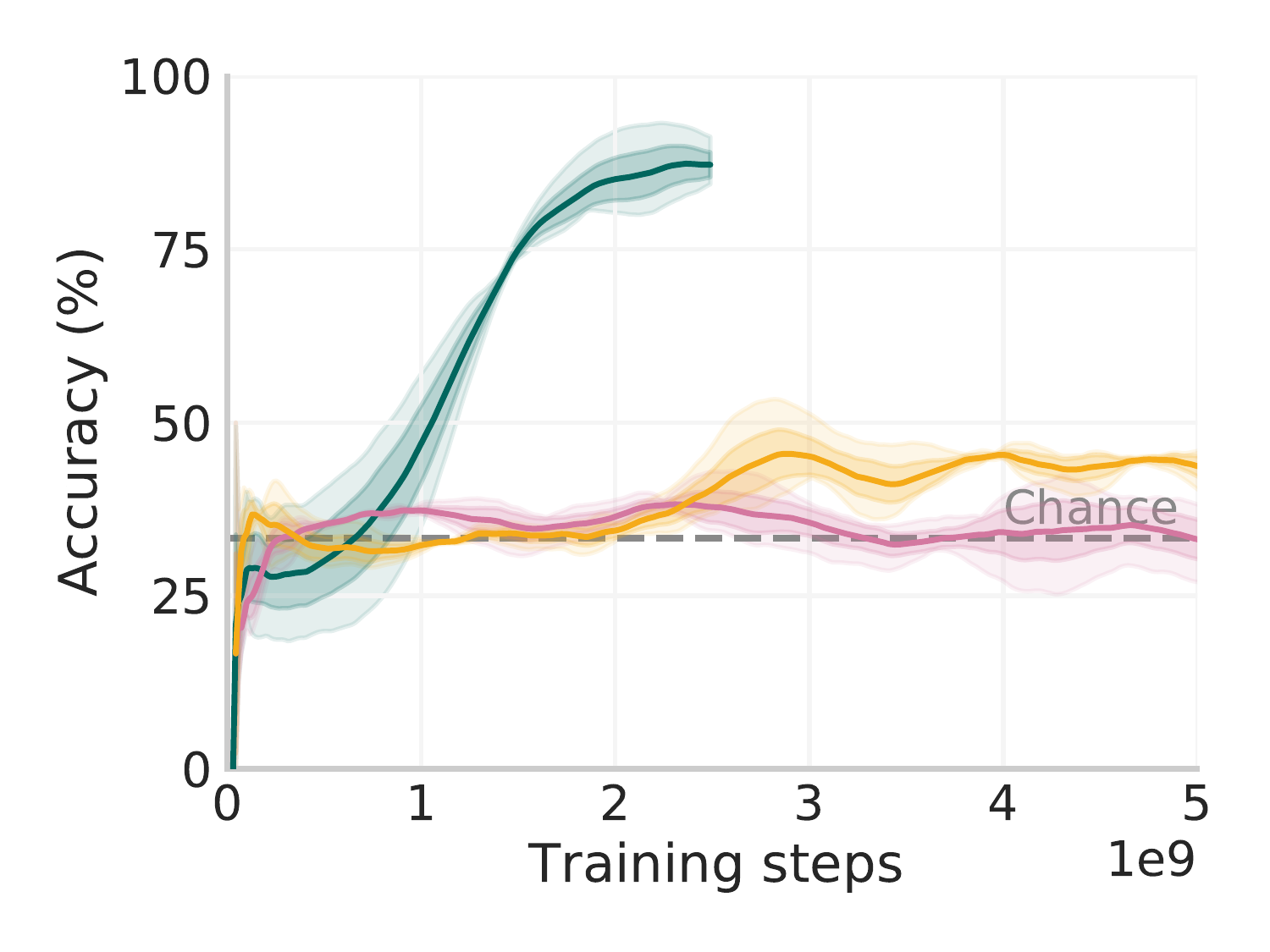}
    \vskip-0.5em
    \caption{Long delays, varying training.}
    \label{fig:exp:objects:curr_long}
    \end{subfigure}%
    \begin{subfigure}[t]{0.33\textwidth}
    \includegraphics[width=\textwidth]{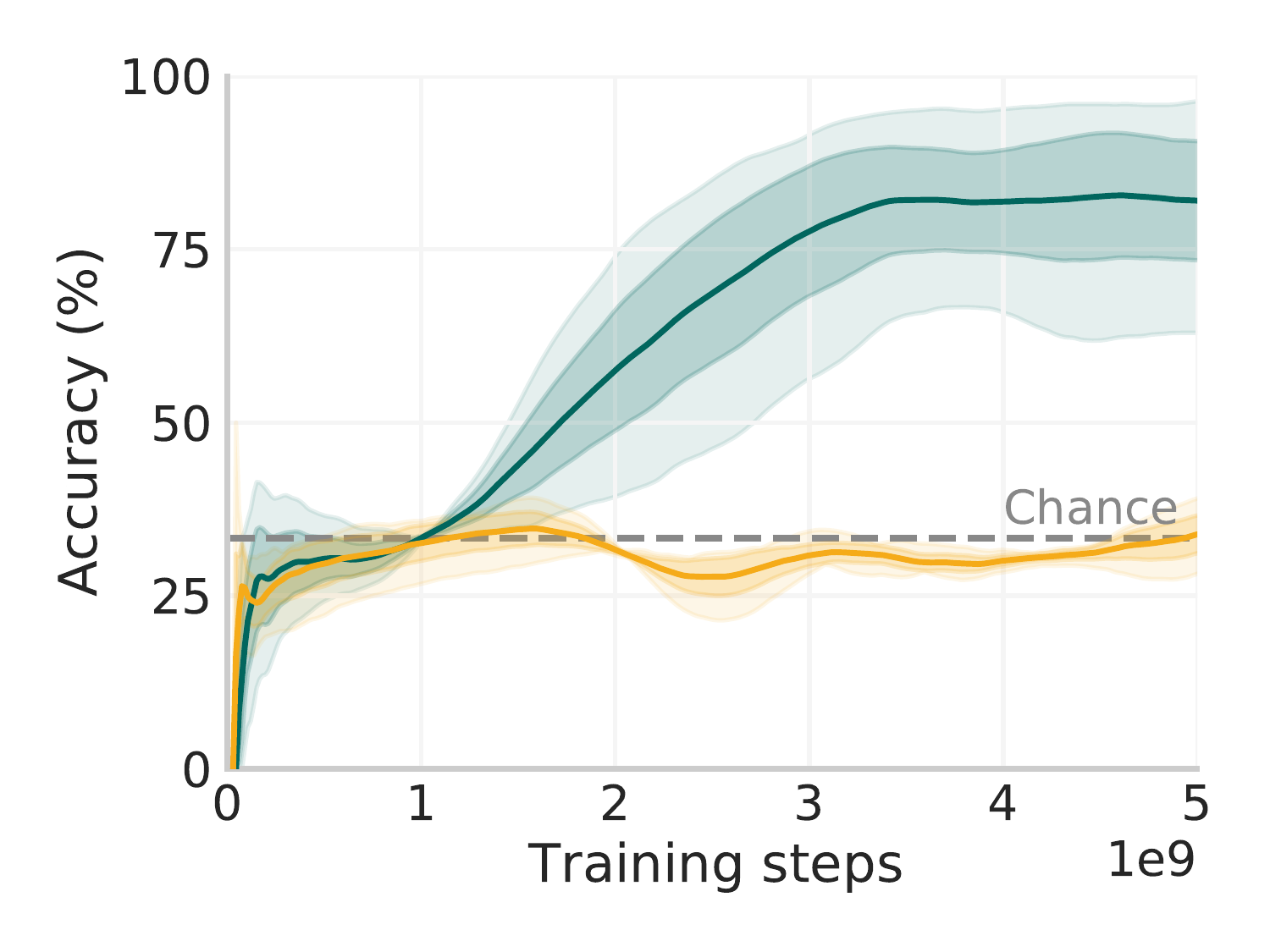}
    \vskip-0.5em
    \caption{Long delays, long-only training.}
    \label{fig:exp:objects:longonly}
    \end{subfigure}
    \caption{The object permanence task---HCAM learns faster than TrXL, and succeeds at longer tasks. (\subref{fig:exp:objects:task}) The task structure.  (\subref{fig:exp:objects:curr_short}) The shortest tasks, with no delays. When trained with varying delay lengths, HCAM rapidly and consistently learns to track object locations. TrXL learns more slowly and less consistently. (\subref{fig:exp:objects:curr_short}) Long tasks, with 30-second delays between each object being revealed. HCAM learns quite well, while TrXL barely achieves above-chance performance. (\subref{fig:exp:objects:longonly}) HCAM can learn the long tasks even without simultaneous training on the shorter tasks to scaffold its learning.  (3 runs per condition. Chance denotes random choice among the boxes, not random actions.)}
    \label{fig:exp:objects}
    \vspace{-0.9em}
\end{figure}

Our next tasks test the ability to remember the identity and location of objects after they have been hidden within visually-identical boxes, despite delays and looking away. These tasks are inspired by work on object permanence in developmental psychology \citep[e.g.][]{baillargeon2004infants}. We implemented these tasks in a 3D Unity environment. The agent is placed in a room and fixed in a position where it can see three boxes (Fig.\ \ref{fig:exp:objects:task}). One at a time, an object jumps out of each box a few times, and then returns to concealment inside. There may be a delay between one object and the next. After all objects have appeared, the box lids close. Then, the agent is released, and must face backwards. Finally, it is asked to go to the box with a particular object, and rewarded if it succeeds.

HCAM solves these tasks more effectively than the baselines. HCAM learns the shortest tasks (with no delays) substantially faster than TrXL (Fig.\ \ref{fig:exp:objects:curr_short}), and LSTMs are unable to learn at all. When trained on tasks of varying lengths, HCAM was also able to master tasks with 30 seconds of delay after each object revealed itself, while TrXL performed near chance even with twice as much training (Fig, \ref{fig:exp:objects:curr_short}). Furthermore, HCAM can learn these long tasks even without the shorter tasks to scaffold it (Fig.\ \ref{fig:exp:objects:longonly}).

\subsection{Rapid word learning with distractor tasks \& generalization across episodes} \label{sec:exp:fastbind}

\begin{figure}[tbh]
    \centering
    \input{figures/task_images/fast_binding/fast_binding_diagram}%
    \begin{subfigure}[b]{0.33\textwidth}
    \centering
    \includegraphics[width=\textwidth]{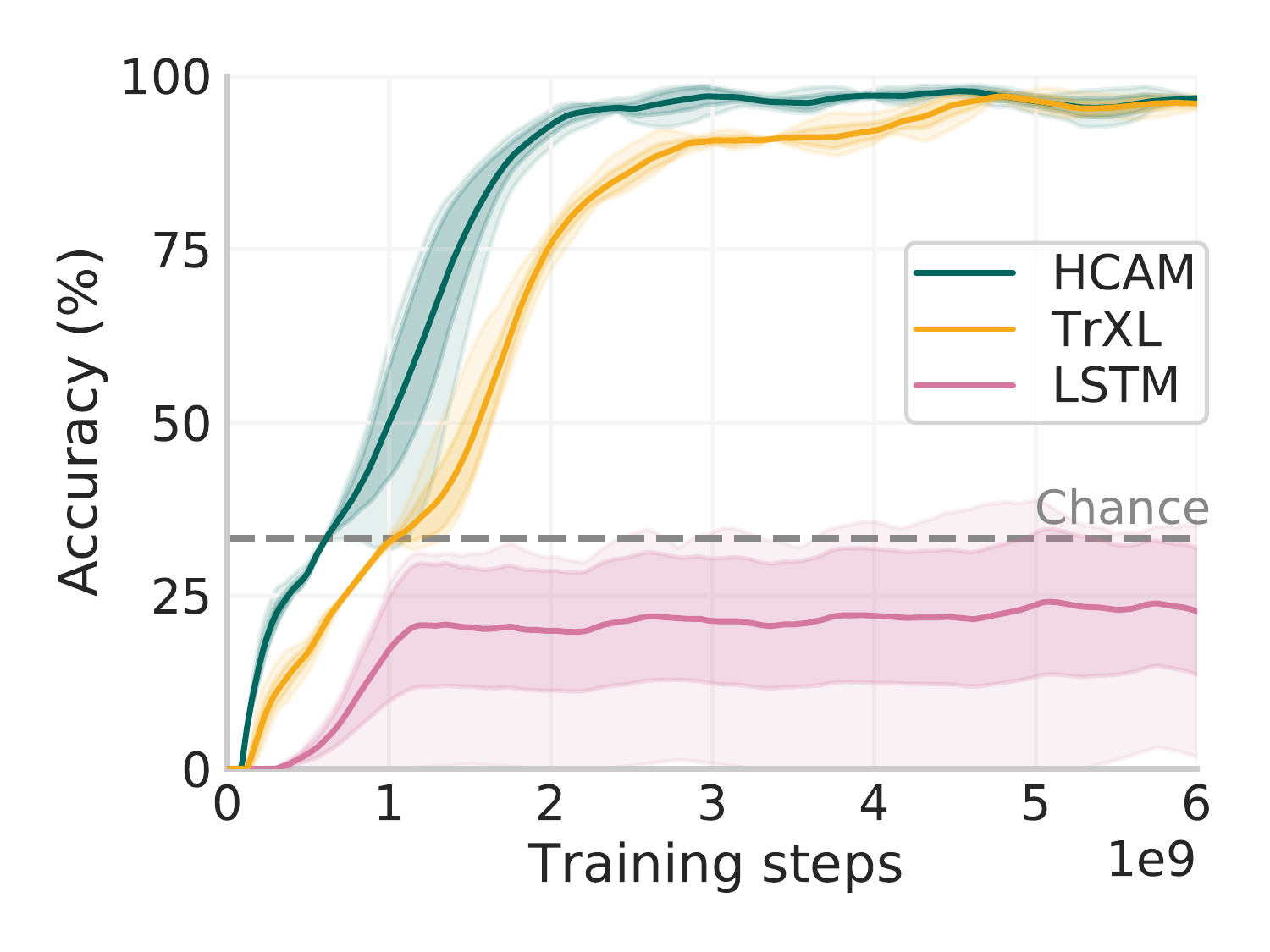}
    \vskip-0.5em
    \captionsetup{width=.8\textwidth}
    \caption{Train task with 2 distractors.}
    \label{fig:exp:fastbind:train}
    \end{subfigure}\\
    \begin{subfigure}{0.33\textwidth}
    \centering
    \includegraphics[width=\textwidth]{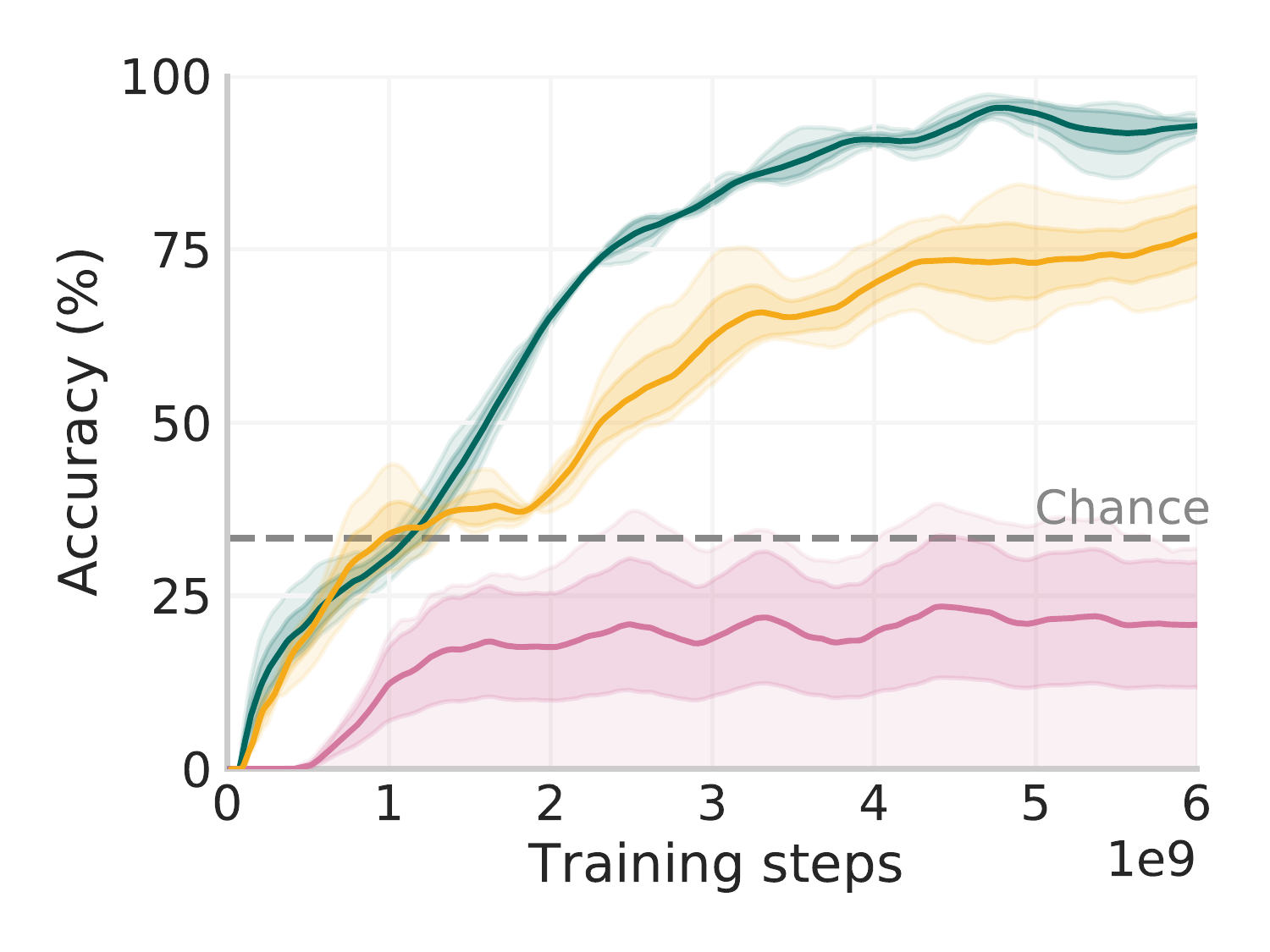}
    \vskip-0.5em
    \captionsetup{width=.92\textwidth}
    \caption{Eval.\ with 10 distractors.}
    \label{fig:exp:fastbind:curr}
    \end{subfigure}%
    \begin{subfigure}{0.33\textwidth}
    \centering
    \includegraphics[width=\textwidth]{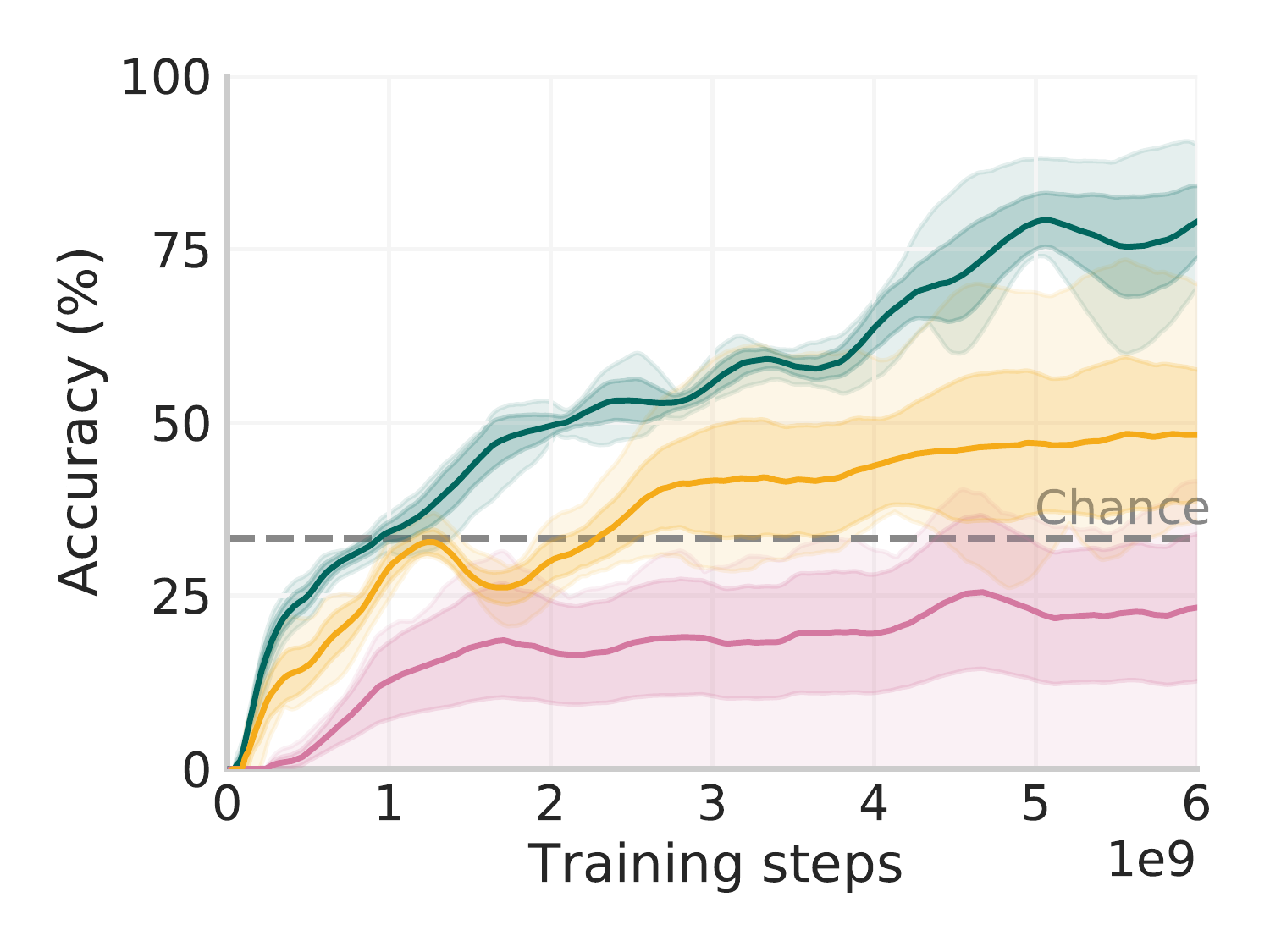}
    \vskip-0.5em
    \captionsetup{width=.92\textwidth}
    \caption{Eval.\ across 4 eps.,\ 0 dist.}
    \label{fig:exp:fastbind:across_eps_4_1}
    \end{subfigure}%
    \begin{subfigure}{0.33\textwidth}
    \centering
    \includegraphics[width=\textwidth]{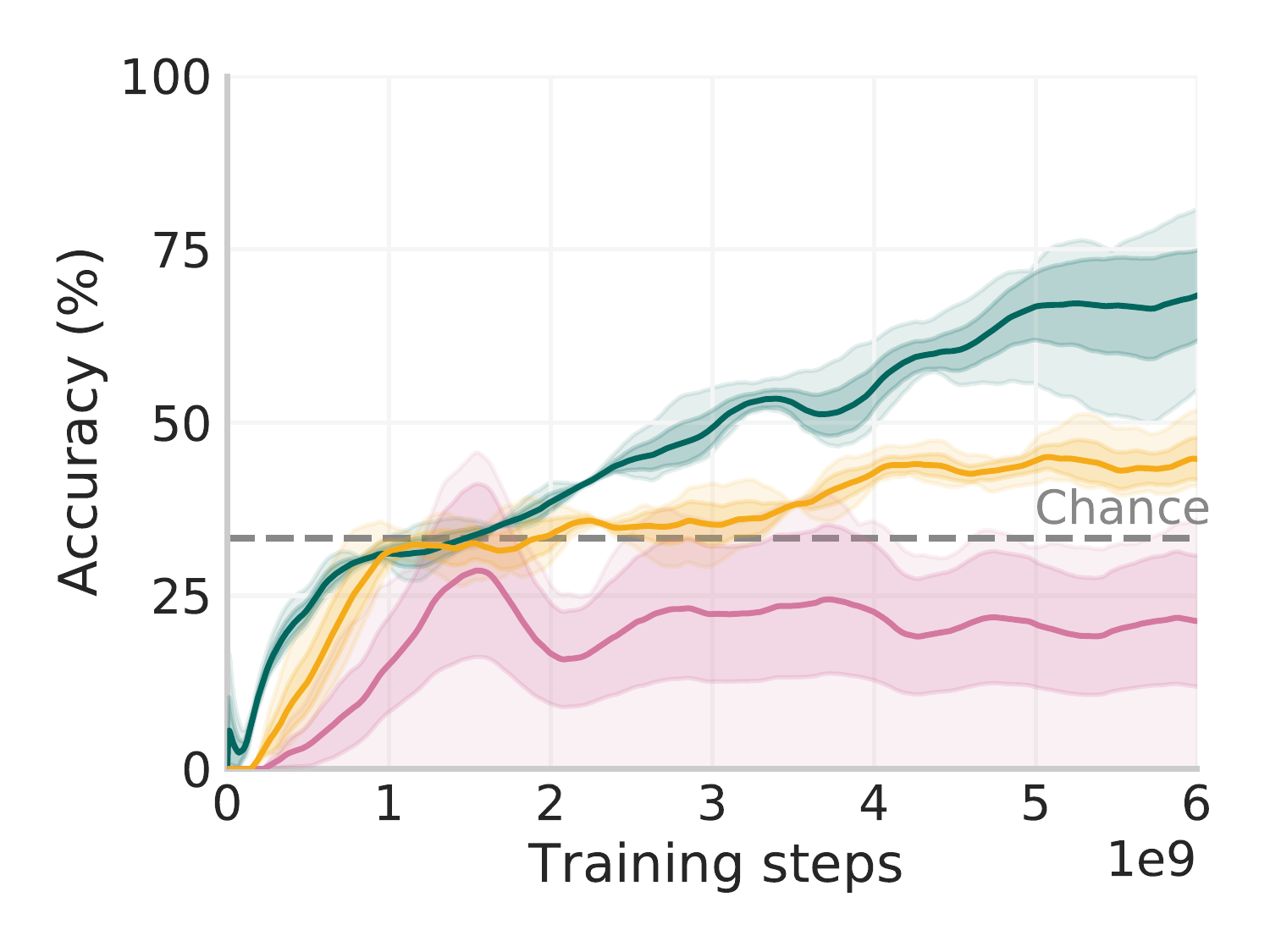}
    \vskip-0.5em
    \captionsetup{width=.92\textwidth}
    \caption{Eval.\ across 2 eps.,\ 5 dist.\ each.}
    \label{fig:exp:fastbind:across_eps_2_5}
    \end{subfigure}%
    \caption{The rapid word-learning tasks. (\subref{fig:exp:fast_binding_tasks:distractors}) Training tasks consist of learning three words (nouns), completing distractor tasks, and then finding an object named using one of the three learned words. We evaluated generalization to larger numbers of distractors. (\subref{fig:exp:fast_binding_tasks:episodes}) During training, the agent never had to recall words for more than a single episode. However, we also evaluated the agent on its ability to recall a word learned several episodes earlier. This tests the ability of our memory to bridge from rapid meta-learning to longer-term retention. (\subref{fig:exp:fastbind:train}) Agents with HCAM or TrXL memories learn the hardest training task (2 distractors) well, although the agent with HCAM is faster to learn. When trained only on single episodes with 0-2 distractor phases: (\subref{fig:exp:fastbind:curr}) HCAM outperforms TrXL at extrapolation to 10 distractor phases. (\subref{fig:exp:fastbind:across_eps_4_1}-\subref{fig:exp:fastbind:across_eps_2_5}) HCAM can generalize strongly out-of-distribution to recalling words that were learned several episodes earlier, despite having completed other episodes in the intervening time. (3 runs per condition. Chance denotes random performance on the final evaluation choice; this requires complex behaviours: learning all names and completing all distractors and intermediate episodes to reach the final choice. This is why LSTM performance is below chance in one seed.)}
    \label{fig:exp:fast_binding}
    \vspace{-1.1em}
\end{figure}

The preceding tasks focused on passively absorbing events, and then recalling them after delays. We next turn to a set of rapid-word-learning tasks (Fig.\ \ref{fig:exp:task_overview:fast_binding}) that require actively acquiring knowledge, and maintaining it in more challenging settings, with intervening distractor tasks (Fig: \ref{fig:exp:fast_binding_tasks:distractors}). These tasks are based on the work of \citet{hill2020grounded}, who showed that transformer-based agents in a 3D environment can meta-learn to bind objects to their names after learning each name only once. That is, immediately after a single exposure to novel object names, their agent can successfully lift objects specified by those names. Their agent then discards its memory before proceeding to the next episode. However, in more complex environments tasks will not be so cleanly structured. There will often be distracting events and delays---or even learning of other information---before the agent is tested.

We therefore created more challenging versions of these tasks by inserting distractor tasks between the learning and test phases. In each distractor task we placed the agent in a room with a fixed set of three objects (that were never used for the word-learning portion), and asked it to lift one of those objects (using fixed names). By varying the number of distractor phases, we were able to manipulate the demands on the memory. We trained our agents on tasks with 0, 1, or 2 distractor phases, and evaluated their ability to generalize to tasks with more distractors. HCAM and TrXL could learn the training tasks equally well, and both showed some ability to generalize to longer tasks (App.\ \ref{app:supp_exp:fastbind}). However, HCAM was able extrapolate better to tasks with 10 distractor phases (Fig.\ \ref{fig:exp:fastbind:curr}), and even showed some extrapolation to 20 distractors, an order of magnitude more than trained (Fig. \ref{fig:supp_exp:fastbind_harder:20}). 

\paragraph{Maintaining knowledge across episodes zero-shot}
It is a fundamental limitation of most meta-learning paradigms that they discard the information they have learned after each episode. The agents used by \citet{hill2020grounded} discarded their memories end of each episode. Children, by contrast, can rapidly learn new words and maintain that knowledge even after learning other new words. This leads to a question: if HCAM can effectively maintain a word-object mapping despite intervening distractor tasks, could it maintain this knowledge across entire episodes of learning new words?

To answer this question, we used the agents trained on single-episode tasks. We evaluated these agents on tasks where they were sometimes tested on a word from several episodes before, despite having learned and been tested on other words in the intervening period (Fig.\ \ref{fig:exp:fast_binding_tasks:episodes}). In training, the agents were never required to maintain a set of words once they encountered a new learning phase with new words. Nevertheless, our agent with HCAM is able to generalize well to recalling words four episodes later (Fig.\ \ref{fig:exp:fastbind:across_eps_4_1}). It can even extrapolate along multiple dimensions, recalling words two episodes later with more distractor phases per episode than it ever saw in training (Fig.\ \ref{fig:exp:fastbind:across_eps_2_5}). In App.\ \ref{app:supp_exp:attention_analyses} we show how the attention patterns of the hierarchical memory support this generalization across episodes. \textbf{Agents with HCAM are able to generalize to memory tasks that are far from the training distribution.}


\subsection{Comparisons with alternative memory systems} \label{sec:exp:other}

There has been an increasing interest in architectures for memory in RL, as well as in deep learning more broadly. In order to help situate our approach within this literature, we benchmarked our model against tasks used in several recent papers exploring alternative forms of memory. HCAM is able to perform well across all the tasks---reaching near-optimal performance, comparable with memory architectures that were specifically engineered for each task---and outperforms strong baselines.\vspace{-0.5em}

\begin{figure}[tbh]
    \centering
    \begin{subfigure}[b]{0.33\textwidth}
    \includegraphics[width=\textwidth]{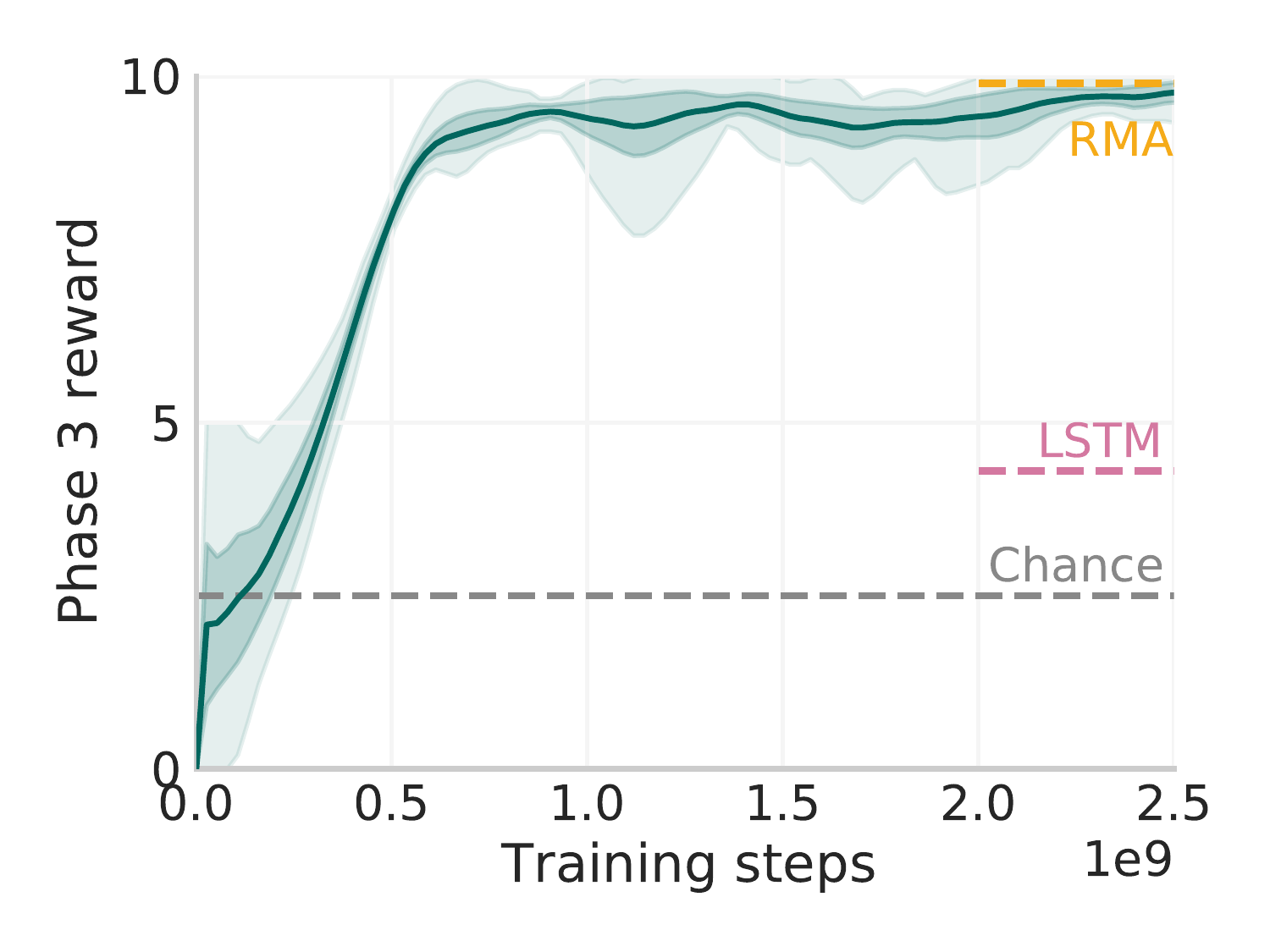}
    \vskip-0.5em
    \caption{Passive Visual Match.}
    \label{fig:exp:other_papers:tvt}
    \end{subfigure}%
    \begin{subfigure}[b]{0.33\textwidth}
    \includegraphics[width=\textwidth]{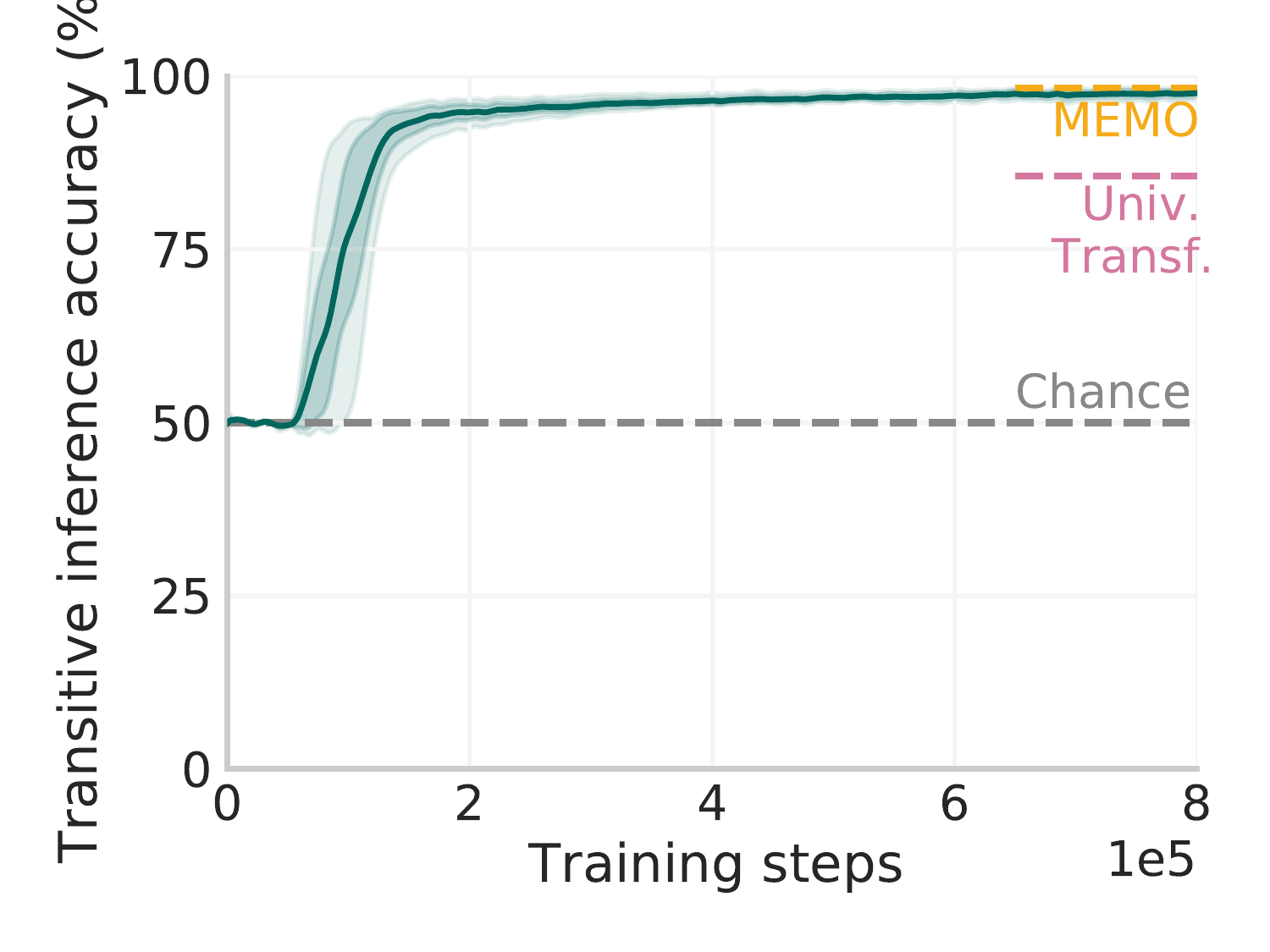}
    \vskip-0.5em
    \caption{Paired Associative Inference.}
    \label{fig:exp:other_papers:pai}
    \end{subfigure}%
    \begin{subfigure}[b]{0.33\textwidth}
    \includegraphics[width=\textwidth]{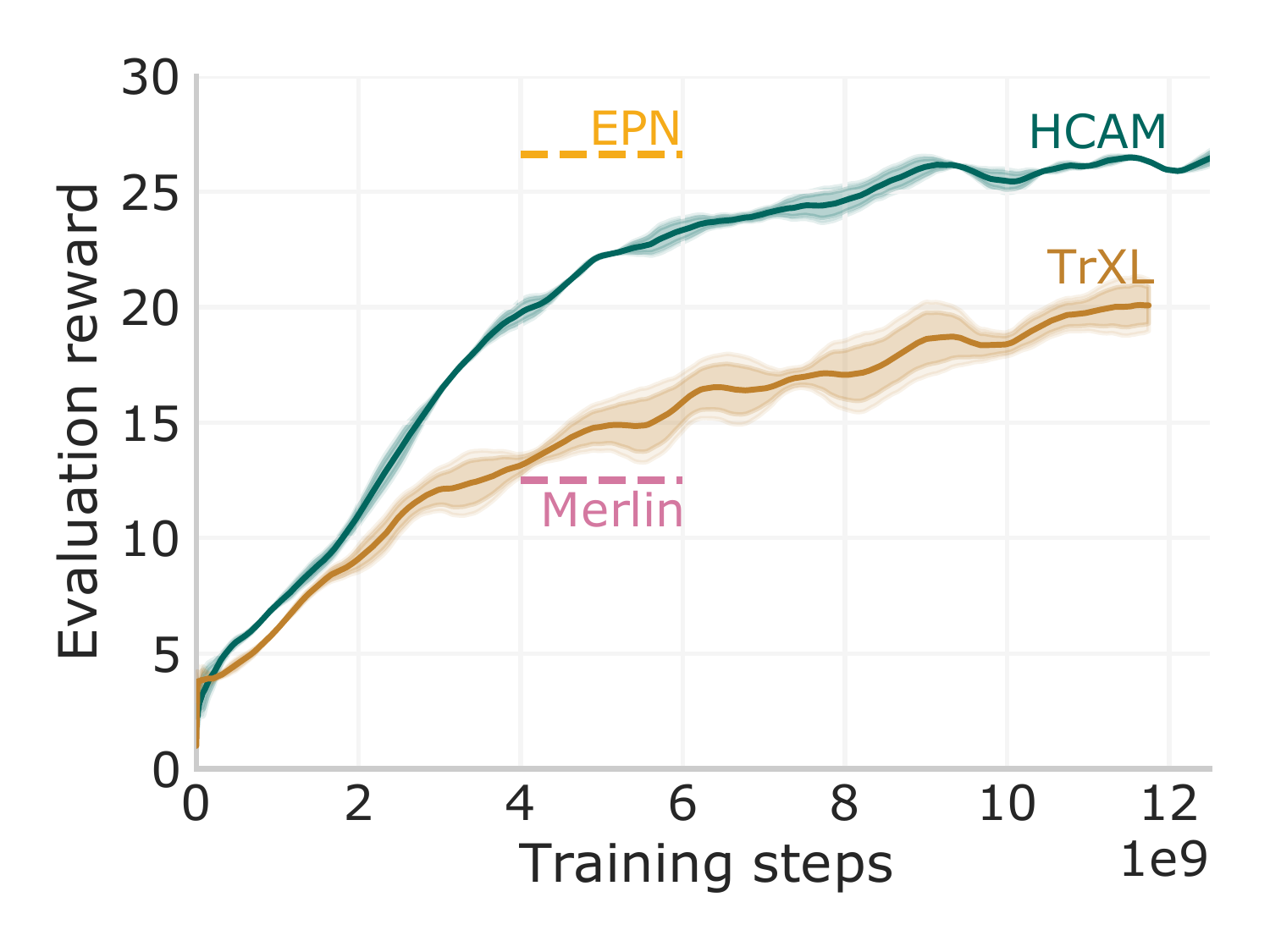}
    \vskip-0.5em
    \caption{One-Shot StreetLearn.}
    \label{fig:exp:other_papers:ossl}
    \end{subfigure}%
    \caption{Evaluating HCAM on tasks from other memory papers. Our agent achieves performance competitive with the specialized model proposed to solve each task. Top results (gold) and selected baselines (red) from each paper are indicated. (\subref{fig:exp:other_papers:tvt}) The Passive Visual Match task \cite{hung2019optimizing}.  (\subref{fig:exp:other_papers:tvt}) The Paired Associative Inference task \citep{banino2020memo}. HCAM achieves near optimal performance, comparable to MEMO and above Universal Transformers \citep{dehghani2018universal}. (\subref{fig:exp:other_papers:ossl}) The One-Shot StreetLearn environment \citep{ritter2020rapid}. HCAM achieves comparable performance to EPN on this difficult navigation task, although it is slower to learn. Both substantially outperform strong baselines. (Some prior results were estimated from figures. HCAM results aggregate over: (\subref{fig:exp:other_papers:tvt}) 6 seeds. (\subref{fig:exp:other_papers:pai}) 3 seeds, selected by validation accuracy. (\subref{fig:exp:other_papers:ossl}) 2 seeds.)}
    \label{fig:exp:other_papers}
    \vspace{-0.75em}
\end{figure}

\paragraph{Passive Visual Match} First, we compare to the Passive Visual Match task from \citet{hung2019optimizing}. In this task, the agent must remember a color it sees at the beginning of the episode, in order to choose a matching color at the end (Fig.\ \ref{fig:exp:task_overview:tvt}).  HCAM achieves reliably accurate performance (Fig.\ \ref{fig:exp:other_papers:tvt}), comparable to RMA, the best model from \citet{hung2019optimizing}. It would be interesting to combine HCAM with the value transport idea Hung et al. proposed, in order to solve their active tasks. \vspace{-0.5em}

\paragraph{Paired Associative Inference}
We next considered the Paired Associative Inference task \citep{banino2020memo}. In each episode, the agent receives a set of pairings of stimuli, and it must chain together transitive inferences over these pairings in order to respond to a probe (Fig.\ \ref{fig:exp:task_overview:pai}). These tasks therefore require sequential reasoning over multiple memories. HCAM achieves comparable performance to MEMO on the length 3 version of this task, and substantially outperforms baselines like Universal Transformer \citep{dehghani2018universal}. This is likely because HCAM, like MEMO, respects the chunk structure of experiencing pairs of images.\vspace{-0.5em}

\paragraph{One-Shot StreetLearn}
Finally, we evaluated HCAM on One-Shot StreetLearn (Fig.\ \ref{fig:exp:task_overview:ossl}) proposed by \citet{ritter2020rapid}, based on \citet{mirowski2019streetlearn}. One-shot StreetLearn is a challenging meta-learning setting. The agent is placed in a neighborhood, and must navigate to as many goals as it can within a fixed time. The agent receives its current state and goal as visual inputs (Google StreetView images) and must learn about the neighborhood in early tasks within an episode, in order to navigate efficiently later. Ritter et al. designed a specialized Episodic Planning Network (EPN) to solve these tasks. HCAM---despite its more general purpose architecture---can achieve comparable performance to EPN (although HCAM is somewhat slower to reach this level of performance). Ritter et al. showed that EPN achieves near-optimal planning in later parts of each episode, and therefore \textbf{HCAM must be planning close to optimally to match EPN's performance.} Strong baselines perform much worse. Merlin \citep{wayne2018unsupervised}---a strong agent with a learned episodic key-value memory and auxiliary unsupervised training---only achieves around half the performance of HCAM and EPN. TrXL is also less effective. This highlights the value of more sophisticated memory architectures, and in particular emphasizes the value of sequences stored in memory, rather than single vectors: EPN stores transitions (sequences of length 2) and HCAM stores longer sequences, while Merlin has only single vector values for each key (as does TrXL).

\section{Discussion}

In this paper we have proposed the Hierarchical Chunk Attention Memory. HCAM allows agents to attend in detail to past events without attending to irrelevant intervening information. Thus, HCAM effectively implements a form of ``mental time-travel'' \citep{tulving1985memory}. Our main insight is that this ability can be achieved using a hierarchically structured memory that sparsely distributes detailed attention. Specifically, recall uses attention over memory chunk summaries to allocate more detailed attention within only the most relevant chunks. We see this as a potentially important insight, because both mental time travel \citep{suddendorf2009mental} and hierarchical structure \citep{hasson2015hierarchical} are essential to the power of human memory. Furthermore, recent work has suggested that human memory behaves more like a ``quantum wave function'' distribution over several past events \citep{manning2021episodic}, which matches our approach of a distribution over the top-\(k\) relevant chunks. HCAM incorporates a number of important features of human memory.

Correspondingly, we showed that agents with HCAM succeed at a wide range of environments and tasks that humans depend on memory to solve. HCAM allows agents to remember sequential events, such as a ballet, in detail. HCAM allows agents to rapidly learn to navigate near-optimally in a new neighborhood by planning over previous paths; to perform transitive inferences; and to maintain object permanence, despite delays, occlusion and looking away. HCAM allows agents to effectively learn new words from a single exposure, and maintain that knowledge across distractor tasks. The agents can even extrapolate far beyond the training distribution to recall words after subsequent learning episodes \textbf{without ever being trained to do so}---better memory systems can help with the challenge of bridging from meta-learning to continual learning, which is receiving increasing interest \citep{finn2019online,he2019task,ritter2020rapid,caccia2020online,moskvichev2021updater}.

The tasks we considered are challenging because they rely on flexible use of memory. The agent did not know in advance which paths to remember in One-Shot StreetLearn, or which dancer would be chosen in Ballet. Furthermore, we trained the agent simultaneously on procedurally generated episodes where task features varied---e.g. the agent did not know how many distractor phases would appear in an episode, or the length of delays, so it could not rely on a fixed policy that recalls information after a fixed time. HCAM might be especially useful in these difficult, variable settings.

HCAM is robust to hyperparameters such as chunk size (App.\ \ref{app:supp_exp:varying_chunk_size}) and the number \(k\) of memories selected at each layer and step (App.\ \ref{app:supp_exp:varying_k})---it is even able to solve many tasks with \(k=1\). It also outperforms TrXL models that are either twice as wide or twice as deep, and correspondingly have many more parameters and, in the deeper case, make many more attention computations (App.\ \ref{app:supp_exp:matched_params}). Our comparisons to other approaches (Sec. \ref{sec:exp:other}) show that our approach is competitive with state of the art, problem-specific memory architectures, and outperforms strong baselines like Merlin and Universal Transformers. Finally, in hyperparameter sweeps suggested by our reviewers to improve TrXL's performance, we found that HCAM was consistently more robust to hyperparameter variation (App.\ \ref{app:supp_exp:response}). These observations suggest that our results should be relatively generalizable.\vspace{-0.5em}

\paragraph{Self-supervised learning} As in prior memory work \citep{wayne2018unsupervised,fortunato2019generalization,hill2020grounded} we found that self-supervised learning was necessary to train the agent to store all task-relevant information in memory. Training the agent to reconstruct its input observations as outputs was sufficient in the tasks we explored, but more sophisticated forms of auxiliary learning \citep{jaderberg2016reinforcement,banino2021coberl} might be useful in other settings. \vspace{-0.5em}

\paragraph{Memory in RL vs. supervised settings} Our work has focused on improving the memory of a situated RL agent. Compared to supervised settings such as language modelling, where a vanilla transformer can achieve very impressive performance \citep{raffel2020exploring,brown2020language}, RL poses unique challenges to memory architectures. First, sparse rewards present a more impoverished learning signal than settings that provide detailed errors for each output. The ability of HCAM to restore a memory in detail may help ameliorate this problem. Second, the multimodal (language + vision) stimuli experienced by our agent contains much more information per step than the word-pieces alone that a language model experiences, and therefore our setting may place more demands on the memory architecture. Finally, our tasks require access to detailed, structural aspects of past memories, while many existing NLP tasks do not depend significantly on structure --- for example, \citet{pham2020out} show that BERT ignores word order when solving many language understanding benchmarks, and produces equivalent outputs for shuffled inputs. Detailed memory will be most beneficial in settings in which the sequential structure of the past is important.

Nonetheless, we do not rule out possible applications of HCAM in supervised settings, such as language processing \citep{raffel2020exploring,brown2020language}, video understanding \citep{sun2019learning}, or multimodal perception \citep{jaegle2021perceiver}. HCAM would likely be most beneficial in tasks that strongly rely on long-term context and structure, e.g.\ the full-length version of NarrativeQA \citep{kovcisky2018narrativeqa}. Because videos often contain hierarchically-structured events, video models might especially benefit from HCAM-style attention. More broadly, the greater efficiency of sparse, hierarchical attention might allow detailed memory even in settings with limited computational resources, such as embedded systems.\vspace{-0.5em}

\paragraph{Episodic memories} \citet{nematzadeh2020memory} suggested endowing transformers with external episodic memories. Several recent papers have proposed systems that can be interpreted as episodic memories from this perspective, specifically looking up related contexts from the training corpus using either nearest neighbors \citep{khandelwal2019generalization}, or attention \citep{yogatama2021adaptive}. However, these approaches have generally only used this type of external memory to predict outputs, rather than allowing the model to perform further computations over the memories it recalls, as in HCAM. Thus, these memories would be inadequate for most RL tasks, which require planning or reasoning with memories.

Various works have proposed other types of external/episodic memory, both in RL specifically \citep[e.g.][]{wayne2018unsupervised,hung2019optimizing,fortunato2019generalization} and in deep learning more broadly \citep[e.g.][]{santoro2016meta, graves2016hybrid}. These approaches have generally not stored memories with the hierarchical summary-chunk structure of HCAM. \citet{ritter2020rapid} proposed an episodic memory for navigation tasks that stores state transitions; this architecture can be seen as a step towards our approach of storing sequences as chunks. Indeed, in the challenging city navigation domain that Ritter et al. explored, we showed that HCAM achieves comparable performance to their EPN architecture, despite HCAM being much more general. Furthermore, both HCAM and EPN substantially outperform Merlin \citep{wayne2018unsupervised}, a strong baseline that learns to store vector memories, rather than the rich representations of sequences stored by HCAM (or transitions stored by EPN). This highlights the value of rich, sequential memories. We suggest that episodic memories could benefit from moving beyond the idea of keys and values as single vectors. \textbf{It can be useful to store more general structures---such as a time-sequence of states---as a single ``value'' in memory.}

One benefit of episodic memory is that memory replay can support learning \citep{kumaran2016learning, mcclelland1995there}, in particular by ameliorating catastrophic interference. It would therefore be interesting to explore whether HCAM's memory could be used for training. For example, could memory summaries be used to locate similar or contrasting memories that should be interleaved together \citep{mcclelland2020integration} to improve continual learning?\vspace{-0.5em}

\paragraph{Transformer memories} 
Many recent works have attempted to improve the computational efficiency of transformer attention over long sequences \citep[e.g.][]{kitaev2020reformer,wang2020linformer}. However, these approaches often perform poorly at even moderately long tasks \citep{tay2021long}. Similarly, we found that on tasks like Ballet or One-Shot StreetLearn, Transformer-XL can perform suboptimally even when the entire task fits within a single, relatively short attention window. However, there could potentially be complementary benefits to combining approaches to obtain even more effective attention with hierarchical memory, which should be investigated in future work. In particular, some recent work has proposed a simple inductive bias which allows Transformers to benefit from evaluation lengths much longer than they were trained on \citep{press2021train}---while this strategy would likely not help with recalling specific instances in detail, it might be complementary to an approach like ours.

Other works have used a hierarchy of transformers with different timescales for supervised tasks \citep{liu-lapata-2019-hierarchical,zhang2019hibert,yang2020html,li2020isarstep}, for example encoding sentences with one transformer, and then using these embeddings as higher-level tokens. This approach improves performance on tasks like document summarization, thus supporting the value of hierarchy. \citet{luong2015effective} also showed benefits of both local and global attention computation in LSTM-based models. However, we are not aware of prior transformers in which the coarse computations are used to select chunks for more detailed computation, as in HCAM (although recent work has demonstrated top-\(k\) patch selection in CNNs \citep{cordonnier2021differentiable}); nor are we aware of prior works that have demonstrated their approaches in RL, or beyond a single task domain.\vspace{-0.5em}

\paragraph{Memory and adaptation} One major benefit of memory is that a model can flexibly use its memories in a goal- or context-dependent way in order to adapt to new tasks \citep{tulving1985memory,suddendorf2009mental,kumaran2016learning}. Adult humans use our recall of rich, contextual memories in order to generalize effectively \citep{suddendorf2009mental,NGO2021}. While our tasks required some forms of goal-dependent memory use---e.g.\ combining old paths to plan new ones---it would be interesting to evaluate more drastic forms of adaptation. Recent work shows that transforming prior task representations allows zero-shot adaptation to substantially altered tasks \citep{lampinen2020transforming}. HCAM could potentially learn to transform and combine memories of prior tasks in order to adapt to radically different settings. Exploring these possibilities offers an exciting future direction.\vspace{-0.5em}

\paragraph{Limitations \& future directions} \label{sec:discussion:limitations}

While we see our contribution as a step towards more effective mental time-travel for agent memory, many aspects could be further improved. Our implementation requires the ability to keep each previous step in (hardware) memory, even if the chunk containing that step is irrelevant. This approach would be challenging to scale to the entire lifetime of episodic memory that humans retain. Nonetheless, storing the past is feasible up to tens of thousands of steps on current accelerators. This could be extended further by using efficient \(k\)NN implementations, which have recently been used to perform \(k\)NN lookup over an entire language dataset \citep{khandelwal2019generalization}.

The challenge of scaling to longer-term memory would be helped by deciding what to store, as in some episodic memory models \citep{graves2016hybrid}, soft forgetting of rarely retrieved memories, and similar mechanisms. However, such mechanisms are simultaneously limiting, in that the model may be unable to recall knowledge that was not obviously useful at the time of encoding. Over longer timescales, HCAM might also benefit from including more layers of hierarchy---grouping memory chunks into higher-order chunks recursively to achieve logarithmic complexity for recall. Even over moderate timescales, more intelligent segmentation of memory into chunks would potentially be beneficial---human encoding and recall depends upon complex strategies for event segmentation \citep{zacks2007event,sols2017event,chan2017lingering,lu2020learning,chang2021relating}. HCAM might also benefit from improved chunk summaries; mean-pooling over the chunk is a relatively na\"{i}ve approach to summarization, so learning a compression mechanism might be beneficial \citep{rae2019compressive}. These possibilities provide exciting directions for future work. \vspace{-0.5em}


\paragraph{Conclusions}

We have proposed a Hierarchical Chunk Attention Memory for RL agents. This architecture allows agents to recall their most relevant memories in detail, and to reason over those memories to achieve new goals. This approach outperforms (or matches the optimal performance of) a wide variety of baselines, across a wide variety of task domains. It allows agents to remember where objects were hidden, and to efficiently learn to navigate in a new neighborhood by planning from memory. It allows agents to recall words they have learned despite distracting intervening tasks, and even across episodes. These abilities to learn, adapt, and maintain new knowledge are critical to intelligent behavior, especially as the field progresses towards more complex environments. We hope that our work will motivate further exploration of hierarchically structured memories, in RL and beyond. Hierarchical memory may have many benefits for learning, reasoning, adaptation, and intermediate- or long-term recall.  

\begin{ack}
We would like to acknowledge Adam Santoro, Emilio Parisotto, Jay McClelland, David Raposo, Wilka Carvalho, Tim Scholtes, Tamara von Glehn, S\'{e}bastien Racani\`{e}re, and the anonymous reviewers for helpful comments and suggestions.
\end{ack}

\bibliographystyle{plainnat}
\bibliography{htm}

\appendix

\section{Methods} 

\subsection{Open source attention module} \label{app:methods:opensource}

We have open-sourced a Jax/Haiku implementation of our Hierarchical Attention Module, which can be found at: \url{https://github.com/deepmind/deepmind-research/tree/master/hierarchical_transformer_memory/hierarchical_attention}

We have also released our Ballet and Rapid Word Learning environments, and include links to those and the released versions of other environments we used below. 

\subsection{Agent architecture and training} \label{app:methods:agents}

\begin{table}[H]
    \centering
    \caption{Hyperparameters used in main experiments. Where only one value is listed across multiple columns, it applies to all. Where hyperparameters were swept, parameters used for HCAM and TrXL are reported separately.}
    \resizebox{\textwidth}{!}{
    \begin{tabular}{|c|c|c|c|c|c|c|}
      \hline
      & \textbf{Ballet} & \textbf{Objects} & \textbf{Words} & \textbf{Image} & \textbf{Streets} & \textbf{Associative}  \\ \hline
      \hline
      All activation fns & \multicolumn{6}{c|}{ReLU} \\ \hline
      State dimension & \multicolumn{6}{c|}{512} \\ \hline
      Memory dimension & \multicolumn{6}{c|}{512} \\ \hline
      Memory layers & \multicolumn{6}{c|}{4} \\ \hline
      Memory num. heads & \multicolumn{6}{c|}{8} \\ \hline
      HCAM chunk size & 32 & 16 & 16 & 16 & 4 & 2 \\ \hline
      HCAM chunk overlap & \multicolumn{4}{c|}{-} & 1 & -\\ \hline
      HCAM top-\(k\) & 8 & 8 & 16 & 16 & 32 & 32 \\ \hline
      HCAM total num. chunks & \multicolumn{6}{c|}{always greater than max episode length // chunk size} \\ \hline
      TrXL extra length & 256 & 512 & 256 & NA & 200 (full) & NA \\ \hline
      \hline
      Visual encoder & CNN & \multicolumn{5}{c|}{ResNet}\\ \hline 
      Vis. enc. channels & \multicolumn{6}{c|}{(16, 32, 32)} \\ \hline 
      Vis. enc. filt. size & (9, 3, 3) & \multicolumn{5}{c|}{(3, 3, 3)}\\ \hline 
      Vis. enc. filt. stride & (9, 1, 1) & \multicolumn{5}{c|}{(2, 2, 2)}\\ \hline 
      Vis. enc. num. blocks & NA & \multicolumn{5}{c|}{(2, 2, 2)}\\ \hline 
      \hline
      Language encoder & \multicolumn{3}{c|}{1-layer LSTM} & \multicolumn{3}{c|}{NA} \\ \hline
      Lang. enc. dimension & \multicolumn{3}{c|}{256} & \multicolumn{3}{c|}{NA} \\ \hline
      Word embed. dimension & \multicolumn{3}{c|}{32} & \multicolumn{3}{c|}{NA} \\ \hline
      \hline
      Policy \& value nets & \multicolumn{6}{c|}{MLP with 1 hidden layer with 512 units.} \\ \hline
      Reconstruction decoders & \multicolumn{6}{c|}{Architectural transposes of the encoders, with independent weights.} \\ \hline
      \hline
      Recon. loss weight (HCAM) & \multicolumn{2}{c|}{1.} & 0.3 & 1. & 0.3 & NA \\ \hline
      Recon. loss weight TrXL & \multicolumn{3}{c|}{1.} & NA & 1. & NA\\ \hline
      \(V\)-trace loss weight (HCAM) & \multicolumn{2}{c|}{0.1} & 0.3 & 0.1 & 1. & NA \\ \hline
      \(V\)-trace loss weight (TrXL) & 0.3 & \multicolumn{2}{c|}{0.1} & NA & 1. & NA \\ \hline
      \(V\)-trace baseline weight (HCAM) & \multicolumn{2}{c|}{1.} & 0.3 & 1. & 0.3  & NA\\ \hline
      \(V\)-trace baseline weight (TrXL) & 1. & 0.3 & 1. & NA & 1.  & NA\\ \hline
      Entropy weight & \(1\cdot10^{-3}\) & \multicolumn{4}{c|}{\(1\cdot10^{-4}\)} & NA\\ \hline
      \hline
      Batch size & \multicolumn{6}{c|}{32} \\ \hline
      Training unroll length & 64 & 128 & 128 & 128 & 64 & NA \\ \hline
      \hline 
      Optimizer & \multicolumn{6}{c|}{Adam \citep{kingma2014adam}} \\ \hline
      LR & \multicolumn{4}{c|}{\(2\cdot10^{-4}\)} & \(4\cdot10^{-4}\) & \(1\cdot10^{-4}\) \\ \hline
    \end{tabular}}
    \label{tab:app:hyper}
\end{table}

\begin{table}[H]
    \centering
    \caption{Hyperparameter sweeps used in main experiments. The One-Shot StreetLearn and Paired Associative Inference tasks used different chunk size sweeps due to the different task lengths and demands. The Paired Associative Inference tasks do not use RL losses and so did not use the corresponding loss sweeps.} \label{tab:app:hypersweep}
    \begin{tabular}{|c|c|}
      \hline
      Reconstruct. loss weight & \{0.1, 0.3, 1.\}\\
      \(V\)-trace loss weight &  \{0.1, 0.3, 1.\}\\
      \(V\)-trace baseline weight &   \{0.1, 0.3, 1.\}\\
      HCAM chunk size & \{16, 32, 64\}, except Streets=(2, 4, 8, 16) and Associative=None\\
      LR & None, except Streets and Associative=\{1, 2, 4\}\(\cdot 10^{-4}\)\\
      \hline
    \end{tabular}
\end{table}
In Table \ref{tab:app:hyper} we show the hyperparameters used for all experiments, and in Table \ref{tab:app:hypersweep} we show the hyperparameters sweeps used, although we generally used a subset of the full sweep. We swept each hyperparameter with a single seed per condition, and then reran the best parameter settings for each condition with more seeds to get a more robust estimate of the performance of each approach.

In most cases the hyperparameters that were not swept were taken from other sources without tuning for our architecture. In particular, the first tasks we considered were the rapid word learning tasks, and many parameters were taken directly from the hyperparameters of the original paper on which the tasks were based \citep{hill2020grounded}. These hyperparameters were therefore tuned directly by prior researchers for other models, but we found them to work well for our memory as well. The visual encoder, language encoder, unsupervised reconstruction loss etc. were copied from those described in the prior work.

Since we ran all other experiments after the initial word learning experiments, we used many of these same hyperparameters on other experiments, such as the visual and language encoder architecture across all experiments. However, some hyperparameters do differ across tasks due to specific task features. For example, the visual encoder for the ballet tasks is set to have a filter size of 9 because this is the resolution of each square in the grid, and the entropy cost for the ballet tasks was chosen from our prior work \citep{hill2019environmental} which used a similar grid world action space. These decisions were shared across all architectures, so should not favor our model over the baselines.

\paragraph{Self-supervised reconstruction loss} We used the same reconstruction loss as \citet{hill2020grounded}, namely reconstructing the language with a softmax cross-entropy loss, and reconstructing the image pixels (normalized to range [0, 1] on each color channel) with a sigmoid cross-entropy loss. The image reconstruction loss was averaged across all pixels and channels, while the language reconstruction loss was summed across the sequence. 

\subsubsection{Bug fixed between original and revised versions of this paper} \label{app:methods:bug_fixed}

Shortly before the camera-ready deadline for NeurIPS, we discovered a bug in the configuration of the HCAM in the Ballet, Words, and Passive Visual Match domains: the local attention window was much longer than intended. Fixing this bug did not substantially alter results in the Ballet or Passive Visual Match tasks, but did change our results somewhat in the Rapid Word Learning tasks. The qualitative patterns of extrapolation and generalization to multiple episodes remain the same, but generalization of HCAM is somewhat worse, although still much better than the baseline models. This does not substantially affect the conclusions of the paper. We have revised the Rapid Word Learning plots in the main text to reflect these updated results, and included evaluation on the original levels in Fig. \ref{fig:supp_exp:fastbind_harder_holdouts}. However, note that our supplemental analyses in this domain were carried out with the longer attention window.

\subsection{Plotting methods}
In all plots, each curve is an average across multiple runs. The \(x\)-axis is always the number of agent steps (actions taken/frames seen) during training. The number of learner updates is generally \(2000-4000\times\) smaller with a batch size of 32 trajectories per update, and unroll lengths of 64-128. The dark regions around the curve show \(\pm\)SD across runs, the light regions show the total range. The plots are smoothed by interpolation with a triangular window, with width and sampling frequency chosen to present results clearly depending on the speed and variability of learning the different tasks. All figures were made with seaborn \citep{Waskom2021} and matplotlib \citep{Hunter2007}.

\subsection{Compute resources}
All experiments were run using Google TPU 
\ifanonymized
  v2 and v3
\else
  v2, v3, and v4
\fi
devices. Each run lasted between a few hours and a few days depending on the experiment. We ran actors/evaluators on CPUs. We estimate the total time needed to reproduce all experiments (including baselines and experiments in appendices) to be around 1000 TPU-hours + 300000 CPU hours.

\section{Tasks} \label{app:tasks}

We have uploaded selected video recordings of the HCAM-based agent performing our main tasks 
\ifanonymized
in the supplemental materials for the paper.
\else
at \url{https://www.youtube.com/playlist?list=PLE5lx5-YU_Hr8Q9IgTAfisJ6XCy3Jhh6F}
\fi

\subsection{Open source or released tasks}

We have open-sourced our Ballet environment at: \url{https://github.com/deepmind/deepmind-research/tree/master/hierarchical_transformer_memory/hierarchical_attention}

We have released our rapid-word-learning tasks in the repository for the paper they were based upon \url{https://github.com/deepmind/dm_fast_mapping}

The environments from other papers that we used also have corresponding releases:
\begin{enumerate}
    \item Passive Visual Match: \url{https://github.com/deepmind/deepmind-research/tree/master/tvt}
    \item Paired Associative Inference: \url{https://github.com/deepmind/deepmind-research/tree/master/memo}
    \item One-Shot StreetLearn \url{https://github.com/deepmind/deepmind-research/tree/master/rapid_task_solving}
\end{enumerate}

\subsection{Ballet} \label{app:tasks:ballet}
The tasks took place in a \(9 \times 9\) tile room with an extra 1 tile wall surrounding on all sides, for a total of \(11 \times 11\) tiles. This was upsampled at a resolution of 9 pixels per tile to form a \(99 \times 99\) image as input to the agent. The agent was placed in the center of the room, and the dancers were placed randomly in 8 possible locations around it. The dancers always had distinct colors and shapes, selected from 15 shapes and 19 colors. These features merely served to distinguish the dancers. The agent always appeared as a white square. The agent received egocentric inputs (that is, its visual input was centered on its location), as this can improve generalization \citep{hill2019environmental}. 

In Listing \ref{lst:app:tasks:ballet:dances} we show the dance sequences used for the ballet tasks. All dances are 16 steps long. We trained all agents with levels uniformly sampled to have 16 or 48 steps of delay between dances, and 2, 4, or 8 dances. The number of dancers in the room corresponded to the number of dances, such that if there were only 2 dances, there were only 2 dancers, while if there were 8 dances there were 8 dancers. This is why chance-level performance is 50\% with 2 dances, but 12.5\% with 8. The agent was given a reward of 1 for a correct choice, and 0 for an incorrect choice.
\begin{lstfloat}
\begin{lstlisting}
{
"circle_cw": [0, 2, 4, 4, 6, 6, 0, 0, 2, 2, 4, 4, 6, 6, 0, 2],
"circle_ccw": [0, 6, 4, 4, 2, 2, 0, 0, 6, 6, 4, 4, 2, 2, 0, 6],
"up_down": [0, 4, 4, 0, 0, 4, 4, 0, 0, 4, 4, 0, 0, 4, 4, 0],
"left_right": [2, 6, 6, 2, 2, 6, 6, 2, 2, 6, 6, 2, 2, 6, 6, 2],
"diagonal_uldr": [7, 3, 3, 7, 7, 3, 3, 7, 7, 3, 3, 7, 7, 3, 3, 7],
"diagonal_urdl": [1, 5, 5, 1, 1, 5, 5, 1, 1, 5, 5, 1, 1, 5, 5, 1],
"plus_cw": [0, 4, 2, 6, 4, 0, 6, 2, 0, 4, 2, 6, 4, 0, 6, 2],
"plus_ccw": [0, 4, 6, 2, 4, 0, 2, 6, 0, 4, 6, 2, 4, 0, 2, 6],
"times_cw": [1, 5, 3, 7, 5, 1, 7, 3, 1, 5, 3, 7, 5, 1, 7, 3],
"times_ccw": [7, 3, 5, 1, 3, 7, 1, 5, 7, 3, 5, 1, 3, 7, 1, 5],
"zee": [1, 6, 6, 2, 2, 5, 1, 5, 5, 2, 2, 6, 6, 1, 5, 1],
"chevron_down": [7, 4, 3, 1, 0, 5, 1, 5, 1, 4, 5, 7, 0, 3, 7, 3],
"chevron_up": [3, 0, 7, 5, 4, 1, 5, 1, 5, 0, 1, 3, 4, 7, 3, 7],
}
\end{lstlisting}
\caption{Dances used in the ballet task. 0-7 refer to directions of movement, clockwise from 0 = up.}
\label{lst:app:tasks:ballet:dances}
\end{lstfloat}

\subsection{Object permanence}
The tasks took place within a 3D environment created with Unity. The agent received a visual observation of \(96 \times 72 \times 3\) pixel RGB images, and a language observation that was tokenized at the word-level. The agent was initially placed in a fixed position near one wall of the room facing toward the center, and the boxes were randomly placed within the agent's field of view. When each object appeared, it jumped out of its box three times in succession. If there was a delay period, it began after the object returned to its box for the third time. The delay periods we used for the varying length training were 0, 10, 20, and 30 seconds. After the last presentation and delay phase, the lids of the boxes closed.

After the lids of the boxes closed, the agent was allowed to move and look around, and was given the instruction ``look backward.'' The agent was rewarded 0.3 for looking backwards (far enough that the boxes were out of view), and looking backward allowed it to advance to the choice phase of the task. In the choice phase, the agent was told ``go to the box containing the [duck]'' and was rewarded 1 for making the correct choice, and 0 for an incorrect choice.

\subsection{Rapid word learning with distractors}
We used the tasks created by \citet{hill2020grounded} with the following modifications. First, we removed three of the possible objects (trains, robots, and rockets) to be used in the distractor task. We then added 0-20 distractor phases between the word binding and test phases. In each distractor phase, the agent and the three distractor objects were randomly placed in the room, and the agent was asked to lift one of them, e.g. ``lift the rocket.'' The agent received a reward of 0.1 for successfully lifting the right object, and was allowed to progress to the next distractor task. If the agent lifted the wrong object, it was neither rewarded nor allowed to progress until it had lifted the correct object or until 20 seconds had passed. All agents rapidly learned to solve these distractor tasks. A fixed time limit of 450 seconds was used across all episodes, after which the episode terminated with reward 0 regardless of what phase the agent was in. 

For the multi-episode evaluation tasks, we simply combined the number of episodes we wished to test across into a single ``super-episode.'' For the final test phase, where we tested earlier words, the distractor objects were always taken from the same learning phase as the target object (to ensure that the agent remembered the exact name-object pairings, rather than simply which name appeared with which group of objects). We set a time limit of 450 seconds to complete \emph{all} the sub-episodes. Agents with HCAM and TrXL memories were able to consistently complete the super-episodes within this time limit---even though TrXL could not choose the correct objects, it was consistently reaching the end and choosing \emph{some} object for a chance at the final reward. However, the LSTM-based agents often timed out on these multi-episode evaluations.

\subsection{Comparisons to other papers} \label{app:tasks:other}

The Passive Visual Match and Paired Associative Inference tasks were used unmodified. The StreetLearn \citep{mirowski2019streetlearn} images and maps we used for the One-Shot StreetLearn were a more recent version than those used by \citet{ritter2020rapid}. Because the task difficulty is fixed through the sampling of neighborhoods from the larger city graph, this should not substantially alter the difficulty of the tasks. We received permission from an author on each paper to use their tasks. 

All three tasks from other papers have been released under Apache licenses, and the open source code can be found at:
\begin{itemize}
    \item Passive visual match: \url{https://github.com/deepmind/deepmind-research/tree/master/tvt/dmlab}
    \item Paired Associative Inference: \url{https://github.com/deepmind/deepmind-research/tree/master/memo}
    \item One-Shot StreetLearn: \url{https://github.com/deepmind/deepmind-research/tree/master/rapid_task_solving}.
\end{itemize}

\paragraph{The PAI task}
In order to apply HCAM to the supervised PAI task, we took the following steps. We embedded all the input memories and probes using a single shared embedding layer. The structure of the memories for the PAI task matches the structure of HCAM's contents, where each pair of associated images corresponds to a single chunk in memory (of length 2). We therefore created a HCAM-style memory containing these embedded contents, and keyed by their summaries (averages across each chunk). We then provided the embedded query as input to the multi-layer HCAM model, but used the same set of embedded task memories at every layer. After 4 HCAM layers, we averaged-pooled across the sequence of resulting embeddings, and then performed a linear projection to produce a final output embedding. We then compared this ouptut embedding to the embeddings of the two possible choices using dot products. These dot products were used as logits in a softmax to choose the answer, and the model was trained using a cross-entropy loss. We did not use a self-supervised reconstruction loss for this setting.

\section{Detailed results}

In Table \ref{tab:app:numericalresults} we show the mean performance and standard deviation across runs from our main experiments.

\begin{table}[htb]
    \centering
    \caption{Numerical results from main experiments/figures---mean \(\pm\) standard deviation across 3 runs per condition. Results are average performance (\% correct) across evaluations during the last 1\% of training, except for the One-Shot StreetLearn tasks, where they are average reward during the last 1\% of training. (Note that on some levels LSTMs were not consistently completing the task before the episode time limit. Incomplete episodes are scored as 0.)} \label{tab:app:numericalresults}
    \begin{tabular}{cccccc}
      Experiment & Level & Fig. & HCAM & TrXL & LSTM\\ \hline
      \multirow{3}{*}{Ballet} & 2 dances, delay 16 & \ref{fig:exp:ballet:2} & \(99.8 \pm 0.3\) & \(96.9 \pm 1.4\) & \(97.4 \pm 1.8\) \\
      & 8 dances, delay 16 &\ref{fig:exp:ballet:8} & \(98.1 \pm 3.3\)  & \(65.7 \pm 13.7\) & \(25.2\pm 2.7\) \\
      & 8 dances, delay 48 &\ref{fig:exp:ballet:8long} & \(97.2 \pm 2.5\) & \(49.2 \pm 13.0\) & \(29.7 \pm 9.4\)\\ \hline
      \multirow{3}{*}{Objects} & No delay, varying train & \ref{fig:exp:objects:curr_short} & \(96.7\pm 0.9\) & \(82.2 \pm 28.7\) & \(33.2 \pm 6.6\) \\
      & Long delay, varying train & \ref{fig:exp:objects:curr_long}  & \(91.7\pm 8.3\) & \(46.1 \pm 20.0\) & \(34.8 \pm 5.5\) \\
      & Long delay, long-only train & \ref{fig:exp:objects:longonly} & \(82.9\pm 16.4\) & \(31.1 \pm 0.6\) & - \\\hline
      \multirow{3}{*}{Words} & 10 distractors & \ref{fig:exp:fastbind:curr} & \(93.0 \pm 4.0\) & \(76.7 \pm 12.5\) & \(21.0 \pm 18.3\) \\
      & 4 episodes, 0 distractor each &  \ref{fig:exp:fastbind:across_eps_4_1}& \(82.8\pm 6.4\) & \(49.3 \pm 16.5\) & \(19.8 \pm 17.4\)  \\
      & 2 episodes, 5 distractors each &  \ref{fig:exp:fastbind:across_eps_2_5}& \(71.1\pm 8.0\) & \(48.7 \pm 13.2\) & \(22.9 \pm 20.1\)\\\hline
      Image & &  \ref{fig:exp:other_papers:tvt} & \(97.0\pm 2.8\) & - & - \\\hline
      Associative & &\ref{fig:exp:other_papers:pai}  & \(97.5\pm 0.9\) & - & - \\\hline
      Streets & & \ref{fig:exp:other_papers:ossl} & \(26.8\pm 0.44\) &  \(19.9\pm 0.65\) & - \\\hline
    \end{tabular}
\end{table}

\section{Supplemental experiments} \label{app:supp_exp}

In this section we present some supplemental experiments and analyses. However, we make several notes here. First, these supplemental analyses were mostly run with a longer local attention window than used in the main text, see App. \ref{app:methods:bug_fixed}, which could potentially affect results, particularly in the rapid word learning domain. Second, we use the original acronym HTM instead of HCAM in most of these plots, because we revised it only after a reviewer pointed out a name clash. 

\subsection{Fast-binding performance on harder hold-out tasks} \label{app:supp_exp:fastbind}
\begin{figure}[htb]
\centering
\begin{subfigure}{0.33\textwidth}
\includegraphics[width=\textwidth]{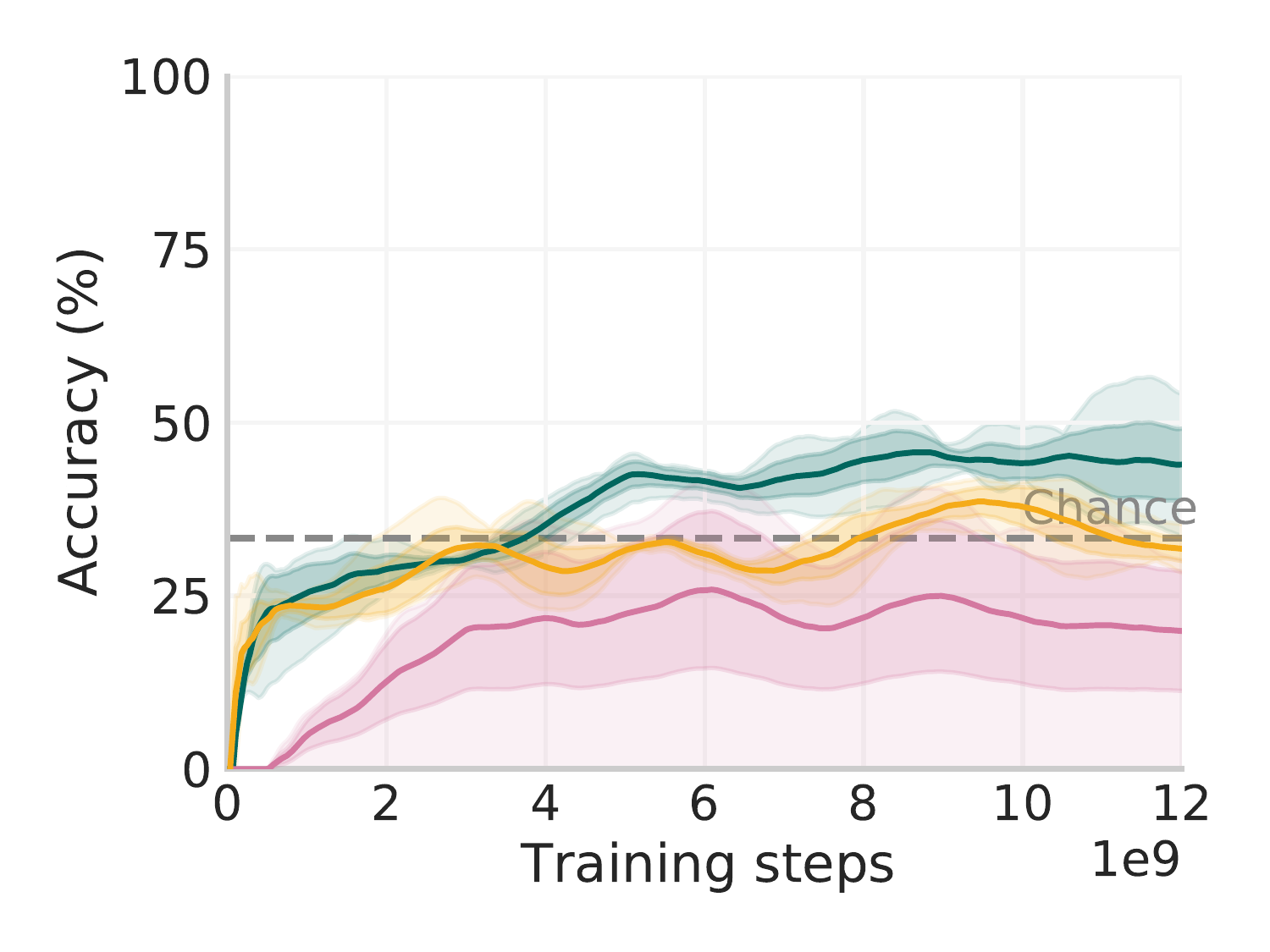}
\captionsetup{width=.8\textwidth}
\caption{Eval. 20 dist.} \label{fig:supp_exp:fastbind_harder:20}
\end{subfigure}%
\begin{subfigure}{0.33\textwidth}
\includegraphics[width=\textwidth]{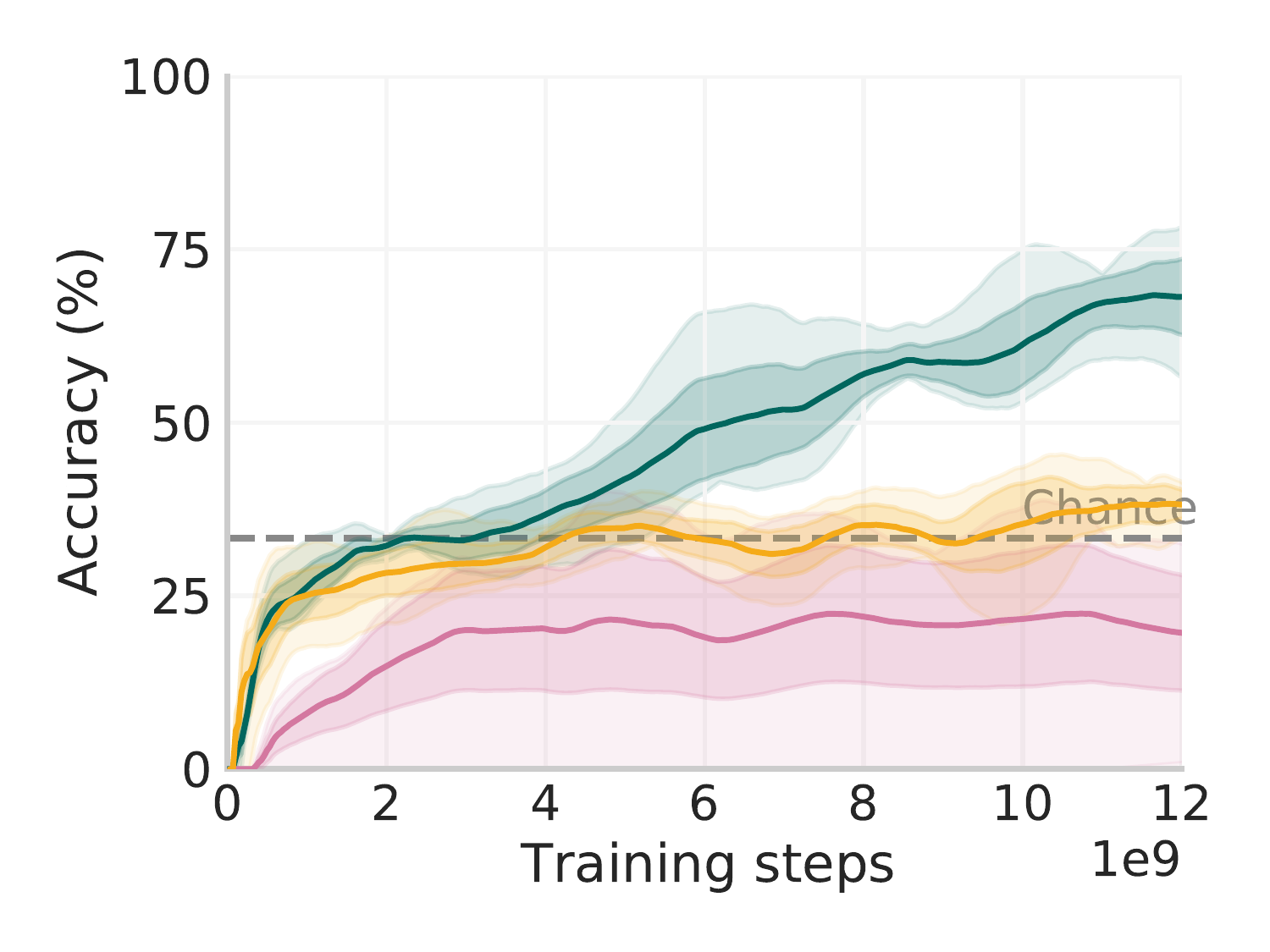}
\captionsetup{width=.8\textwidth}
\caption{Eval. 4 eps., 1 dist.} \label{fig:supp_exp:fastbind_harder:4_1}
\end{subfigure}%
\begin{subfigure}{0.33\textwidth}
\includegraphics[width=\textwidth]{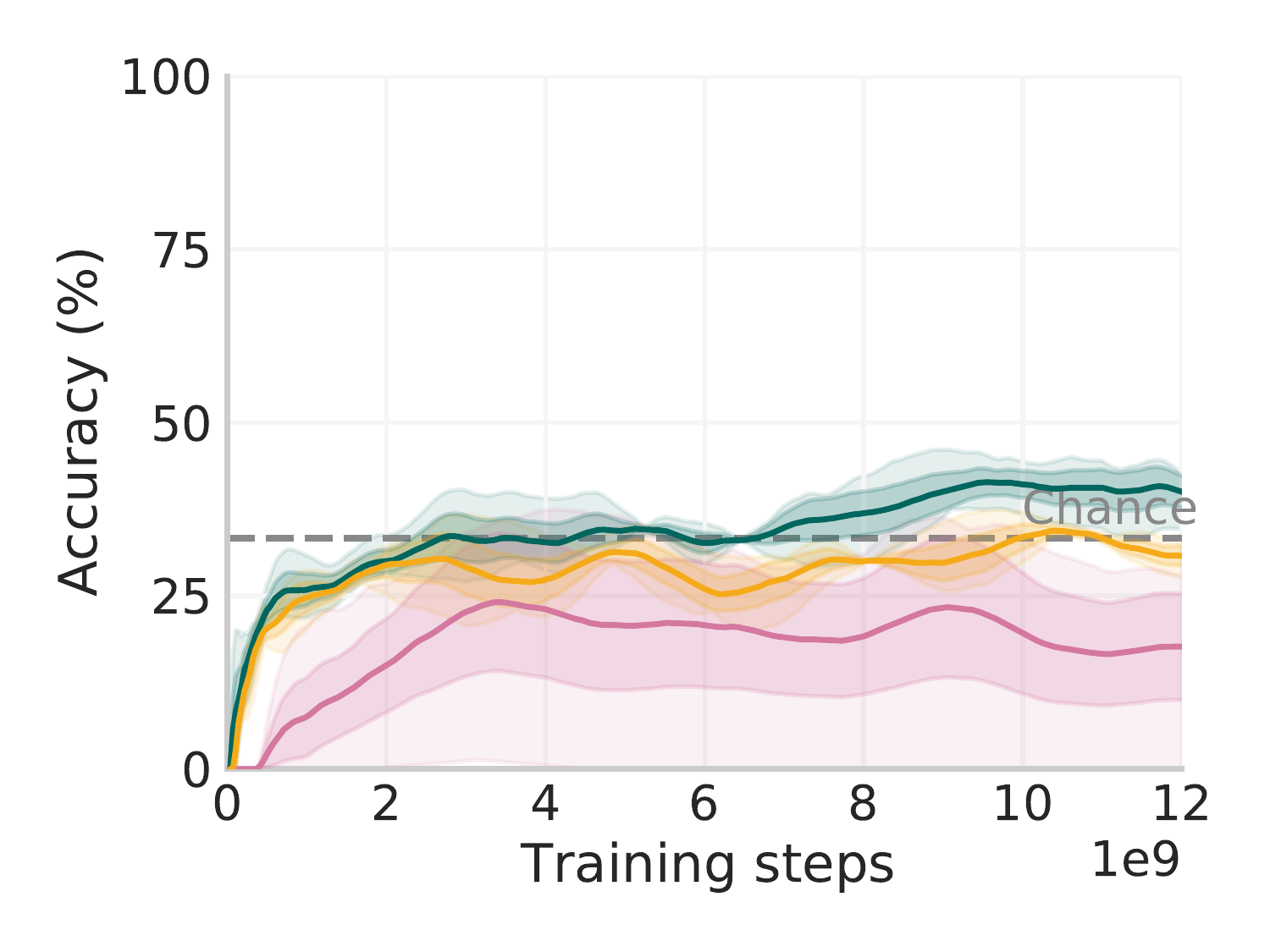}
\captionsetup{width=.8\textwidth}
\caption{Eval. 3 eps., 5 dist.} \label{fig:supp_exp:fastbind_harder:3_5}
\end{subfigure}%
\caption{Evaluating HCAM on the harder generalization tasks we considered in the original version of this paper, after longer training (note horizontal axis). HCAM achieves off-chance performance, and continues to improve as training goes on. (3 seeds per condition.)}
\label{fig:supp_exp:fastbind_harder_holdouts}
\end{figure}

In our original version of this paper, we presented generalization results on a harder set of evaluation tasks. Unfortunately, the high generalization performance on these results seemed to be at least in part due to a bug causing our HCAM memory to have a large local attention window (see above). We therefore changed the main text figures to show performance on slightly easier task variations. However, in Fig. \ref{fig:supp_exp:fastbind_harder_holdouts} we show performance on the original evaluation tasks. HCAM still achieves off-chance performance in 2 out of 3 cases, with fairly decent performance in one case, and performance continues to improve as training goes on.

\subsection{Ballet generalization}  \label{app:supp_exp:ballet_gen}
\begin{figure}[htb]
    \centering
    \begin{subfigure}{0.33\textwidth}
    \includegraphics[width=\textwidth]{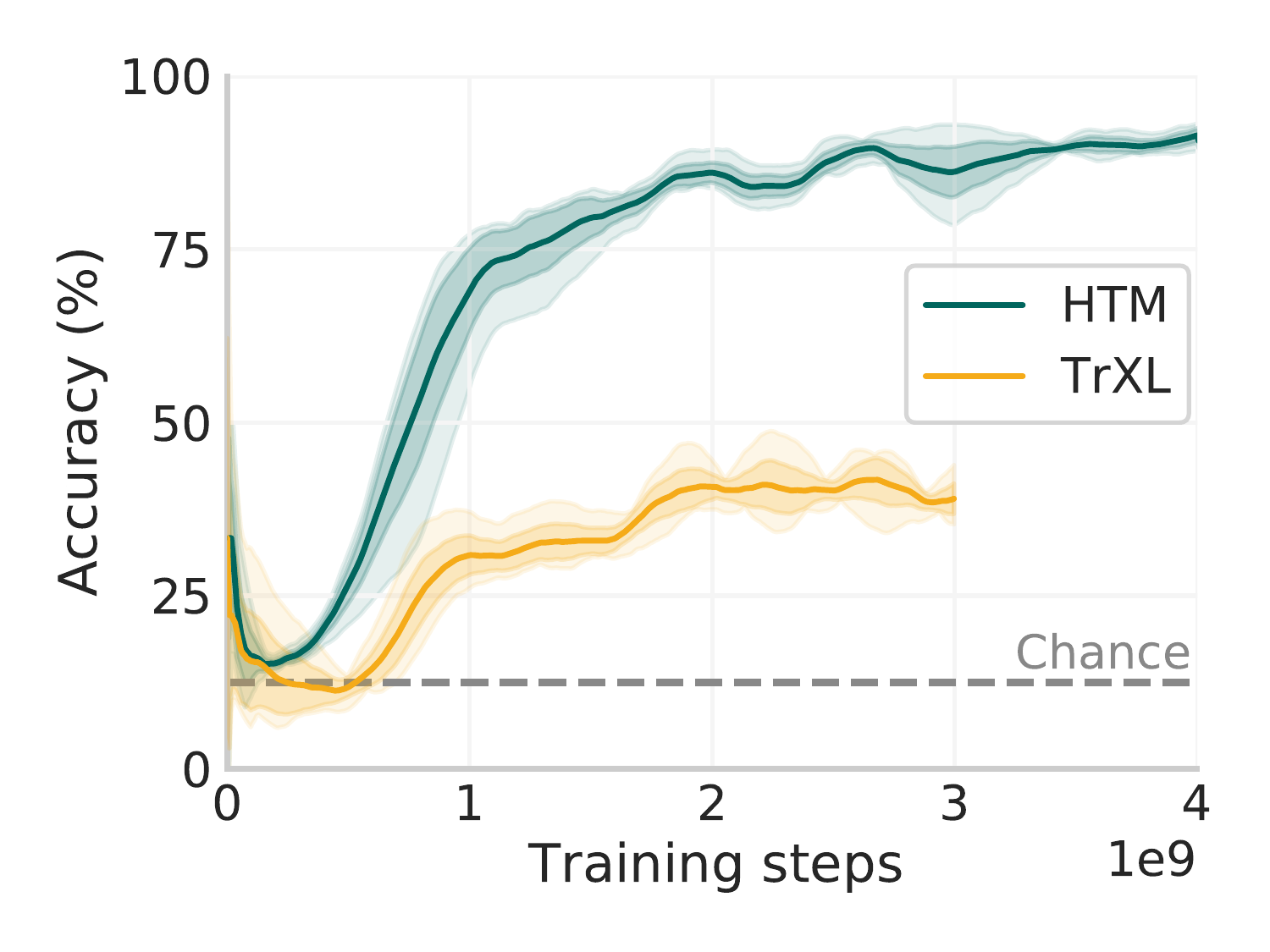}
    \captionsetup{width=.8\textwidth}
    \caption{Evaluate 8 dances, short delays.}
    \label{fig:supp_exp:ballet_gen:8}
    \end{subfigure}%
    \begin{subfigure}{0.33\textwidth}
    \includegraphics[width=\textwidth]{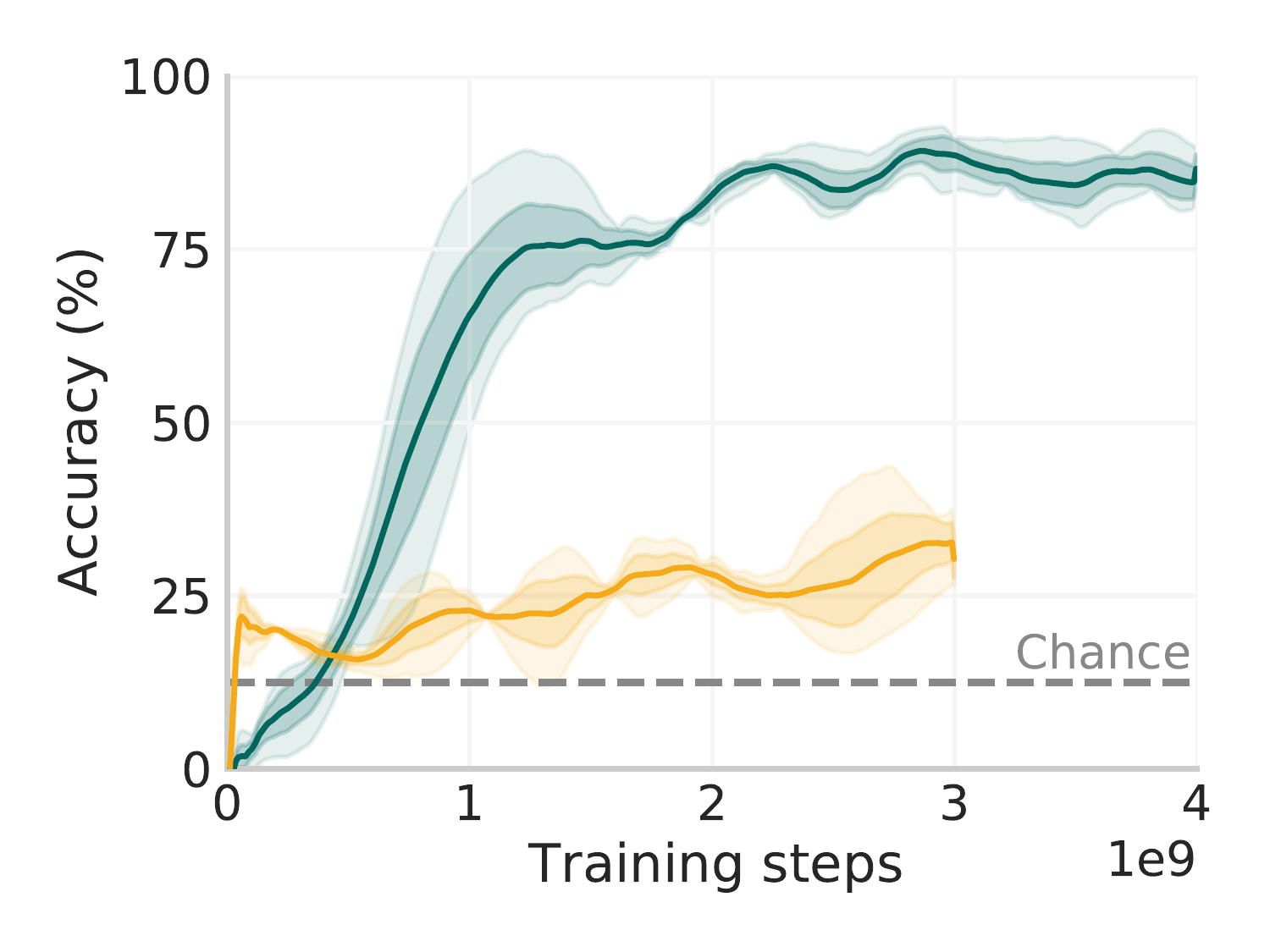}
    \captionsetup{width=.8\textwidth}
    \caption{Evaluate 8 dances, long delays.}
    \label{fig:supp_exp:ballet_gen:8long}
    \end{subfigure}
    \caption{HCAM (labeled as HTM) trained on 2, 4, or 6 dance ballets generalizes well to 8 dance ballets, while TrXL does not. Results are from two seeds in each condition.}
    \label{fig:supp_exp:ballet_gen}
\end{figure}

In the main text Ballet experiments (Fig. \ref{fig:exp:ballet}), we compared differences only in training performance. In Fig. \ref{fig:supp_exp:ballet_gen}, we show that HCAM is also able to generalize well from training on 2, 4, or 6 dances, to evaluation on 8 dances with either short or long delays.

\subsection{Analyzing memory attention in the rapid word learning tasks} \label{app:supp_exp:attention_analyses}

In this section, we show that the HCAM agent's extrapolation to cross-episode evaluation (Fig. \ref{fig:exp:fast_binding_tasks:episodes}) in the rapid word learning tasks is supported by selective patterns of memory access. In summary, we show that when the agent is asked to recall a word from many episodes before, all layers of its memory exhibit significantly higher attention to phase when it learned that word, compared to its attention patterns when tested on a more recent word. 

Specifically, we considered the case of evaluation across 4 episodes, with 1 distractor each. To lay out the ``super-episode'' structure of this setting very explicitly, it proceeds through 4 episodes, each of which has three distinct phases. In other words, the episodes proceed as: learn 1, distract 1, test 1, learn 2, distract 2, test 2, learn 3, distract 3, test 3, learn 4, distract 4, test 4. Tests 1-3 evaluate memory for a word learned in their respective learn phases 1-3, but test 4 tests surprises the agent by asking about a word learned in learning phase 1 (Fig. \ref{fig:exp:fast_binding_tasks:episodes}). As shown in the main text (Fig. \ref{fig:exp:fastbind:across_eps_4_1}), the HCAM-based agent achieves above 90\% performance on test 4, despite never being evaluated on words from earlier episodes during training.

\begin{figure}[htb]
    \centering
    \begin{subfigure}{0.5\textwidth}
    \includegraphics[width=\textwidth]{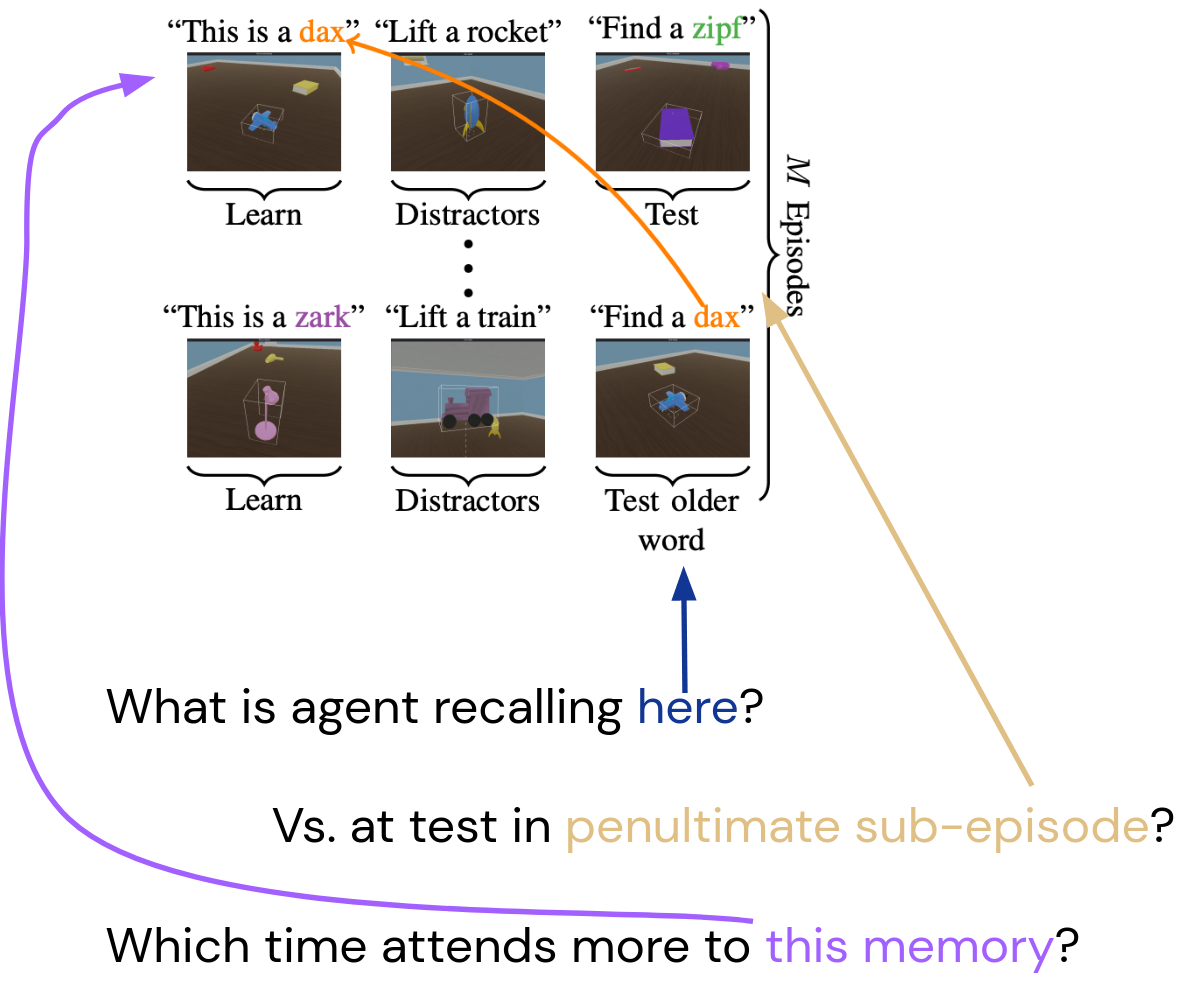}
    \caption{Analysis sketch.}
    \label{fig:supp_exp:fb_attn:diagram}
    \end{subfigure}%
    \begin{subfigure}{0.5\textwidth}
    \includegraphics[width=\textwidth]{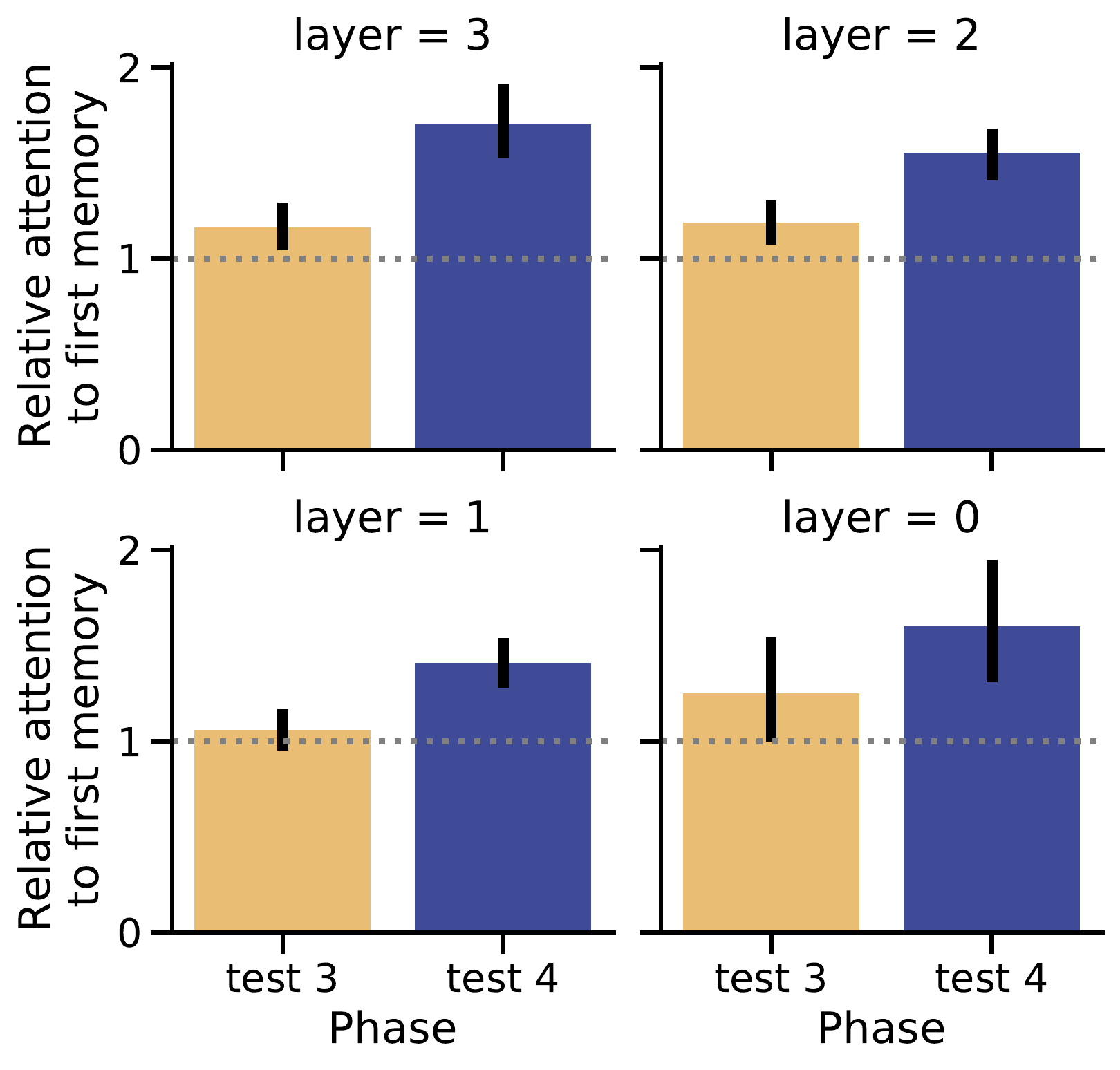}
    \caption{Results.}
    \label{fig:supp_exp:fb_attn:results}
    \end{subfigure}
    \caption{The HCAM-based agent selectively attends to relevant memories in the rapid-word-learning generalization tasks. (\subref{fig:supp_exp:fb_attn:diagram}) We analyze the relative weight of attention to the first memory from the first learning phase, when the agent is asked to recall a word form the first phase in test 4 vs. when it is asked to recall a word from a later phase in test 3. (\subref{fig:supp_exp:fb_attn:results}) Across all 4 memory layers, the agent attends more strongly to its memory of this first learning phase when that memory is relevant---in test 4---compared to when that memory is irrelevant---in test 3. (This plot shows relative attention weights---that is, attention weights divided by average attention weight, so that if the agent were attending uniformly to all memories, their relative attention weights would be 1, indicated by the dotted line. This plot shows averages and 95\%-CIs across the 93 episodes where the agent made a correct choice in test 4, out of 100 total super-episodes run.)} \label{fig:supp_exp:fb_attn:summary}
\end{figure}

To investigate the attention patterns underlying this result, we analyzed what chunks of memory the agent was attending to in test phase 4 (Fig. \ref{fig:supp_exp:fb_attn:summary}). In particular, we ran the agent on 100 of these episodes, and saved its attention weights for each phase of the experiment. In 93 of these episodes, the agent chose correctly in test 4. Within those 93 episodes, we then evaluated the agent's attention to the first memory chunk of learn 1, the learning phase of the first episode (which generally contained most, if not all, of that learn phase). We compare the attention weight on this chunk when the agent is tested on one of these words in test phase 4 to a within-episode control: the relative weight when the agent is tested on a word from learn 3 during test phase 3. Is the agent attending more to its memory of learn phase 1 in test 4, when that memory is relevant, compared to test 3, when it is irrelevant? In fact, we find that across all four layers of the agent's memory, the agent is attending more strongly to the memory when it is relevant than when it is not. That is, the agent is distributing its attention intelligently, in a query-dependent way. It is attending most strongly to memories of the learn 1 phase when it is asked to recall a word from it, compared to a within-super-episode control where it is asked to recall a word from another phase (learn 3). 

\begin{figure}[p]
    \centering
    \begin{subfigure}{0.66\textwidth}
        \includegraphics[width=\textwidth]{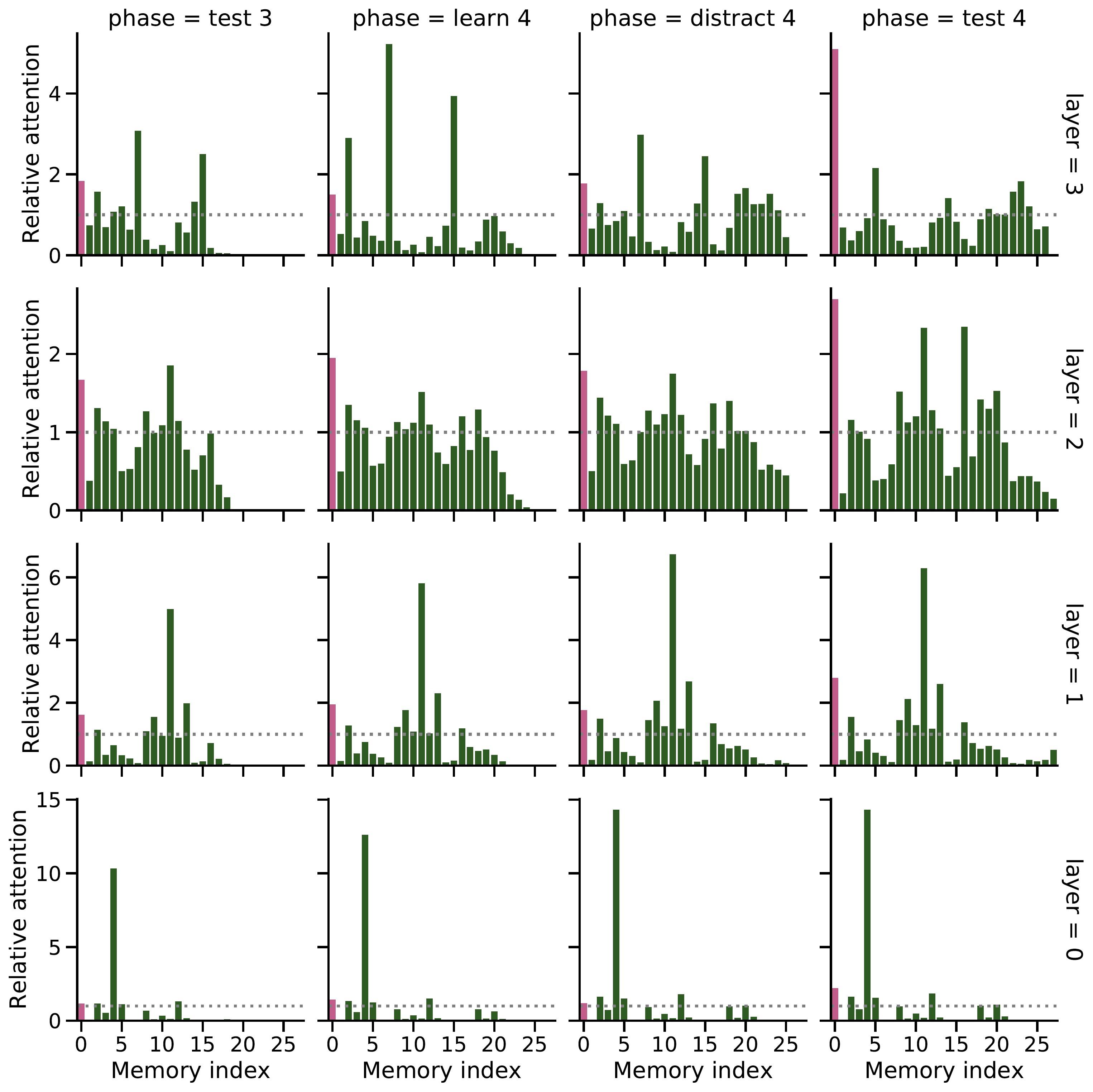}
        \caption{Example super-episode 1}
    \end{subfigure}\\
    \begin{subfigure}{0.66\textwidth}
        \includegraphics[width=\textwidth]{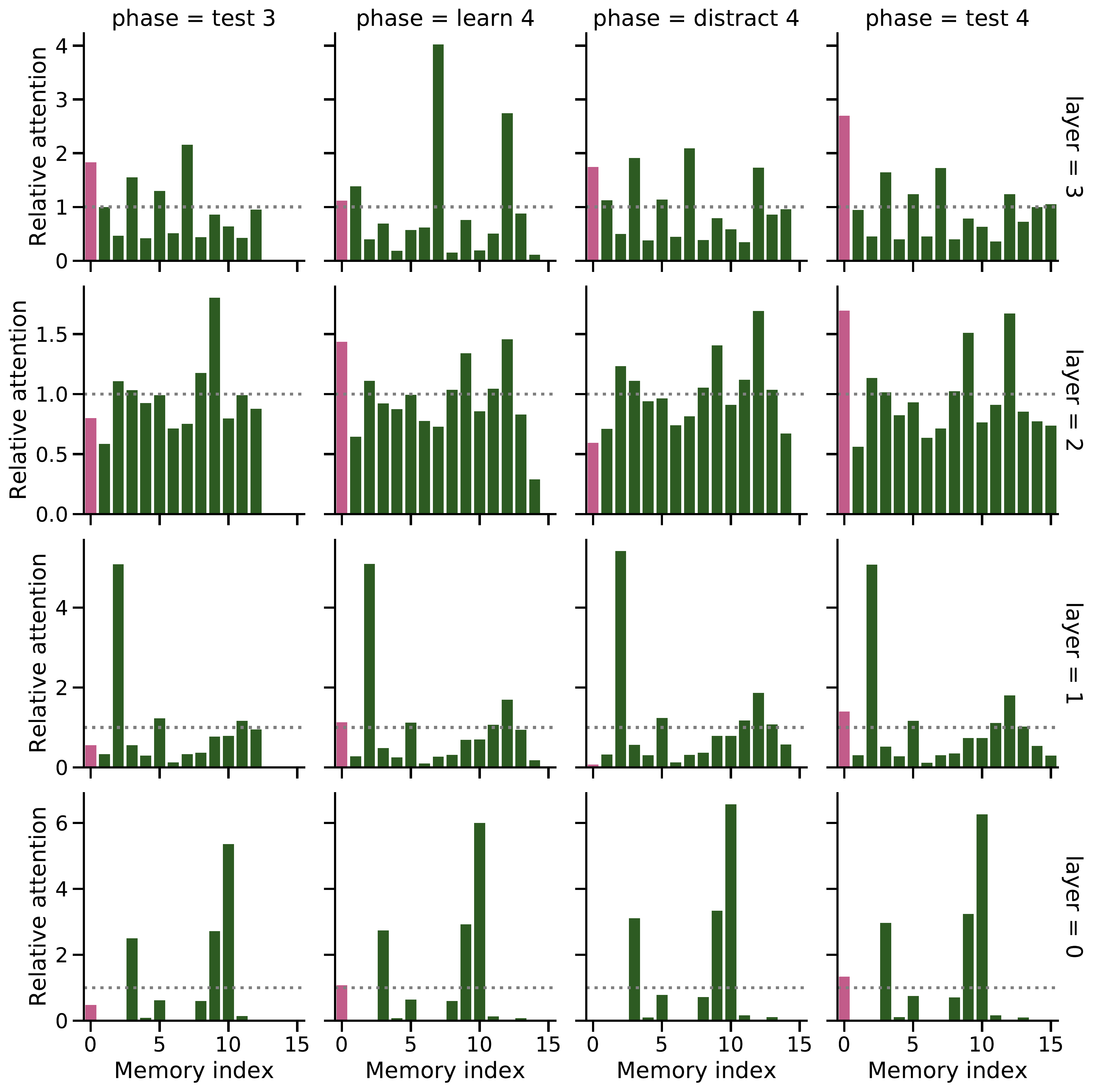}
        \caption{Example super-episode 2}
    \end{subfigure}\\
    \caption{Patterns of attention within four phases of two example (randomly chosen) super-episodes on the rapid-word-learning generalization tasks. The higher layers of the network show clear shifts in attention patterns between the different phases, although with some consistent biases within each super-episode, especailly at the lower layers. The pink bar shows the weight on the first chunk from learn phase 1---the analyis shown in Fig. \ref{fig:supp_exp:fb_attn:summary} corresponds to comparing the pink bars in the first and last column, aggregated across many more episodes. (This plot shows relative attention weights---that is, attention weights divided by average attention weight, so that chance level relative attention would be 1, indicated by the dotted line. Note that these were computed \emph{before} the top-\(k\) operation on the attention weights, which is why more than 16 weights are active. The two super-episodes had different lengths, which is why more memories were stored in the first.)} \label{fig:supp_exp:fb_attn:detailed}
\end{figure}

The results above involve several levels of aggregation: averaging within phases, and across many super-episodes. To give a flavor for the complexity of the full patterns of attention, in Fig. \ref{fig:supp_exp:fb_attn:detailed} we show the average attention weights for every layer across all the stored memories, in the final four phases of two super-episodes. 

\subsection{Varying chunk sizes}  \label{app:supp_exp:varying_chunk_size}
\begin{figure}[htb]
    \centering
    \begin{subfigure}{0.33\textwidth}
    \includegraphics[width=\textwidth]{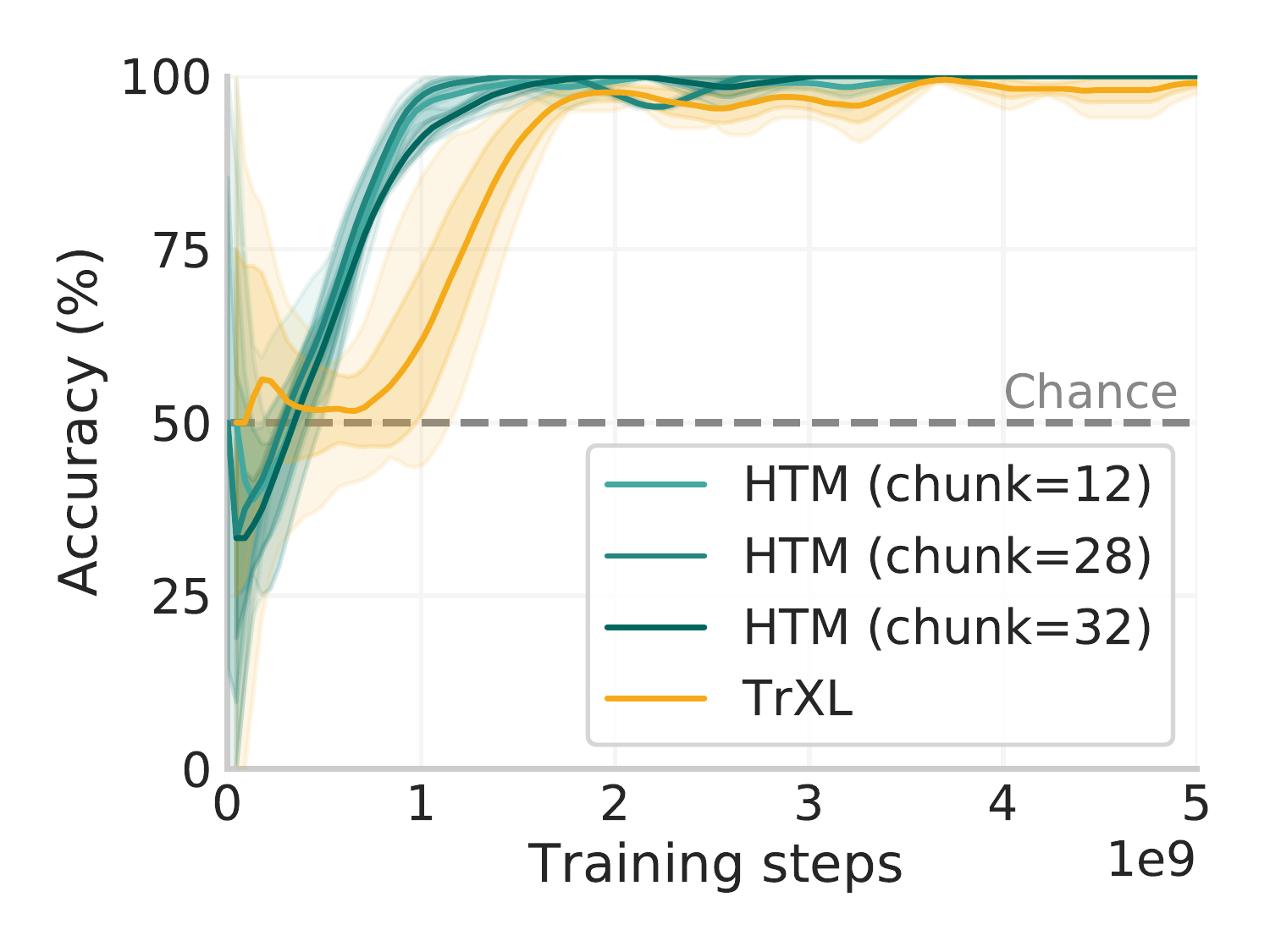}
    \caption{2 dances, short delays.}
    \label{fig:supp_exp:ballet:2}
    \end{subfigure}%
    \begin{subfigure}{0.33\textwidth}
    \includegraphics[width=\textwidth]{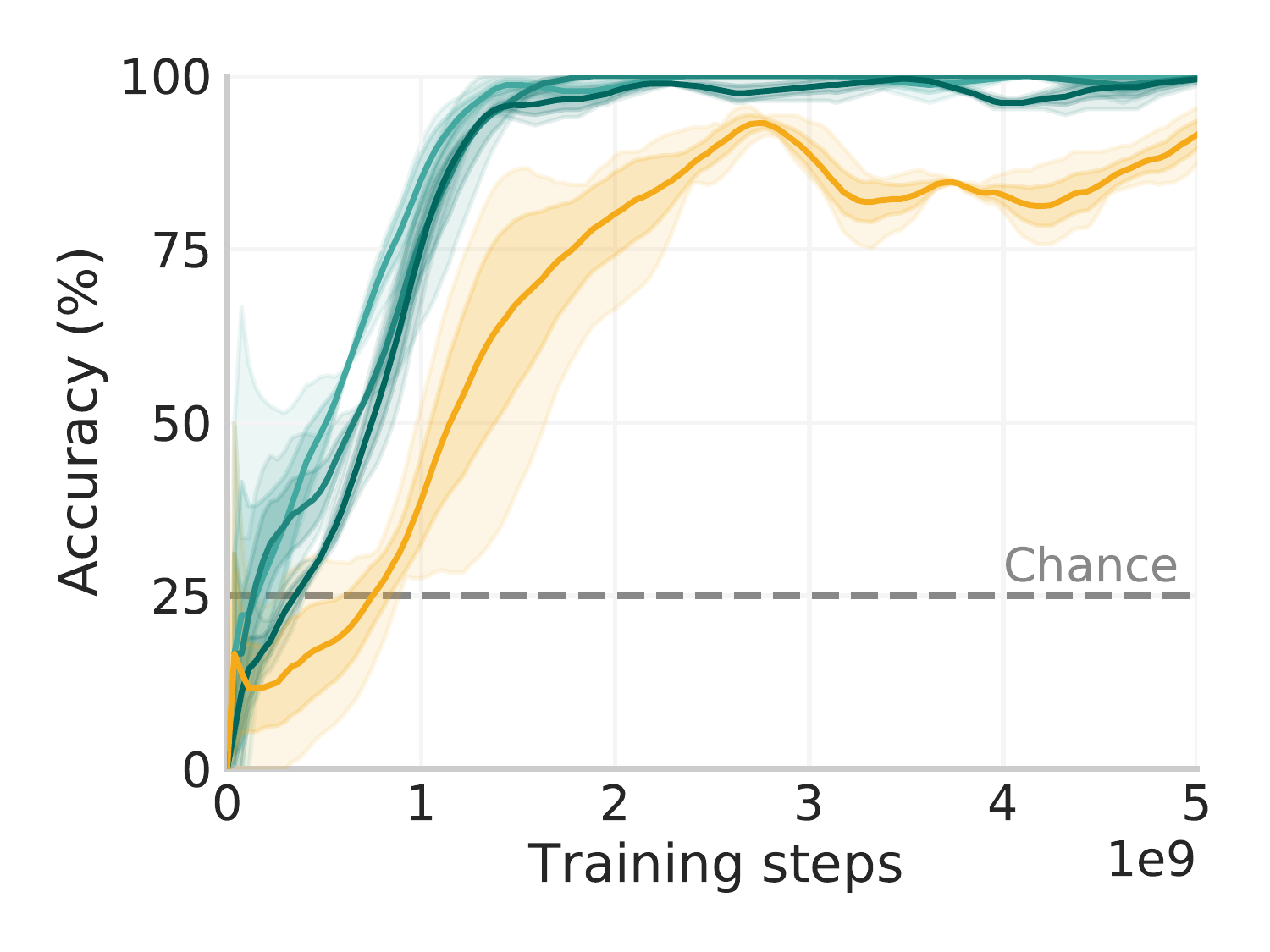}
    \caption{4 dances, short delays.}
    \label{fig:supp_exp:ballet:4}
    \end{subfigure}%
    \begin{subfigure}{0.33\textwidth}
    \includegraphics[width=\textwidth]{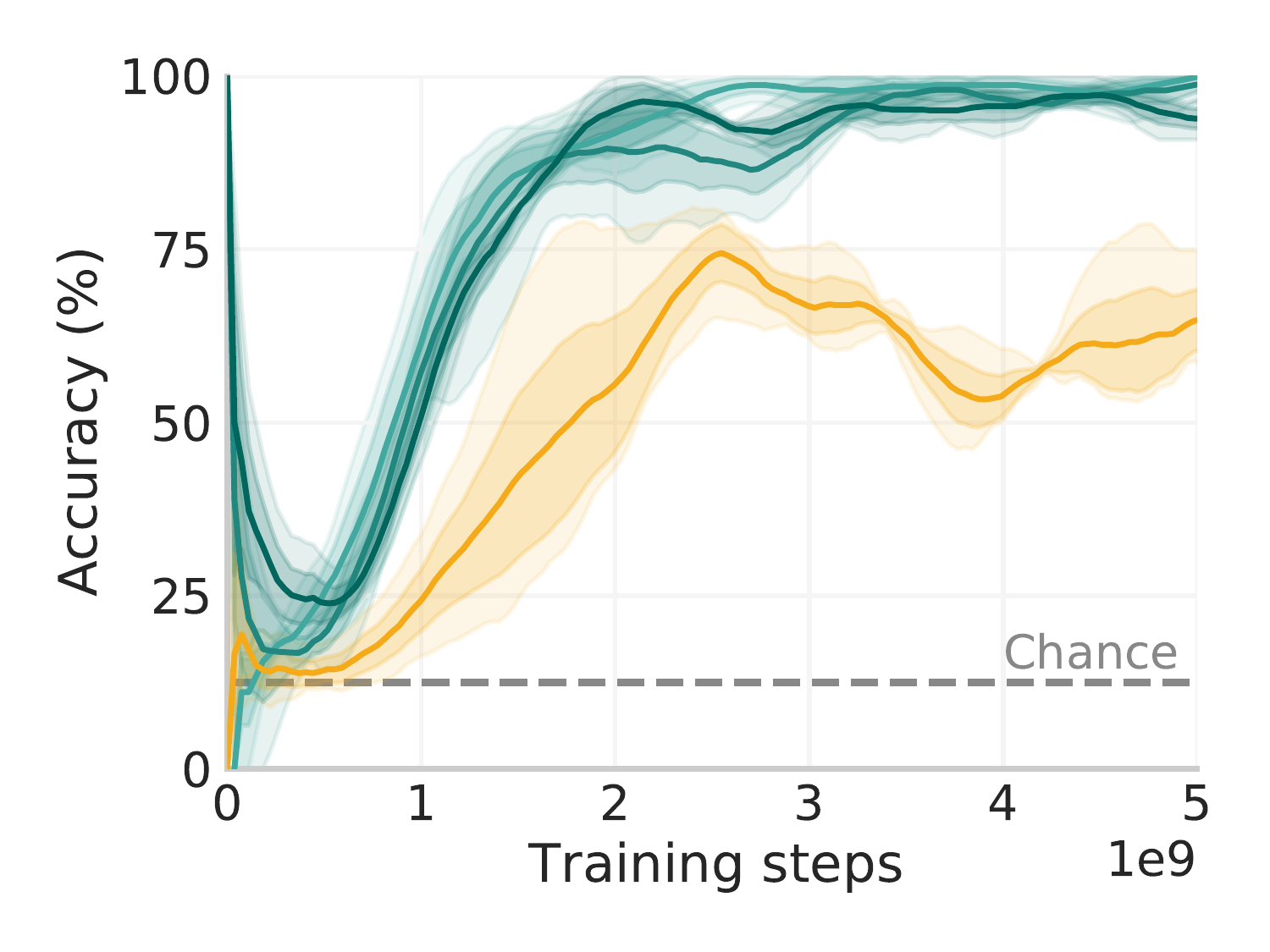}
    \caption{8 dances, short delays.}
    \label{fig:supp_exp:ballet:8}
    \end{subfigure}\\
    \begin{subfigure}{0.33\textwidth}
    \includegraphics[width=\textwidth]{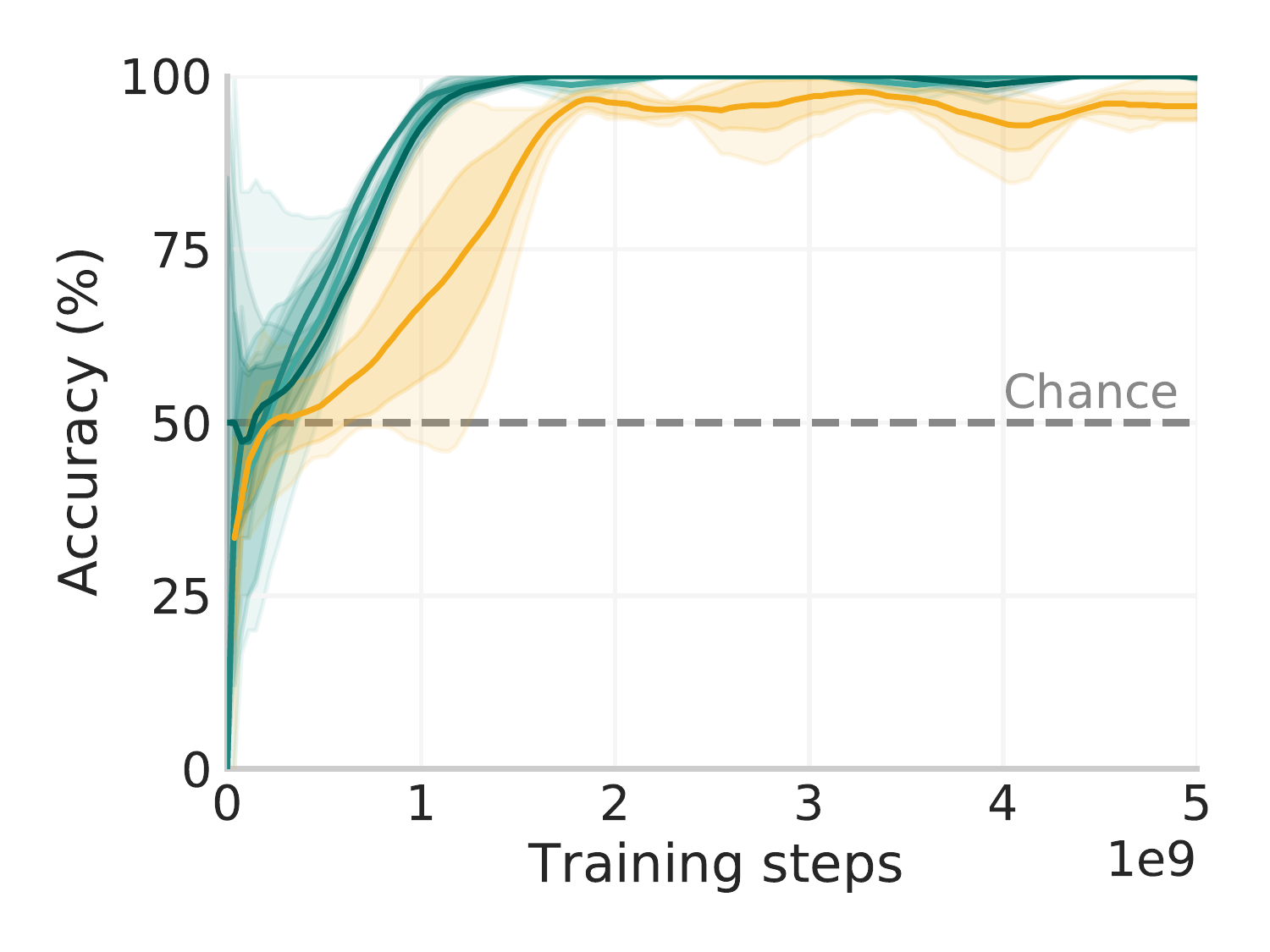}
    \caption{2 dances, long delays.}
    \label{fig:supp_exp:ballet:2long}
    \end{subfigure}%
    \begin{subfigure}{0.33\textwidth}
    \includegraphics[width=\textwidth]{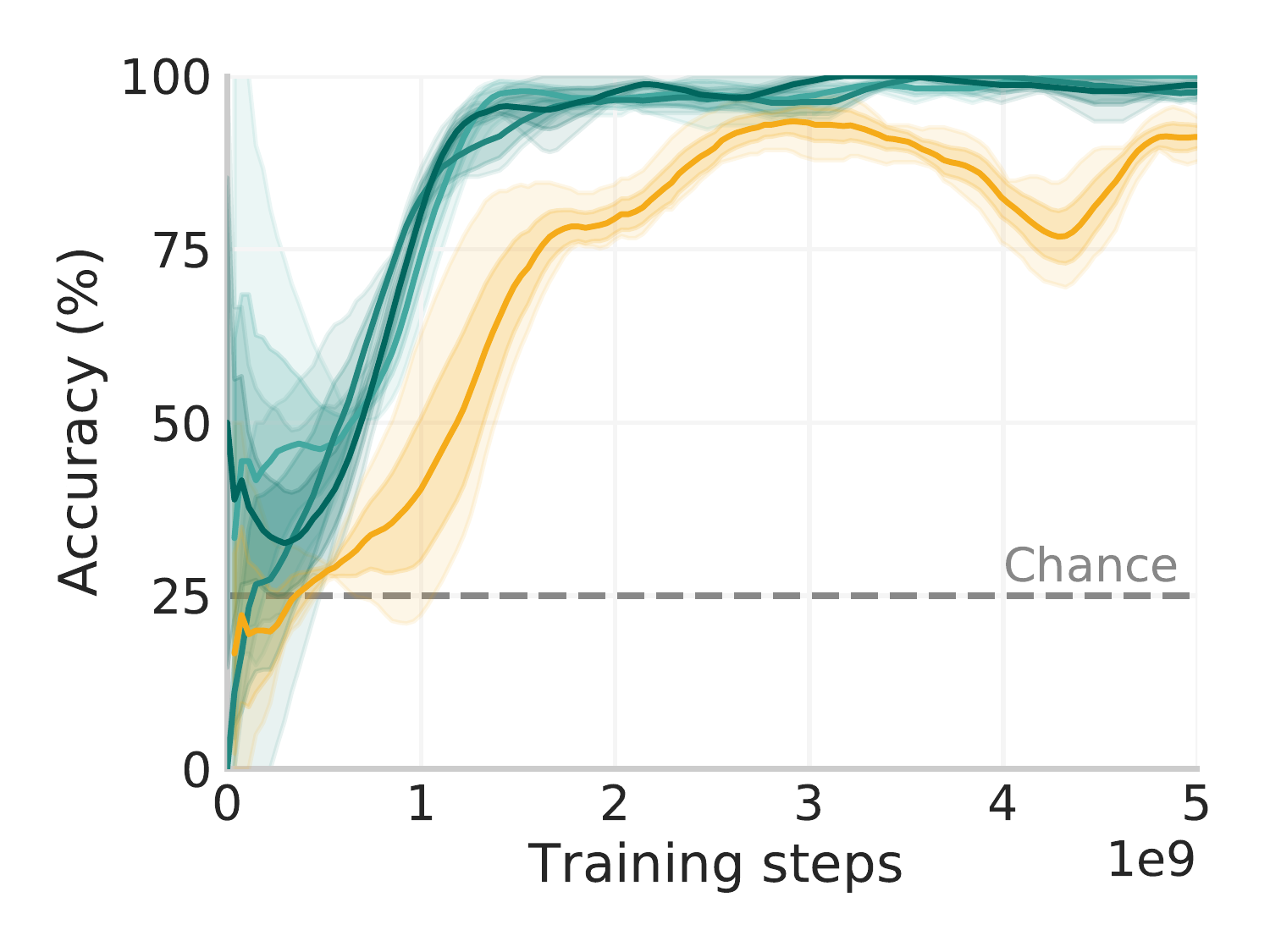}
    \caption{4 dances, long delays.}
    \label{fig:supp_exp:ballet:4long}
    \end{subfigure}%
    \begin{subfigure}{0.33\textwidth}
    \includegraphics[width=\textwidth]{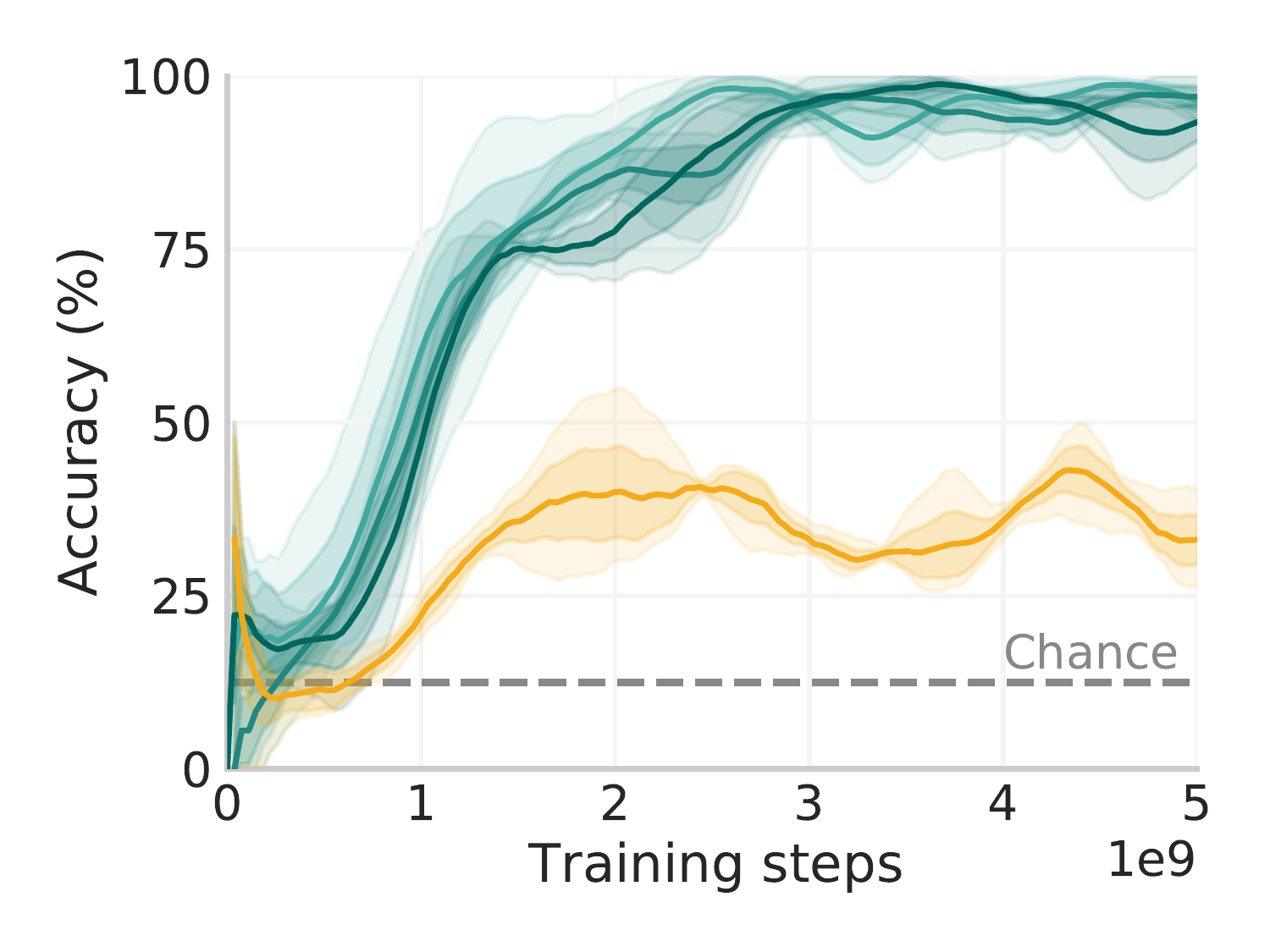}
    \caption{8 dances, long delays.}
    \label{fig:supp_exp:ballet:8long}
    \end{subfigure}
    \caption{Comparison of HCAM (labeled as HTM) with different chunk sizes to TrXL across the different ballet levels. The performance of the HCAM model is robust to varying chunk size, indicating that HCAM does not need a task-relevant segmentation to perform well. The results reported in the main text use chunk size 32; panels \subref{fig:supp_exp:ballet:2}, \subref{fig:supp_exp:ballet:8}, and \subref{fig:supp_exp:ballet:8long} correspond to main text Fig. \ref{fig:exp:ballet}. These comparisons were run before a minor bug was fixed in HCAM memory writing. Results are from three seeds in each condition.}
    \label{fig:supp_exp:ballet}
\end{figure}

In Fig. \ref{fig:supp_exp:ballet} we show that the performance of the HCAM model is robust to varying chunk sizes in the ballet task; therefore its advantage in this task is not due to having additional information about the correct segmentation of the episodes. Furthermore, HCAM performs well even when its chunk size is 12, and so the total number of timepoints it can attend to at each layer is smaller than the number that the TrXL can attend to at each layer. Thus its advantage in these tasks is not due to attending to more of the episode, but rather to attending more effectively. 

\subsection{Varying \(k\) for memory selection} \label{app:supp_exp:varying_k}
\begin{figure}[htb]
    \centering
    \begin{subfigure}{0.33\textwidth}
    \includegraphics[width=\textwidth]{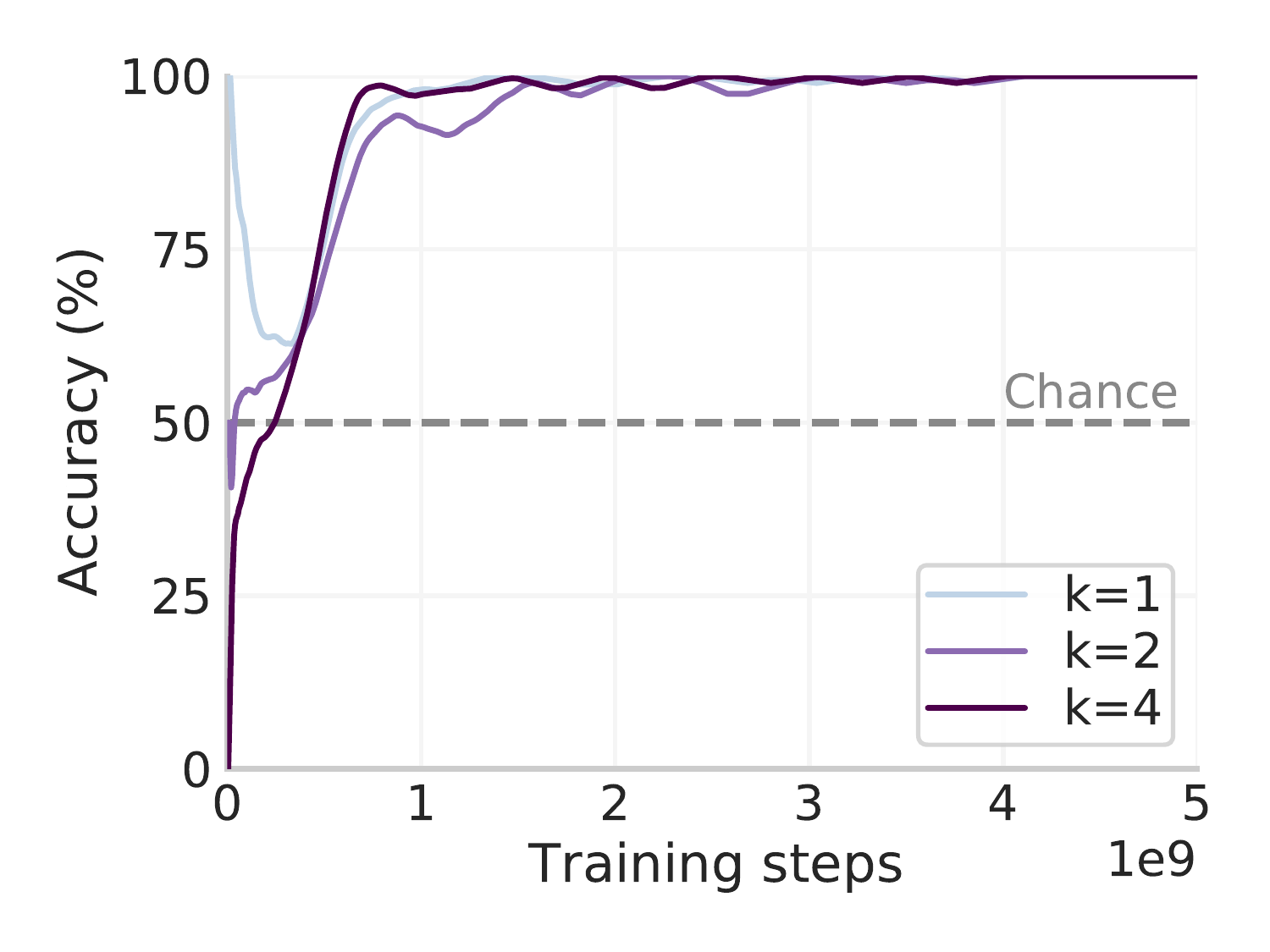}
    \caption{Ballet, 2 dances, short delays.}
    \label{fig:supp_exp:varying_k:ballet2}
    \end{subfigure}%
    \begin{subfigure}{0.33\textwidth}
    \includegraphics[width=\textwidth]{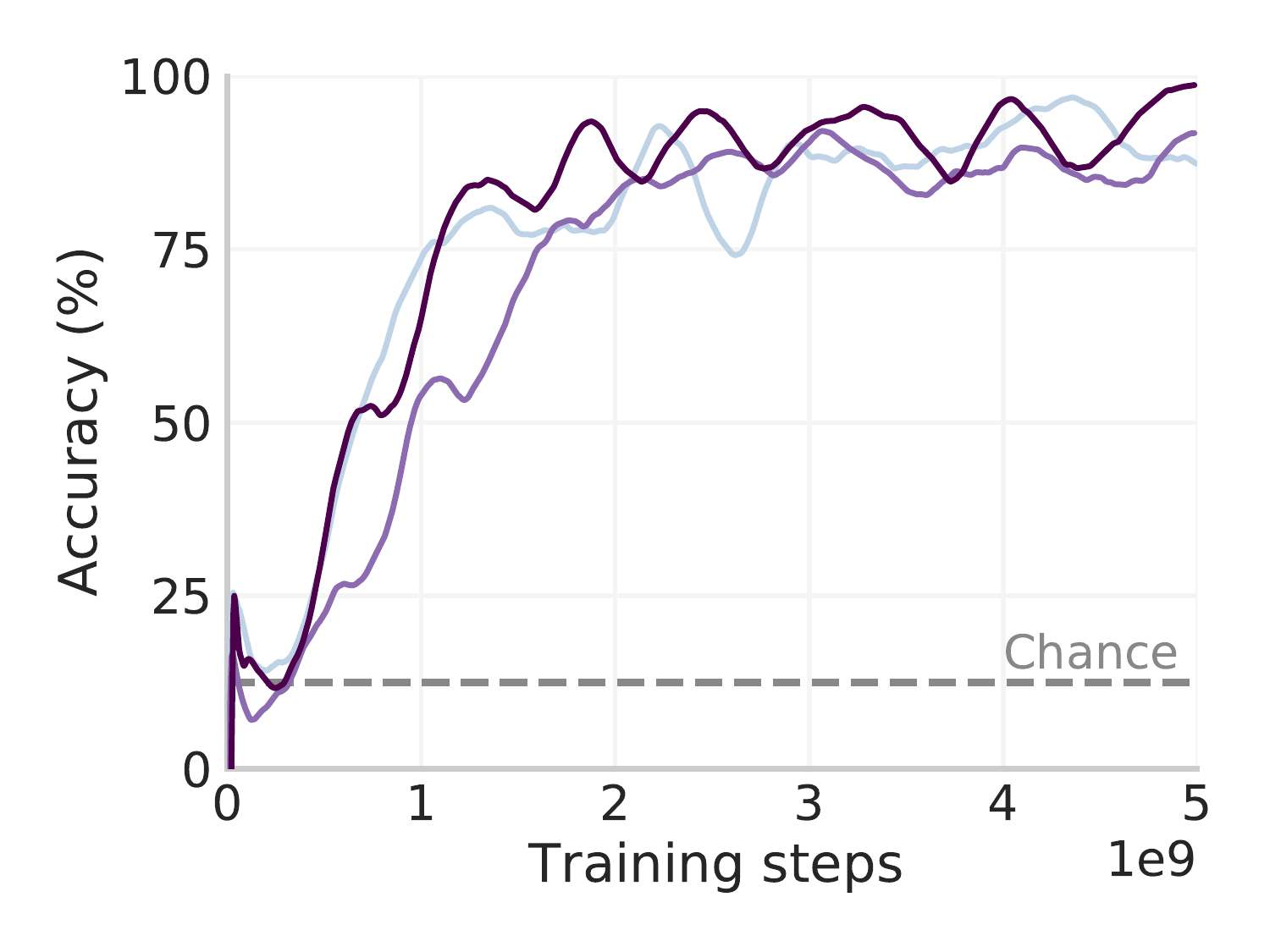}
    \caption{Ballet, 8 dances, long delays.}
    \label{fig:supp_exp:varying_k:ballet8long}
    \end{subfigure}\\
    \begin{subfigure}{0.33\textwidth}
    \includegraphics[width=\textwidth]{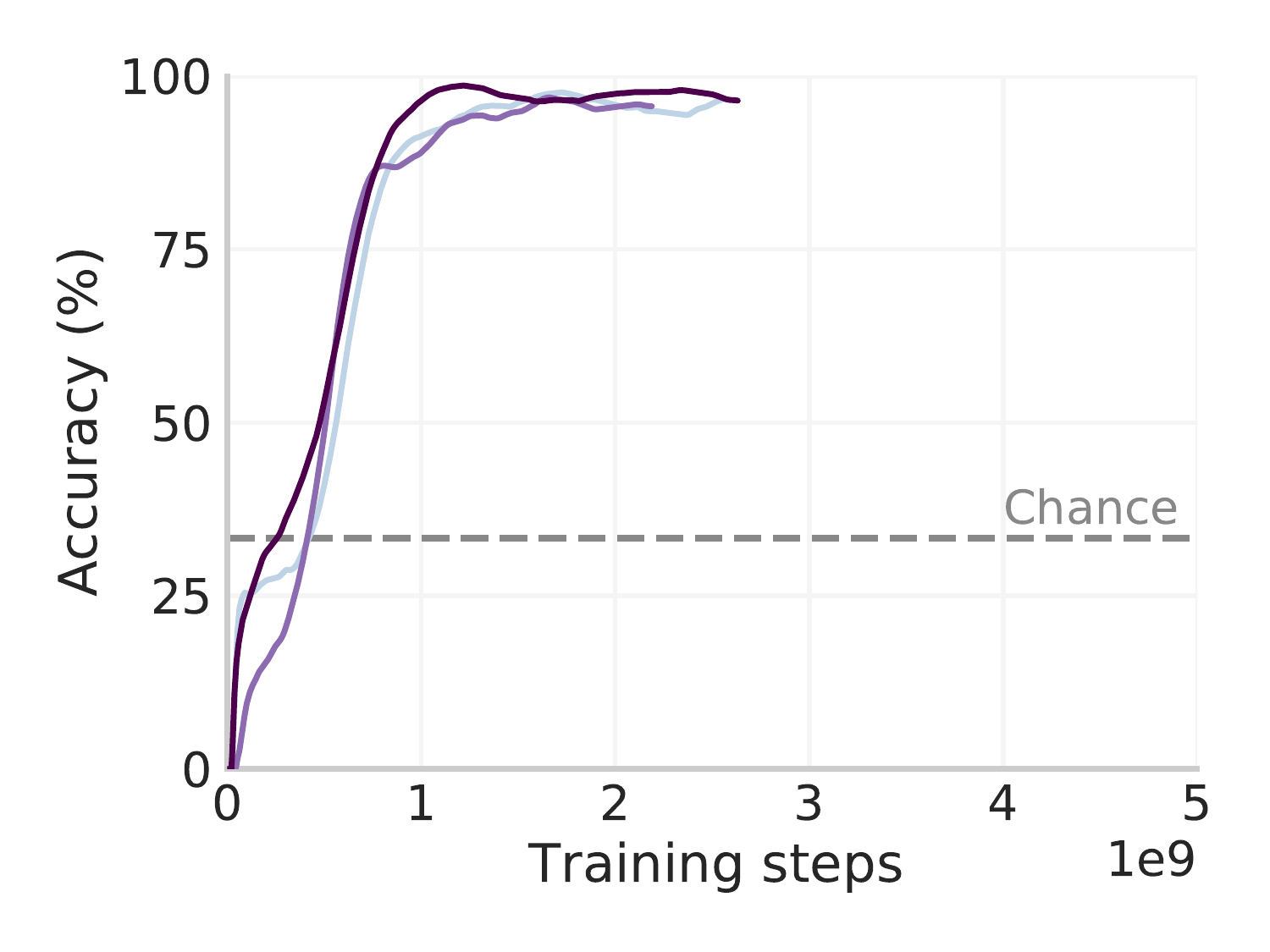}
    \caption{Object permanence, no delays.}
    \label{fig:supp_exp:varying_k:objects}
    \end{subfigure}%
    \begin{subfigure}{0.33\textwidth}
    \includegraphics[width=\textwidth]{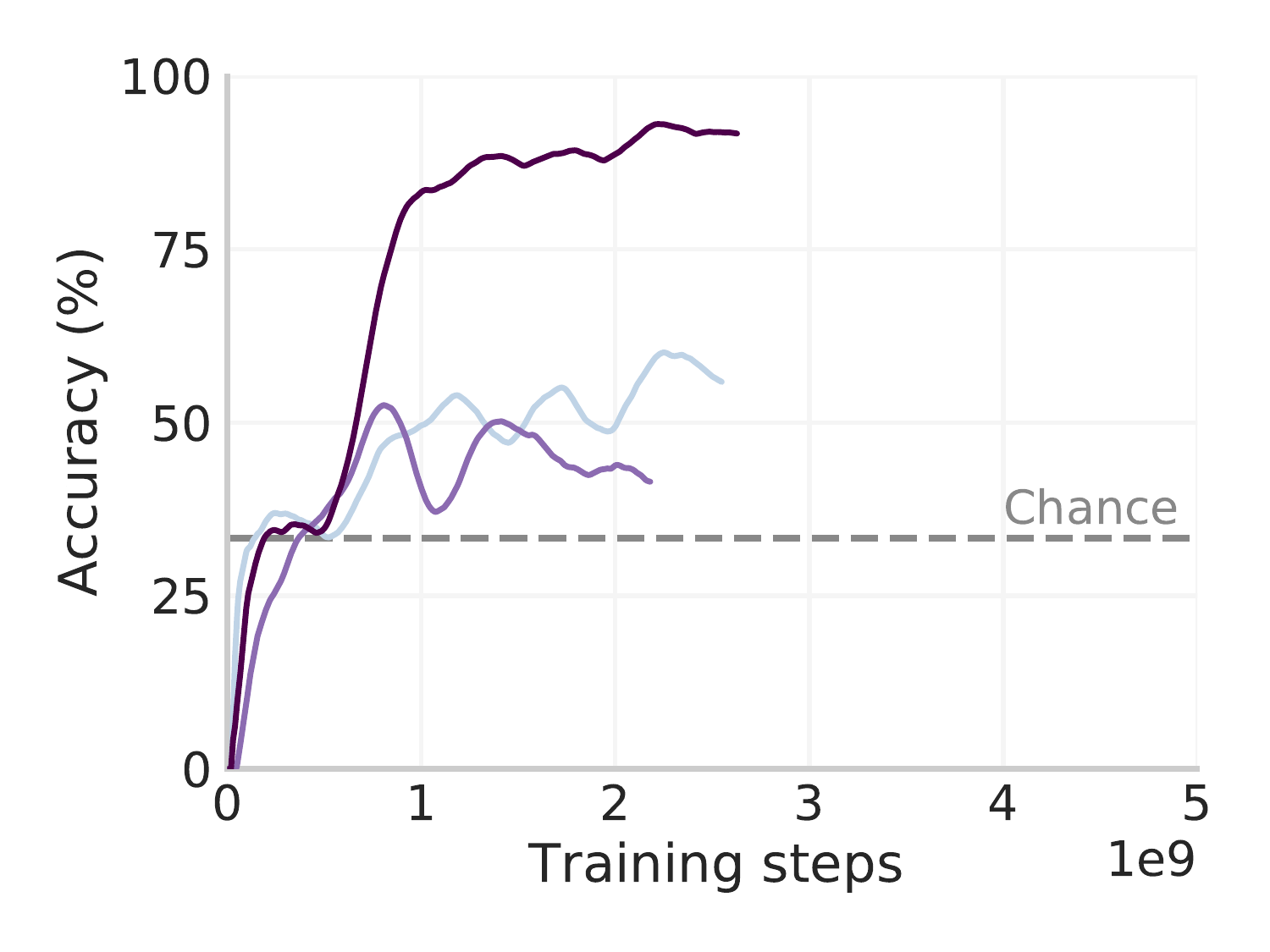}
    \caption{Object permanence, 30s delays.}
    \label{fig:supp_exp:varying_k:objects30s}
    \end{subfigure}\\
    \caption{Varying the number \(k\) of memory chunks selected in the top-\(k\) step of hierarchical attention. Performance is relatively robust to \(k\) smaller than used in the main experiments (\(k=16\)), in fact shorter tasks can be learned even with \(k=1\), while longer tasks require \(k\geq 4\). (\subref{fig:supp_exp:varying_k:ballet2}-\subref{fig:supp_exp:varying_k:ballet8long}) On the ballet tasks, HCAM can learn fairly well even with \(k=1\), both for the shorter and longer tasks. (\subref{fig:supp_exp:varying_k:objects}-\subref{fig:supp_exp:varying_k:objects30s}) On the object permanence tasks, HCAM can learn the shortest tasks well even with \(k=1\), but struggles to learn the longer tasks unless \(k\geq 4\). (One seed per condition.)}
    \label{fig:supp_exp:varying_k}
\end{figure}

In Fig. \ref{fig:supp_exp:varying_k} we show that HCAM is robust to varying the number \(k\) of memory chunks selected in the top-\(k\) step of the hierarchical attention at each layer. Specifically, while the main text experiments used \(k=16\), we show that HCAM is able to perform the ballet and object permanence tasks well even with \(k=4\), and can perform the shorter tasks even with \(k=2\) or even \(k=1\). While it initially surprised us that hard memory selection with \(k=1\) did not harm the optimization process, it resonates with recent results from the Switch Transformer \citep{fedus2021switch}, which found that hard selection of a single expert was effective in a mixture-of-experts style model.

\subsection{The importance of self-supervised learning} \label{app:supp_exp:selfsupervised}
In Fig. \ref{app:supp_exp:selfsupervised}, we show that the self-supervised loss (image + language reconstruction at the agent output) that we used as an auxiliary loss during training is necessary for our model to achieve good performance on the ballet and fast-binding tasks. This is presumably because this loss forces the model to encode the input in detail, and therefore that information is in-principle retrievable from the state representations stored in the agent memory. 
\begin{figure}[htb]
    \centering
    \begin{subfigure}{0.33\textwidth}
    \includegraphics[width=\textwidth]{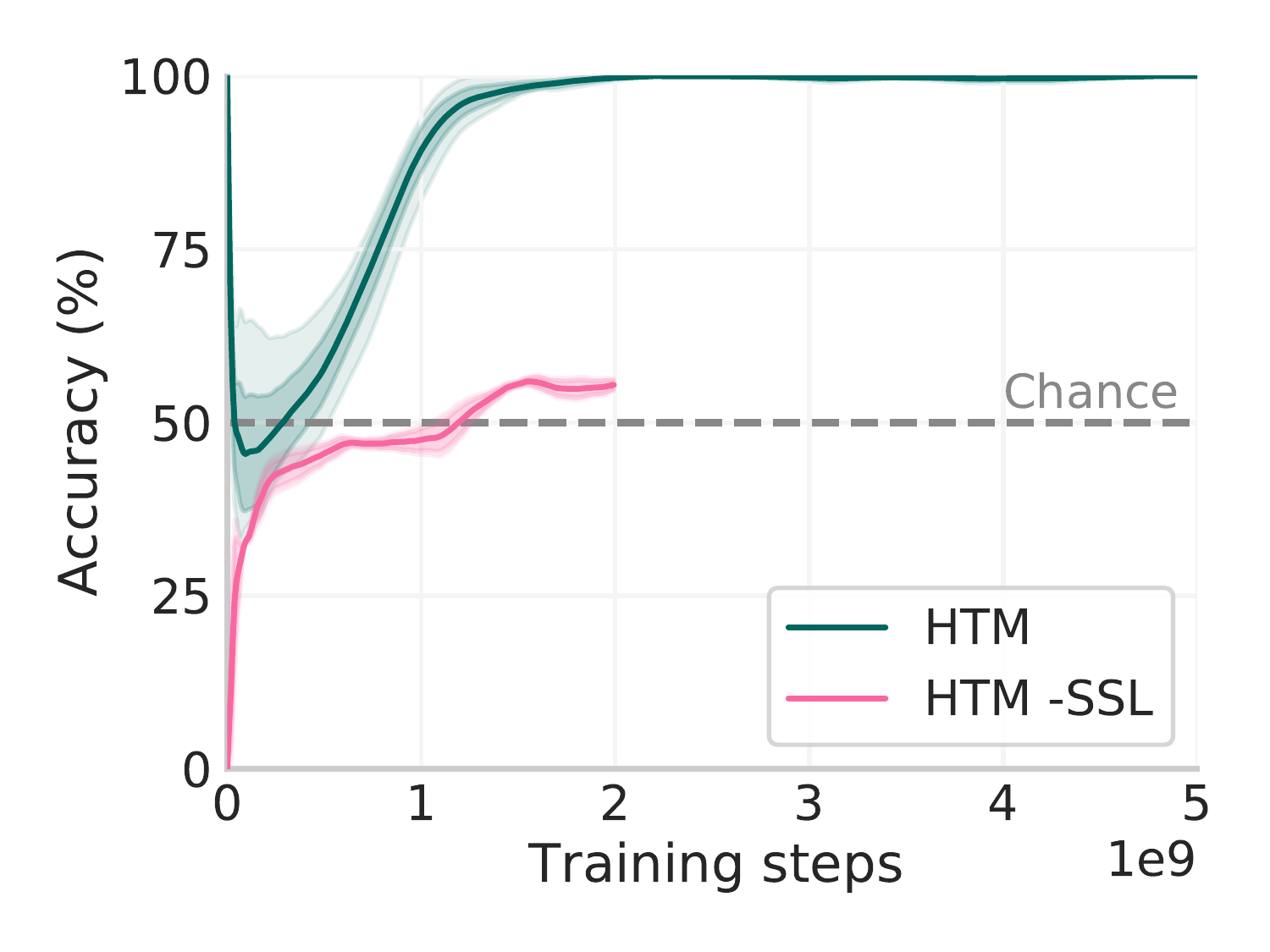}
    \caption{Ballet, 2 dances, short delays.}
    \label{fig:supp_exp:selfsupervised:ballet2}
    \end{subfigure}%
    \begin{subfigure}{0.33\textwidth}
    \includegraphics[width=\textwidth]{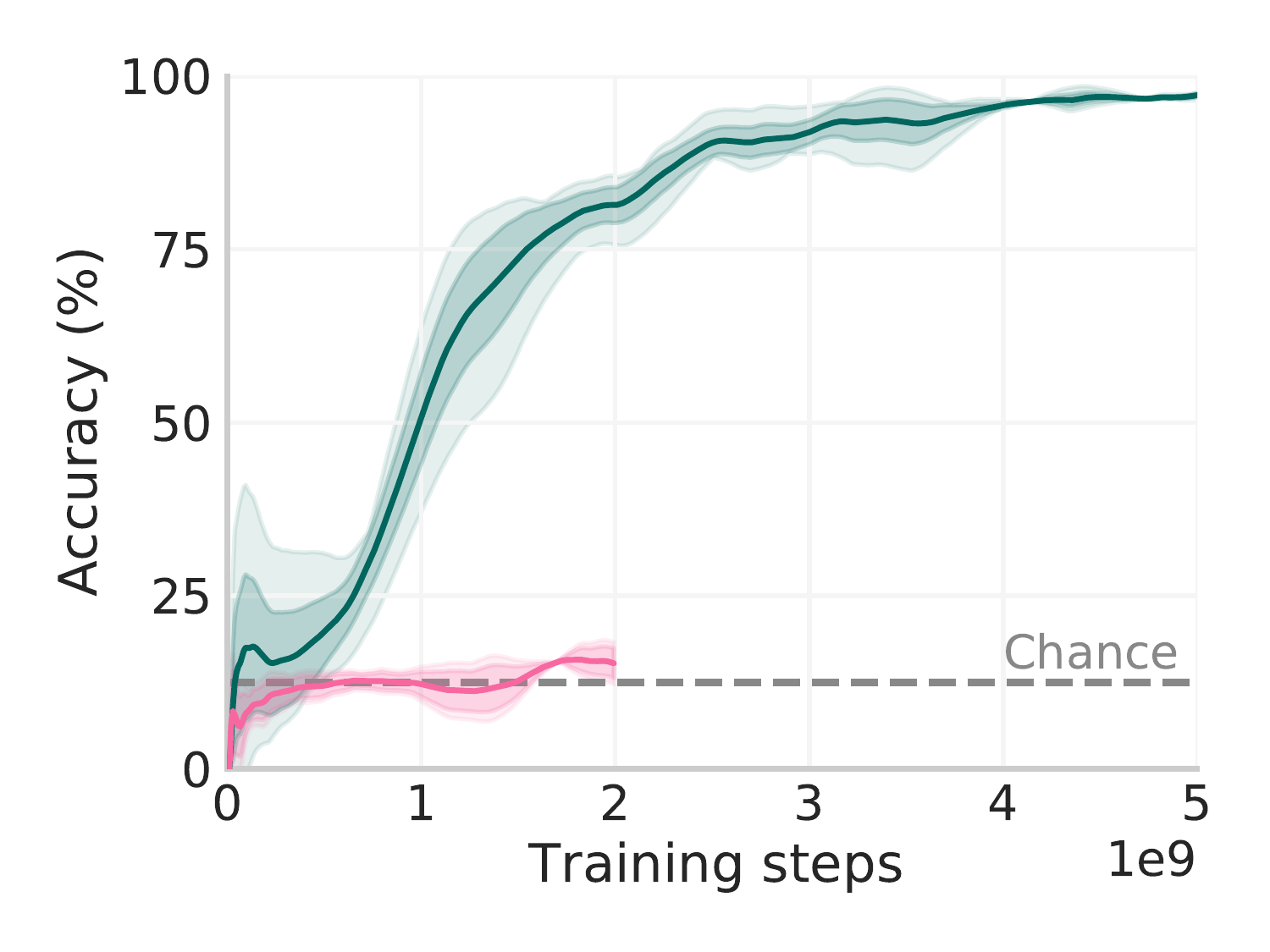}
    \caption{Ballet, 8 dances, long delays.}
    \label{fig:supp_exp:selfsupervised:ballet8long}
    \end{subfigure}\\
    \begin{subfigure}{0.33\textwidth}
    \includegraphics[width=\textwidth]{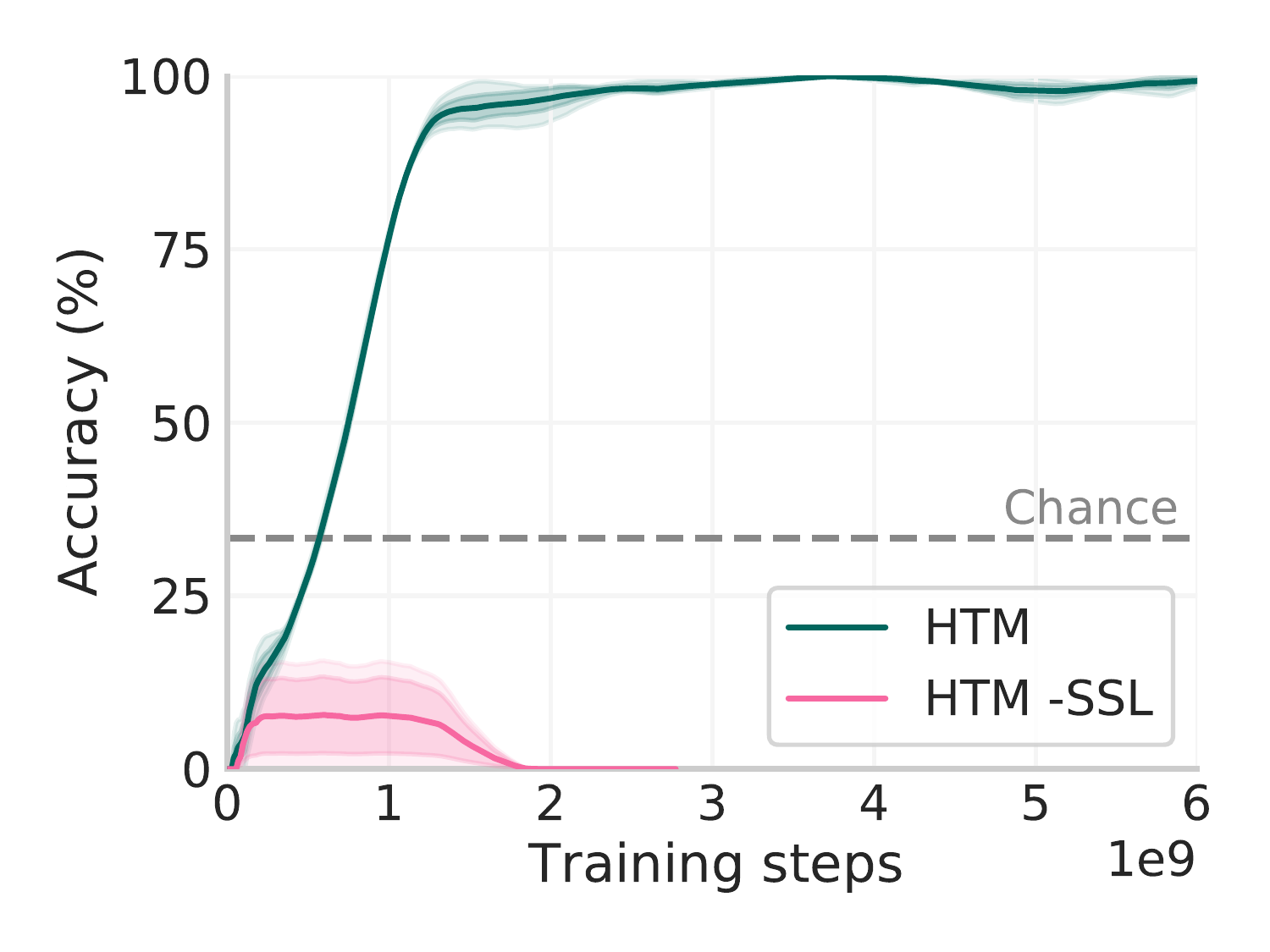}
    \caption{Rapid word learning, no distractors.}
    \label{fig:supp_exp:selfsupervised:fb0}
    \end{subfigure}%
    \begin{subfigure}{0.33\textwidth}
    \includegraphics[width=\textwidth]{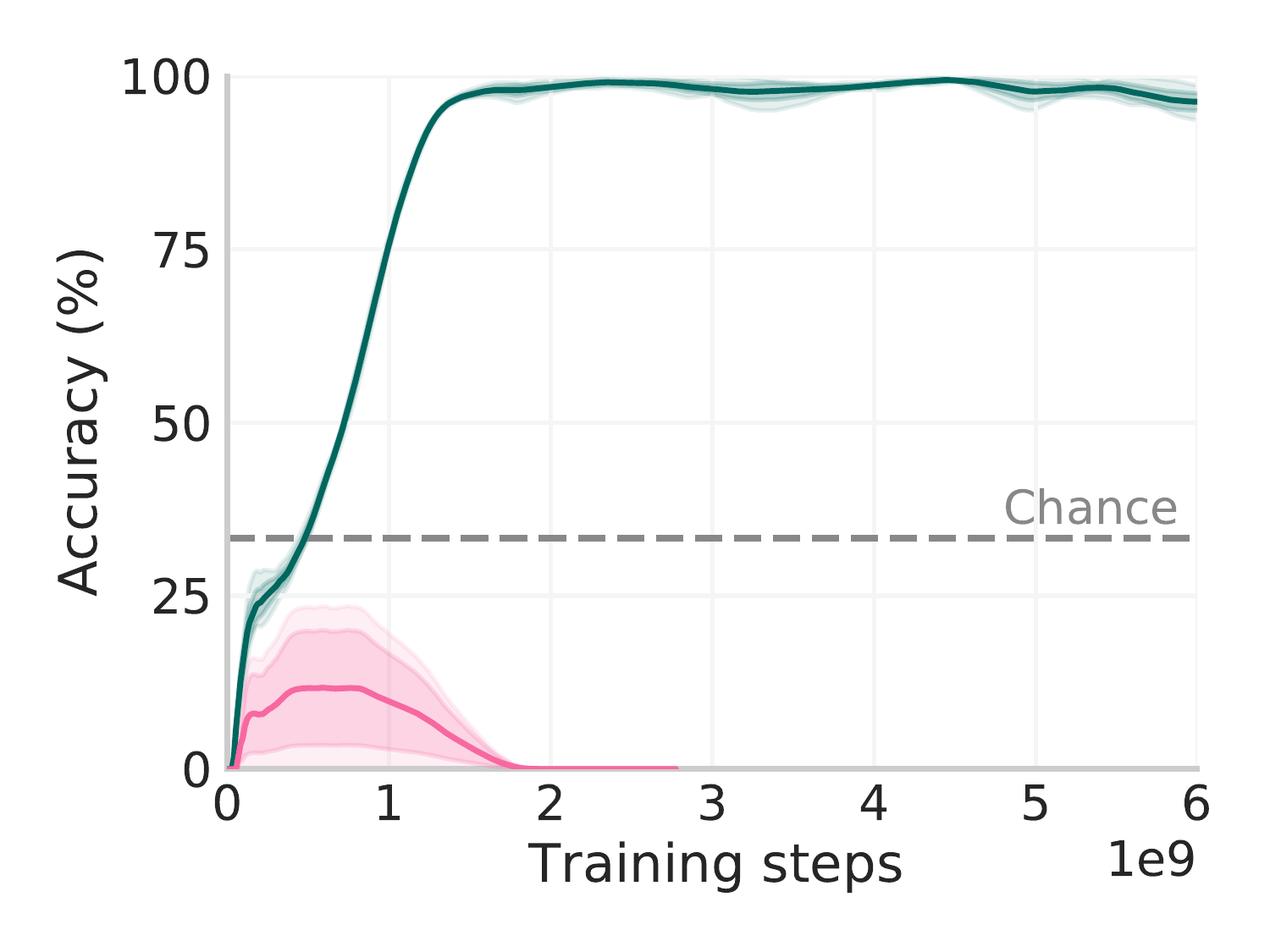}
    \caption{Rapid word learning, 2 distractors.}
    \label{fig:supp_exp:selfsupervised:fb2}
    \end{subfigure}
    \caption{The self-supervised loss (SSL) is necessary for the model to learn appropriate representations. When the SSL is disabled, the model either fails to achieve substantially-above-chance performance, for example in the ballet tasks (\subref{fig:supp_exp:selfsupervised:ballet2}-\subref{fig:supp_exp:selfsupervised:ballet8long}), or fails to learn the tasks to even chance level, as in the rapid word learning tasks (\subref{fig:supp_exp:selfsupervised:fb0}-\subref{fig:supp_exp:selfsupervised:fb2}). (HTM refers to HCAM, see note above. 3 runs per condition for main results, 2 runs per condition for results without SSL.)}
    \label{fig:supp_exp:selfsupervised}
\end{figure}

\subsection{Memory layer gating} \label{app:supp_exp:gating}
\begin{figure}[htb]
    \centering
    \begin{subfigure}{0.33\textwidth}
    \centering
    \includegraphics[width=\textwidth]{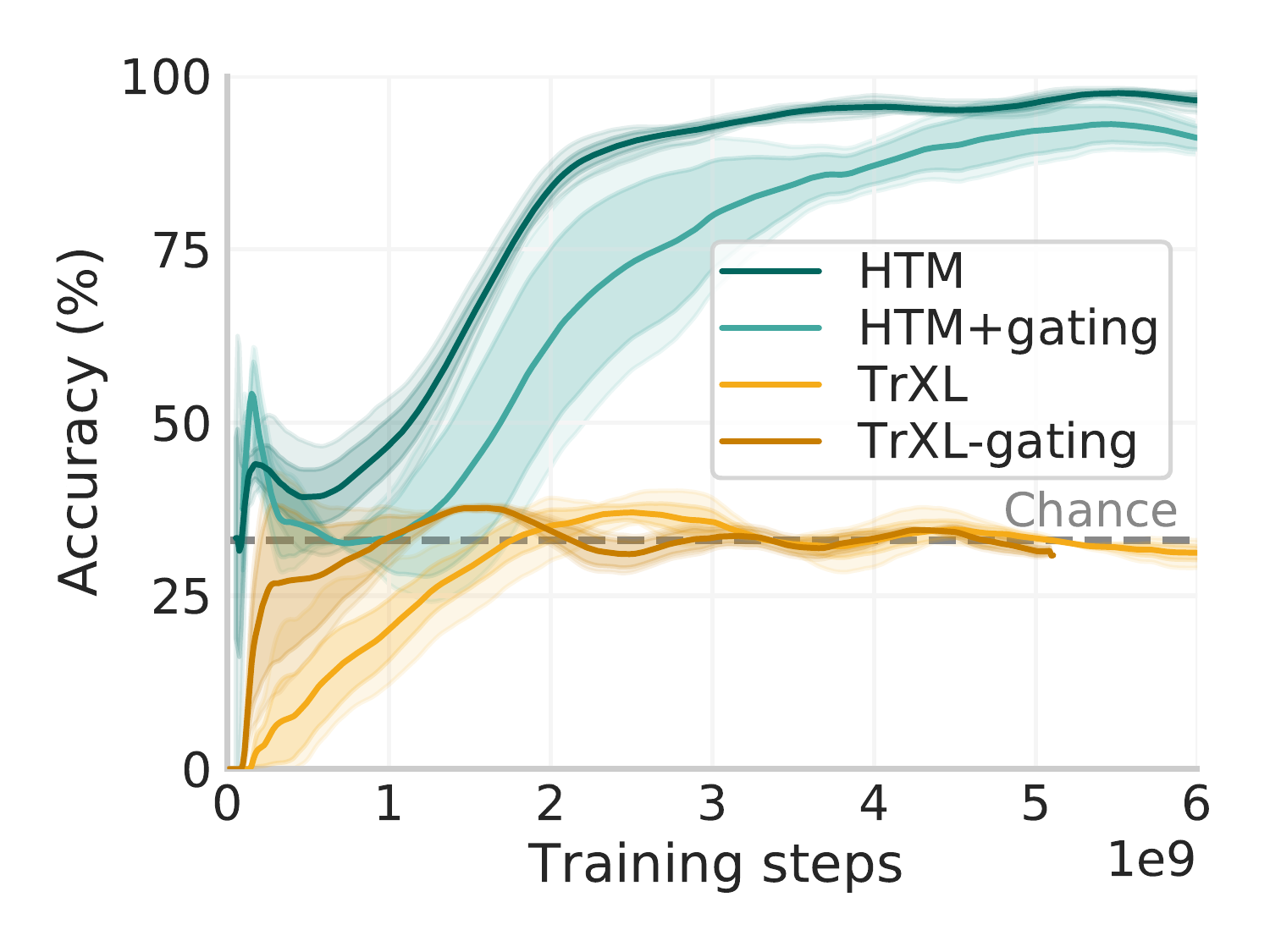}
    \captionsetup{width=.8\textwidth}
    \caption{Train 0-2 distractors,\\evaluate 20.}
    \label{fig:supp_exp:gating_lesion:fb:extrap}
    \end{subfigure}%
    \begin{subfigure}{0.33\textwidth}
    \centering
    \includegraphics[width=\textwidth]{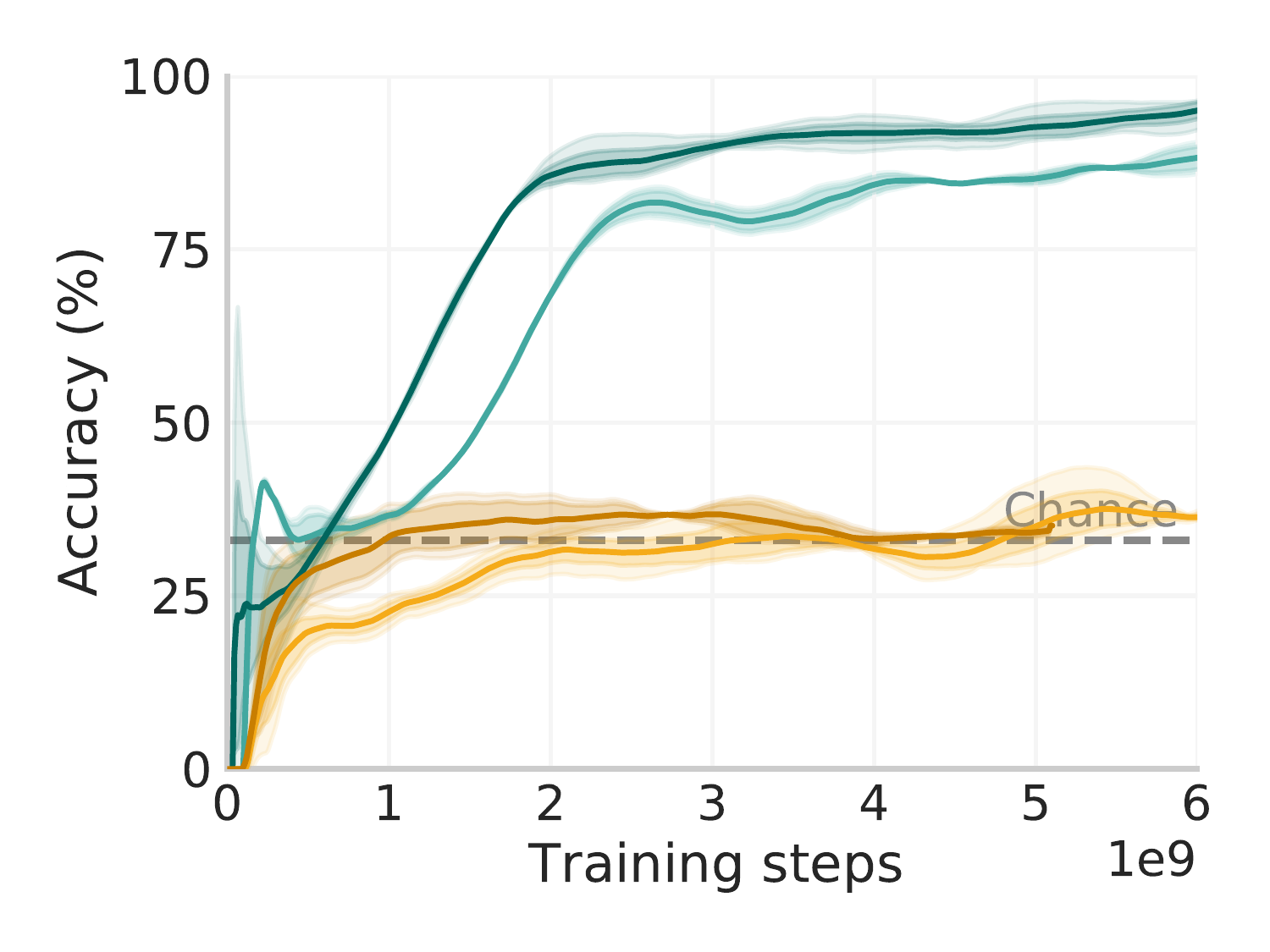}
    \captionsetup{width=.8\textwidth}
    \caption{Evaluate 4 episodes,\\1 distractor each.}
    \label{fig:supp_exp:gating_lesion:fb:across_4_1}
    \end{subfigure}%
    \begin{subfigure}{0.33\textwidth}
    \centering
    \includegraphics[width=\textwidth]{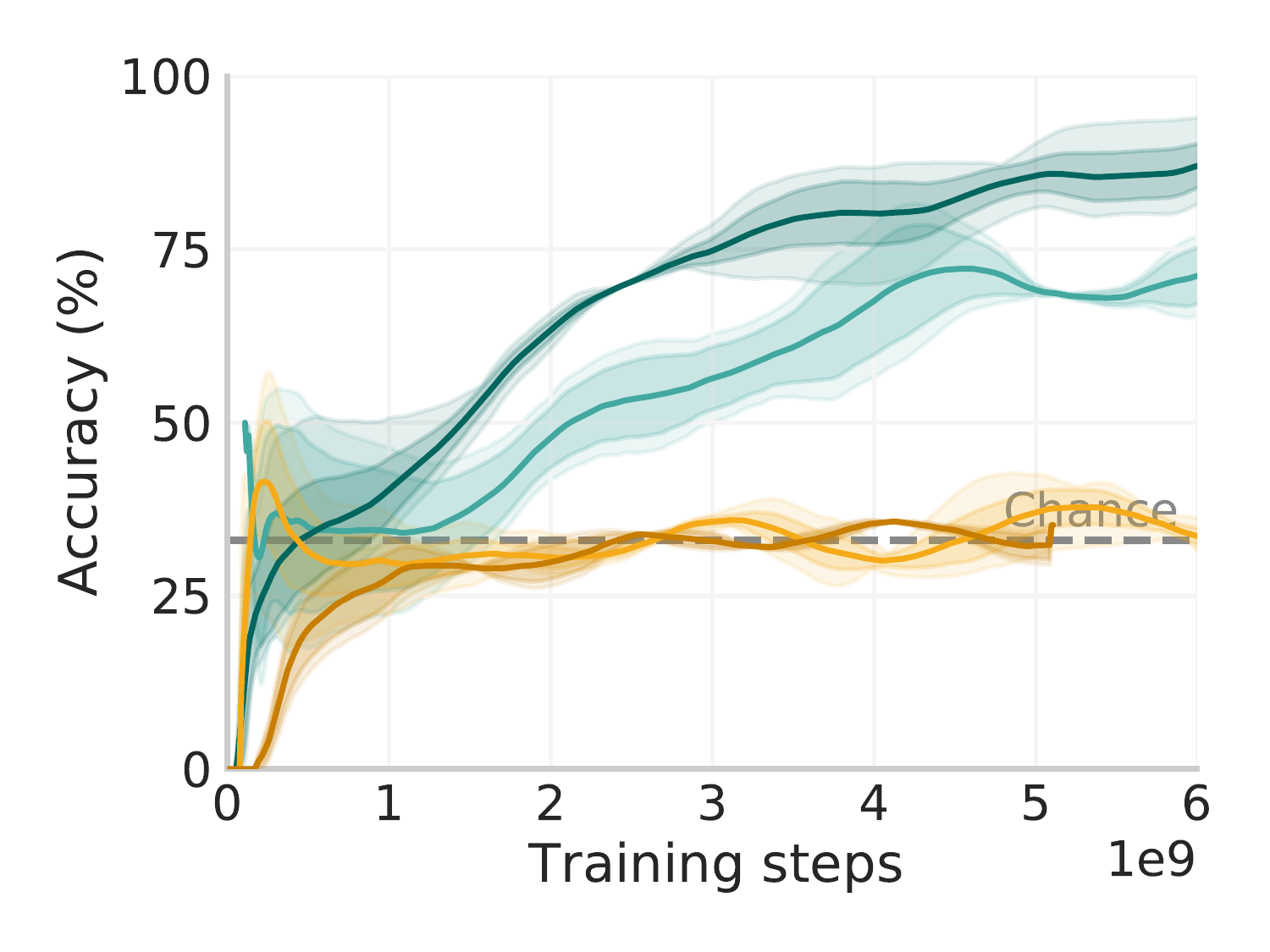}
    \captionsetup{width=.8\textwidth}
    \caption{Evaluate 3 episodes,\\5 distractors each.}
    \label{fig:supp_exp:gating_lesion:fb:across_3_5}
    \end{subfigure}%
    \caption{HCAM (labeled as HTM) performs better without gating \citep{parisotto2020stabilizing} than with gating. On the fast-binding tasks HCAM with gating learns slightly more slowly and generalizes slightly worse than without gating. Gating of memory layers does not appear necessary for TrXL in our tasks, unlike the experiments of \citet{parisotto2020stabilizing}. However, neither gated nor ungated TrXL are able to extrapolate to the tasks that gated or ungated HCAM does. (3 seeds per condition for main runs, 2 per condition for alternatives.)}
    \label{fig:supp_exp:gating_lesion}
\end{figure}

In Fig. \ref{fig:supp_exp:gating_lesion}, we show that HCAM's performance is not enhanced by the gating mechanism proposed by \citet{parisotto2020stabilizing}, and in fact HCAM actually learns slightly more slowly when its layers are gated. This is potentially because HCAM already has some notion of gating in its selection of relevant chunks, and additional gating therefore only interferes with the optimization process. Thus, we did not use gating for HCAM in our main experiments. Furthermore, we found that gating was not necessary to train the TrXL memory on our tasks, although we used it in our main experiments to match \citet{parisotto2020stabilizing}. However, HCAM with or without gating outperforms TrXL with or without gating.

\subsection{Comparing a TrXL that is 2\(\times\) wider/deeper than HCAM} \label{app:supp_exp:matched_params}

\begin{figure}[htb]
\centering
\begin{subfigure}{0.33\textwidth}
\centering
\includegraphics[width=\textwidth]{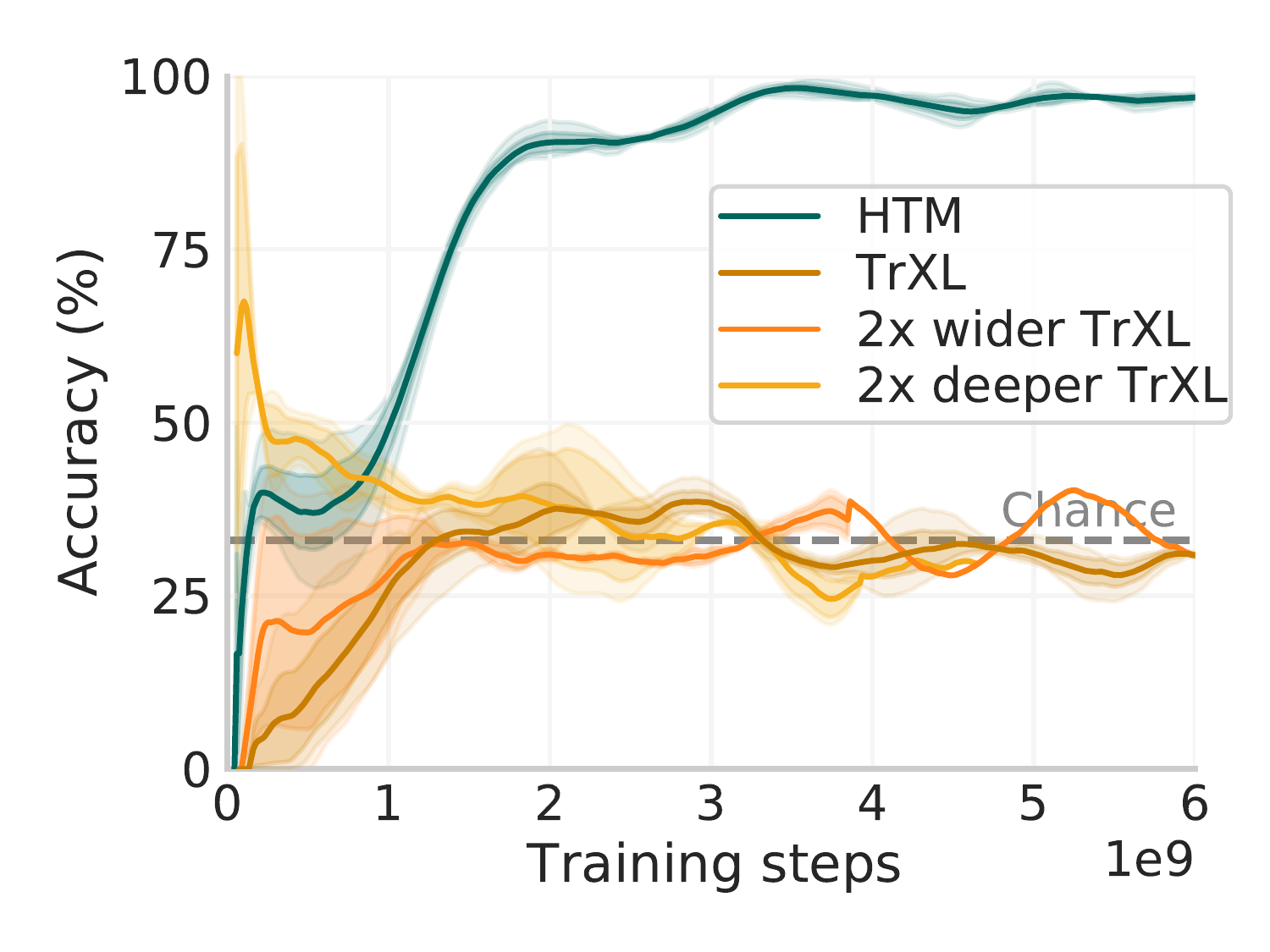}
\captionsetup{width=.8\textwidth}
\caption{Train 0-2 distractors,\\evaluate 20.}
\label{fig:supp_exp:params:fb:extrap}
\end{subfigure}%
\begin{subfigure}{0.33\textwidth}
\centering
\includegraphics[width=\textwidth]{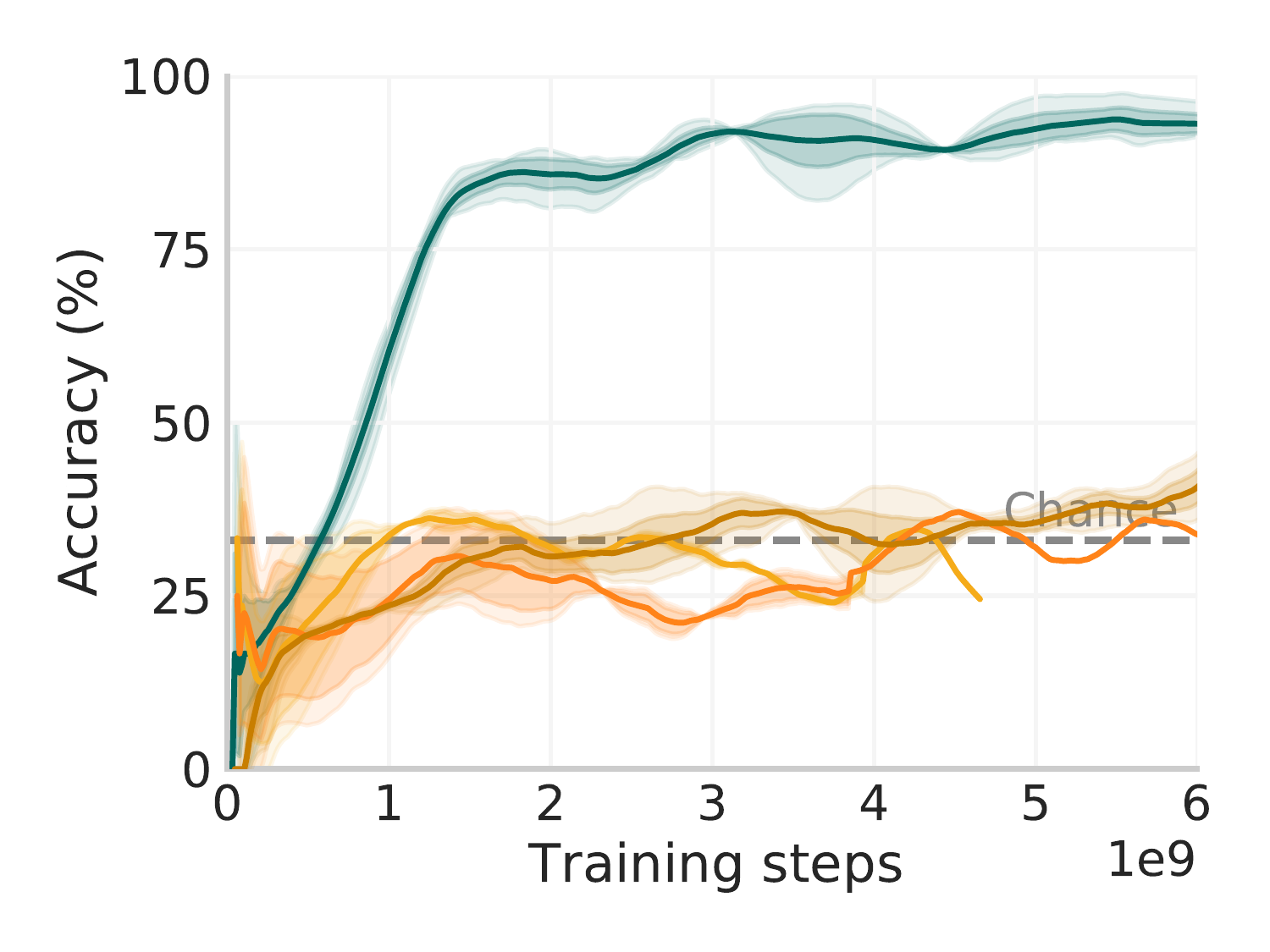}
\captionsetup{width=.8\textwidth}
\caption{Evaluate 4 episodes,\\1 distractor each.}
\label{fig:supp_exp:params:fb:across_4_1}
\end{subfigure}%
\begin{subfigure}{0.33\textwidth}
\centering
\includegraphics[width=\textwidth]{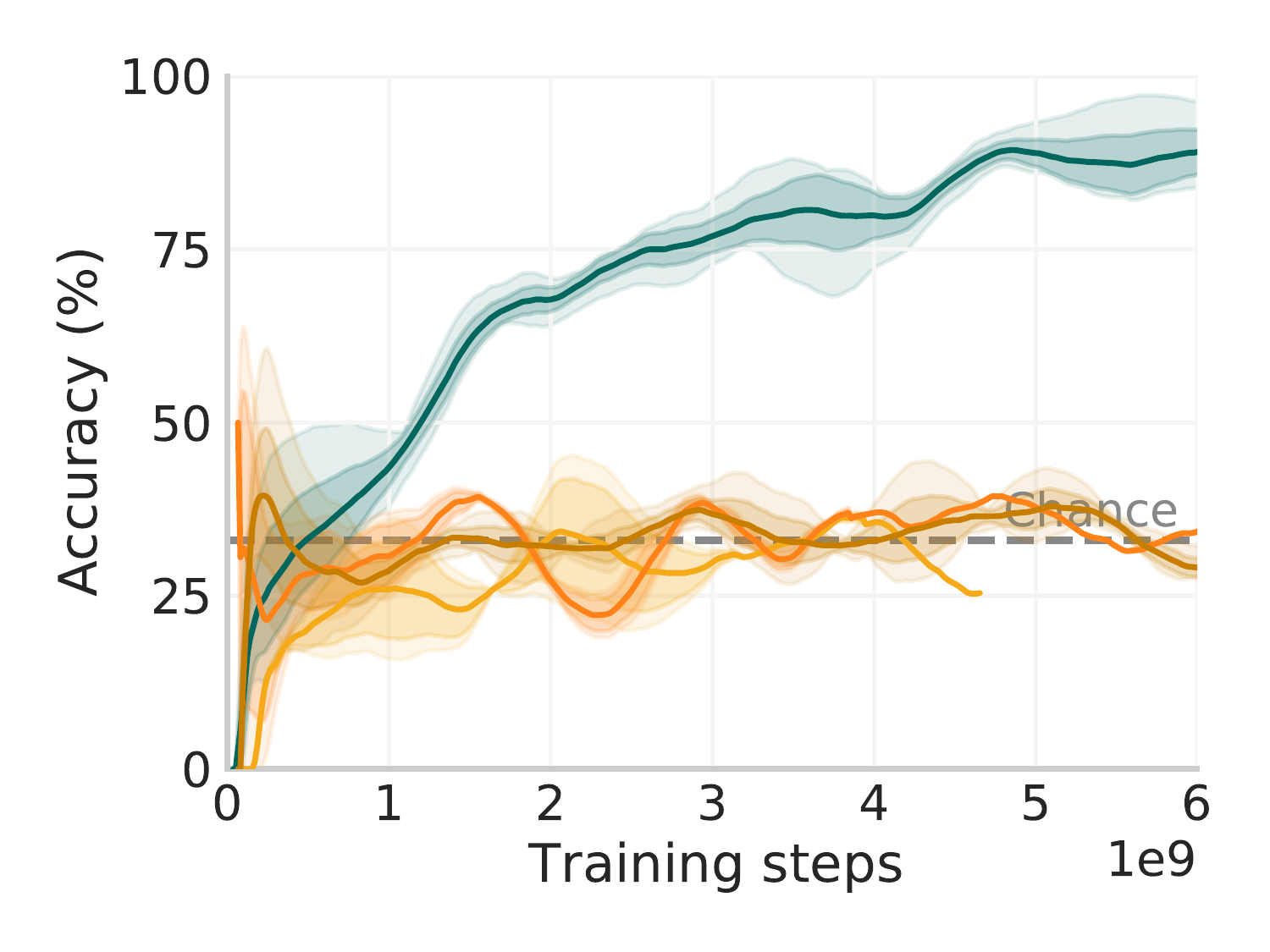}
\captionsetup{width=.8\textwidth}
\caption{Evaluate 3 episodes,\\5 distractors each.}
\label{fig:supp_exp:params:fb:across_3_5}
\end{subfigure}%

\caption{Comparing agents with TrXL memories that have more parameters than HCAM on the rapid word-learning tasks. Neither a TrXL model that is twice as wide, nor one that is twice as deep are able to perform as well as HCAM. Thus, HCAM's advantage is not due to the added blocks or slightly more parameters than our TrXL baseline. (HTM refers to HCAM, see note above. 3 seeds per main condition, 2 seeds per condition for supplemental.)}
\label{fig:supp_exp:matched_params}
\end{figure}

Because our HCAM-based agents have an added HCAM attention block in each memory layer compared to our TrXL-based ones, it might seem that they have somewhat more parameters and greater total depth. However, as noted in Section \ref{app:supp_exp:gating}, HCAM does not use the gating layers used by the TrXL memory \citep{parisotto2020stabilizing} and because of this HCAM uses about 20\% fewer parameters and is in some sense shallower than our TrXL baselines. However, it does have more layers of attention. To ensure that this or other simple factors were not the primary driver of HCAM's advantage, we ran comparisons where we either made the TrXL twice as wide (i.e. each layer had twice as many hidden units, including the attention projections etc.) or twice as deep (i.e. an 8-layer TrXL memory instead of 4-layers as we used for our main experiments). Both of these have substantially more parameters than our HCAM-based models, and the latter is substantially deeper as well. However, we show in Fig. \ref{fig:supp_exp:matched_params} that these much larger TrXL models were also unable to match the performance of HCAM on the rapid word-learning tasks. Thus the advantage of HCAM is not due to parameters or depth alone.

\subsection{Sparsity without hierarchy: a top-\(k\) TrXL} \label{app:supp_exp:topk_trxl}
One possible explanation of our results would be that sparsity alone is sufficient---perhaps the TrXL is suffering from spreading its attention across too many points in the past, but if it were restricted to only a few points hierarchy would not be necessary. To evaluate this possibility, we created a modified TrXL where we imposed sparsity of attention, by truncating its attention to only the top-\(k\) most relevant timepoints. We chose \(k=16\) to match HCAM. We show the results in Fig. \ref{app:supp_exp:topk_trxl}. The top-\(k\) TrXL performs comparably to a standard TrXL on the ballet tasks (i.e. does not perform as well as HCAM), and fails to learn properly on the rapid word-learning tasks, even if allowed to attend to a larger number of points (\(k=32\)). Thus, sparsity without hierarchy does not suffice, and may actually harm learning.
\begin{figure}[htb]
\centering
\begin{subfigure}{0.33\textwidth}
\centering
\includegraphics[width=\textwidth]{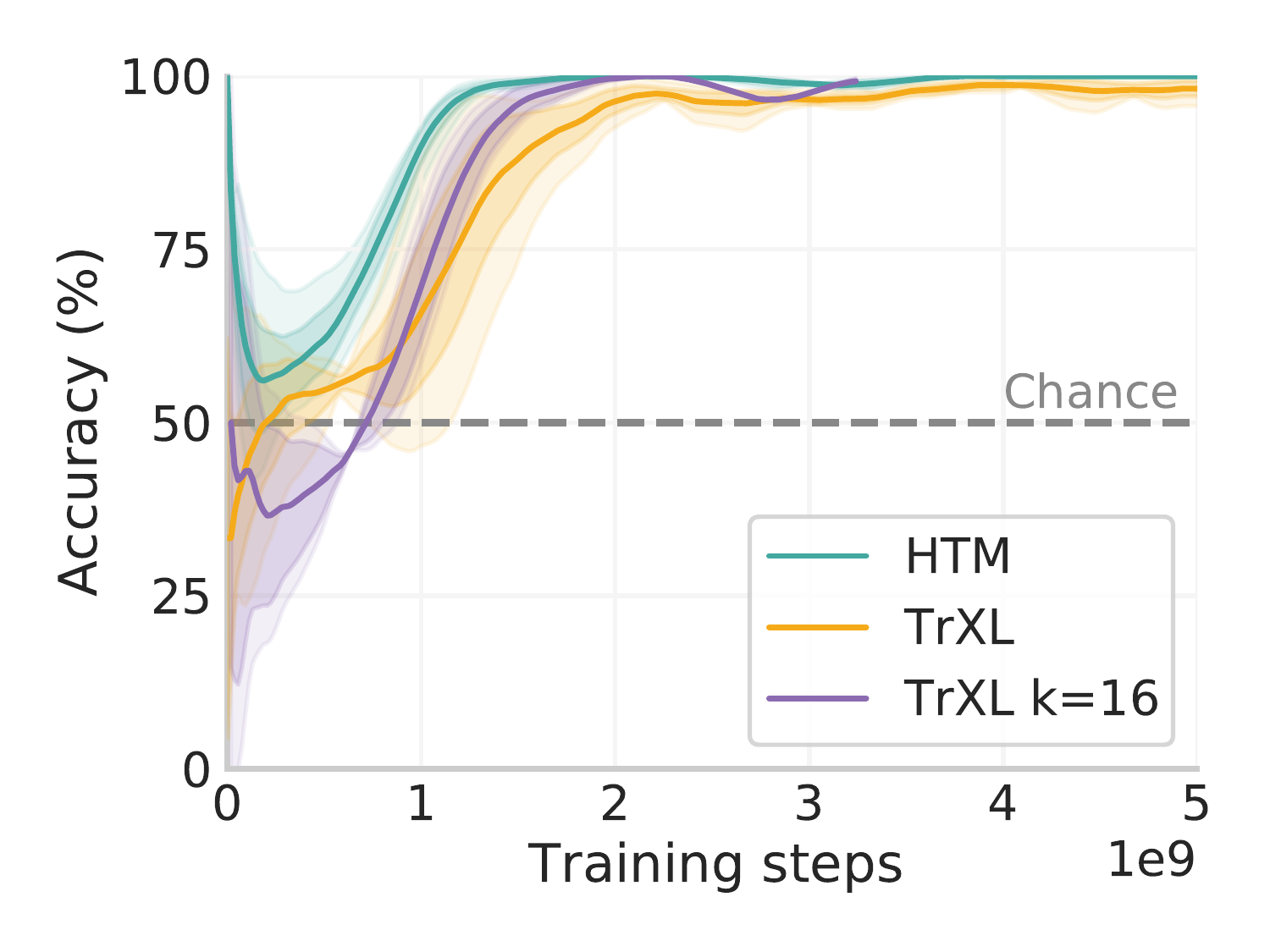}
\caption{Ballet, 2 dances, short delays.}
\label{fig:supp_exp:topktrxl:ballet:2}
\end{subfigure}%
\begin{subfigure}{0.33\textwidth}
\centering
\includegraphics[width=\textwidth]{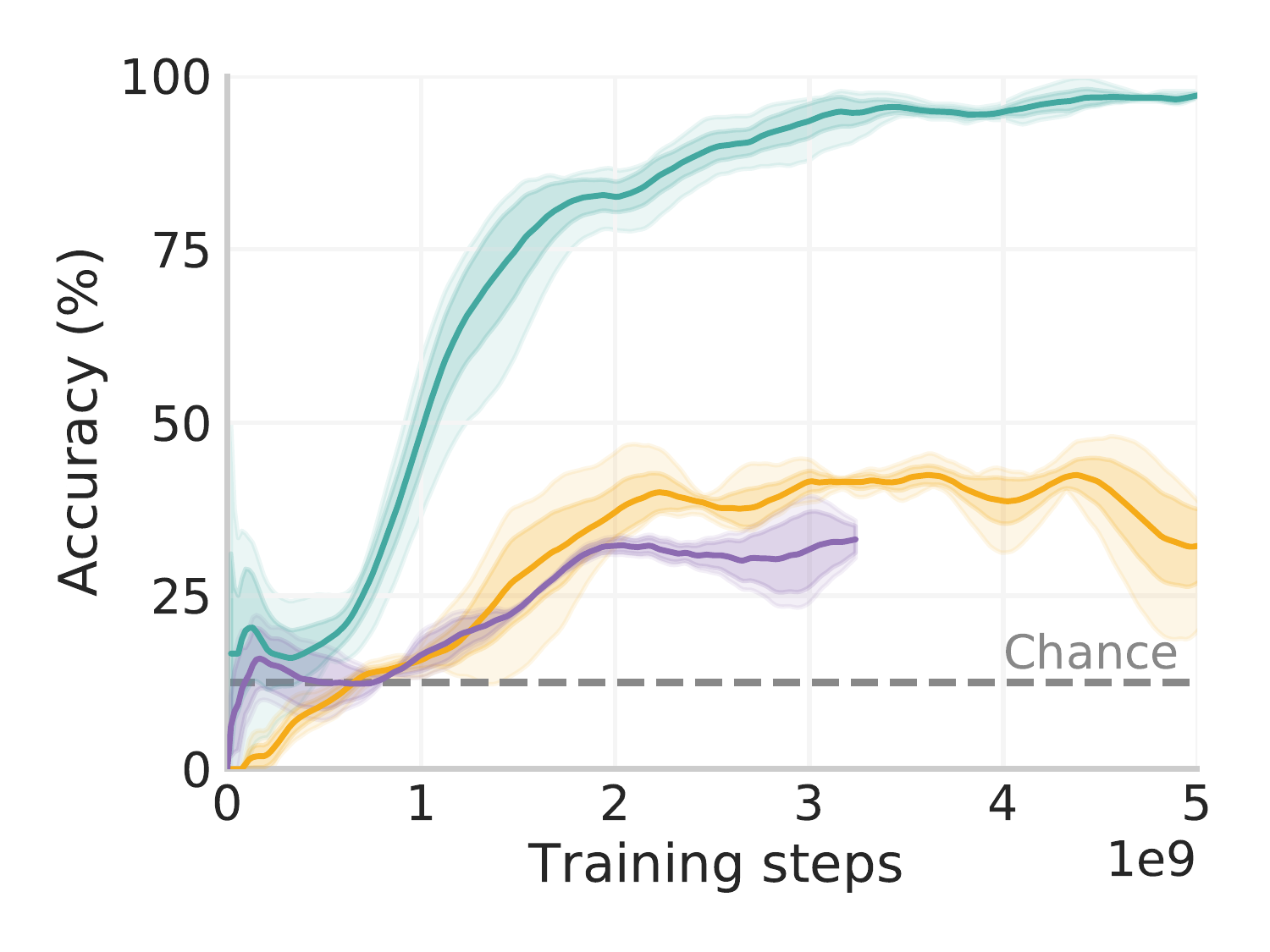}
\caption{Ballet, 8 dances, long delays.}
\label{fig:supp_exp:topktrxl:ballet:8}
\end{subfigure}\\
\begin{subfigure}{0.33\textwidth}
\centering
\includegraphics[width=\textwidth]{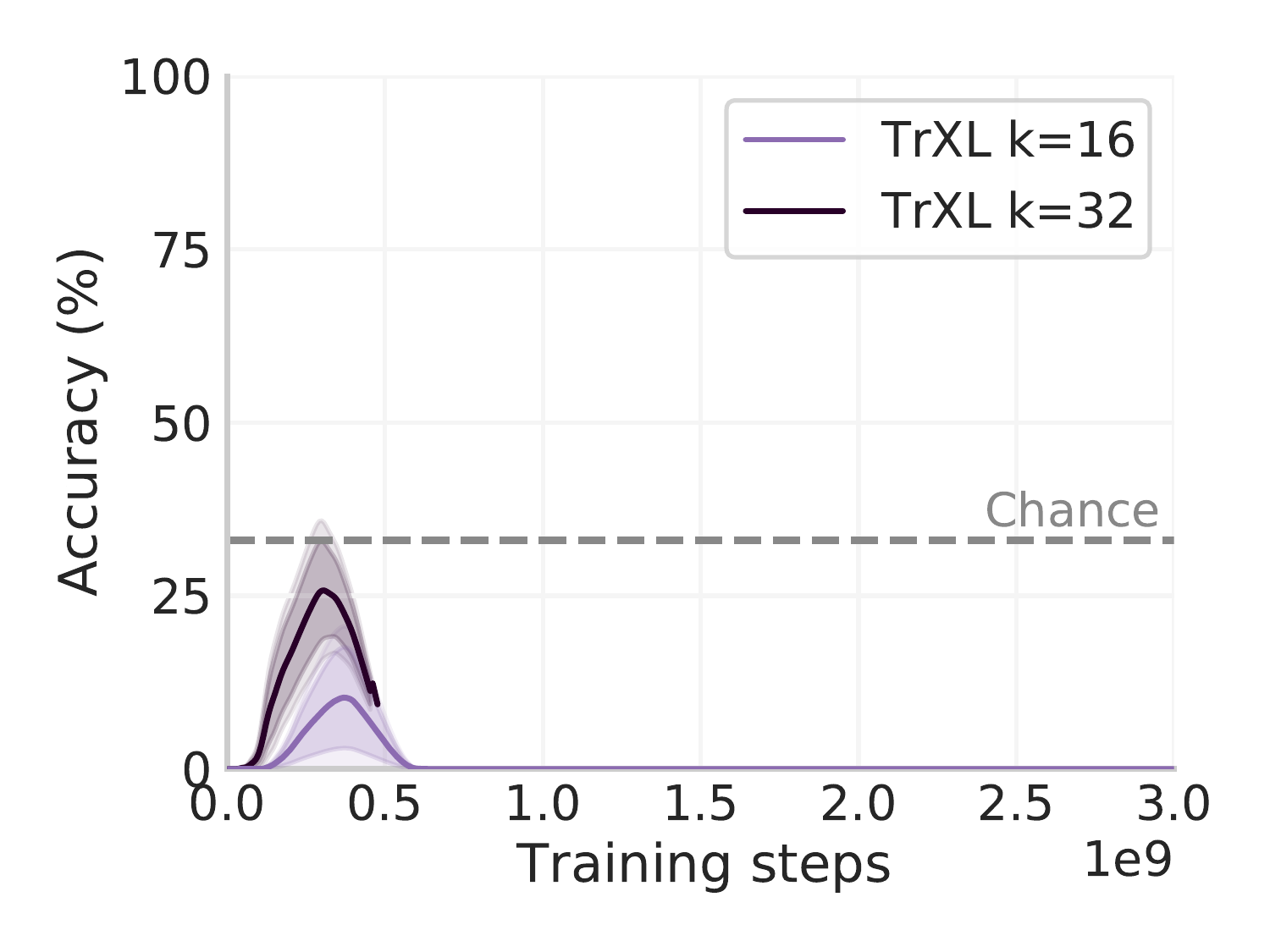}
\captionsetup{width=.8\textwidth}
\caption{Word learning, train:\\no distractors.}
\label{fig:supp_exp:topktrxl:fb:0}
\end{subfigure}%
\begin{subfigure}{0.33\textwidth}
\centering
\includegraphics[width=\textwidth]{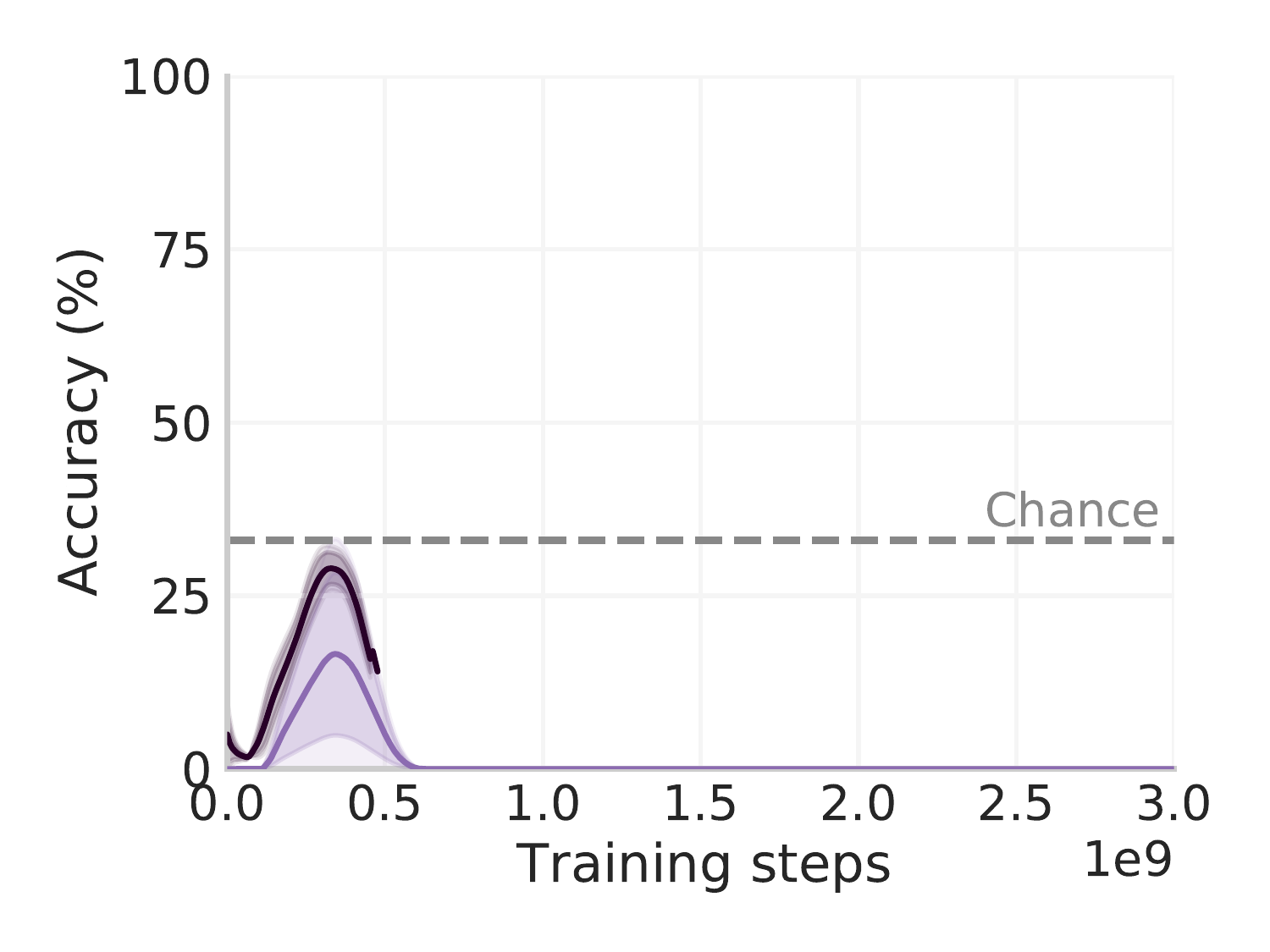}
\captionsetup{width=.8\textwidth}
\caption{Word learning, train:\\2 distractors.}
\label{fig:supp_exp:topktrxl:fb:2}
\end{subfigure}%

\caption{Sparsity alone is not sufficient---a TrXL restricted to attend to only the top-\(k\) timepoints performs comparably to a standard TrXL at the ballet tasks (\subref{fig:supp_exp:topktrxl:ballet:2}-\subref{fig:supp_exp:topktrxl:ballet:8}), but collapses and fails to learn in the rapid word learning tasks (\subref{fig:supp_exp:topktrxl:fb:0}-\subref{fig:supp_exp:topktrxl:fb:2}), even if given a larger \(k\). The advantage of HCAM is not due to sparsity alone. (HTM refers to HCAM, see note above. 3 seeds per main condition, 2 seeds per condition for supplemental.)}
\label{fig:supp_exp:topktrxl}
\end{figure}


\subsection{Compute efficiency assessed by learner FPS} \label{app:supp_exp:fps}
\begin{figure}[htb]
\centering
\includegraphics[width=0.33\textwidth]{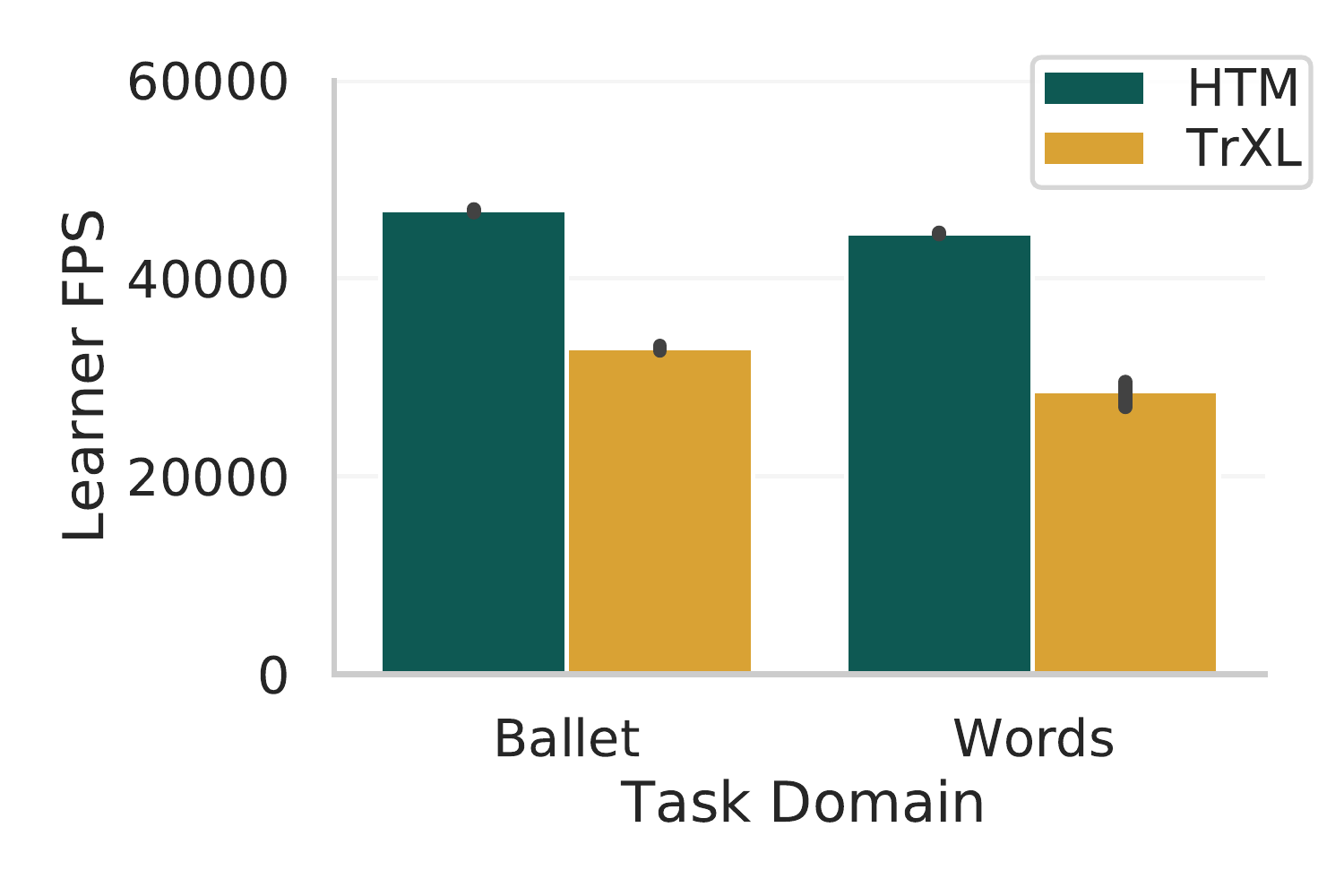}
\caption{HCAM (labeled as HTM) runs at a higher speed (measured in average learner frames processed per second) than TrXL. Results on TPUv3, error bars show 95\%-CI across 3 runs.}
\label{fig:supp_exp:fps}

\end{figure}

One goal of HCAM is that sparser attention might be more efficient than full attention. This is especially true when comparing HCAM without gating to the more computationally intense Gated TrXL \citep{parisotto2020stabilizing}. Correspondingly, we show in Fig. \ref{fig:supp_exp:fps} that HCAM generally runs \(\sim\)30-40\% faster than TrXL. 

\clearpage
\subsection{Results are robust to reviewer-suggested hyperparameter sweeps} \label{app:supp_exp:response}

Our reviewers evaluated the paper carefully, and expressed concerns that there might be bias in our hyperparameter selection. In particular, one reviewer raised a concern that TrXL might learn tasks like Ballet if given a different learning rate. To address these concerns, we ran a set of follow-up hyperparameter sweeps. We swept the learning rate and entropy weight (which we had not varied previously from the values used in prior work) on both the Ballet and Word tasks. We ran a full product of three learning rates above/below our original values (5e-4, 5e-5, 1e-5) and entropy weights \(5\!\times\) more or less than the original value, with 2 seeds per condition (for a total of 24 hyper \(\times\) seed \(\times\) memory type combinations per task domain). We emphasize that the same hyperparameters were tested for both models, and that these sweeps centered on hyperparameter settings that were previously untuned (sourced from prior papers), and that this sweep focuses on learning rate, which a reviewer suggested might particularly benefit TrXL. Our results show that HCAM is more robust to variation in these hyperparameters than TrXL, and generally sweeping these parameters does not improve TrXL's performance substantially beyond the results reported in the main text.

\textbf{Ballet results:} HCAM substantially outperforms TrXL in this sweep as well. First, HCAM is far more robust to varying the hyperparameters—in every hyperparameter setting, agents with HCAM achieved off chance performance (measured as window-averaged performance >5 percentage points above chance-level) on the hard 8-dance tasks within 1 billion steps. By contrast, 58\% of the TrXL jobs did not attain off chance performance on even the easiest task within 1.5 billion steps (when we stopped the training). In addition, 66\% of the HCAM jobs achieved above-75\% performance on the easiest tasks before \emph{any} of the TrXL jobs achieved above-chance performance on any task. The performance of the best TrXL jobs from this sweep is comparable to the performance at the same point in training from our original experiments: about 40-50\% performance on the 8 dance, short delays task, and 25-35\% on the 8 dance, long delays task at 1.5 billion steps. HCAM performed much better than TrXL, with 75\% of the HCAM jobs outperforming even the best TrXL agents on the hardest task, and the best HCAM jobs comparable to the results in the paper, achieving 80-90\% performance on the hardest tasks at 1.5 billion steps. In both 8 dance levels, the advantage of the two HCAM seeds with the best hyperparams over the two TrXL seeds with the best hyperparams is significant by a paired\footnote{paired reflecting the non-independence of the encoder initialization when the agents are initialized with the same random seed, but results are similar with an unpaired test, respectively \(t(2)=29\), \(p=0.001\); \(t(2)=9.5\), \(p=0.01\))} t-test, respectively \(t(1)=21\), \(p=0.03\) and \(t(1) = 101, p=0.006\). In summary, TrXL’s performance at these tasks does not seem to be improved by varying learning rates or entropy weight, and HCAM seems much more robust to variation in these hyperparameters.

\begin{figure}[htb]
\centering
\begin{subfigure}{0.33\textwidth}
\centering
\includegraphics[width=\textwidth]{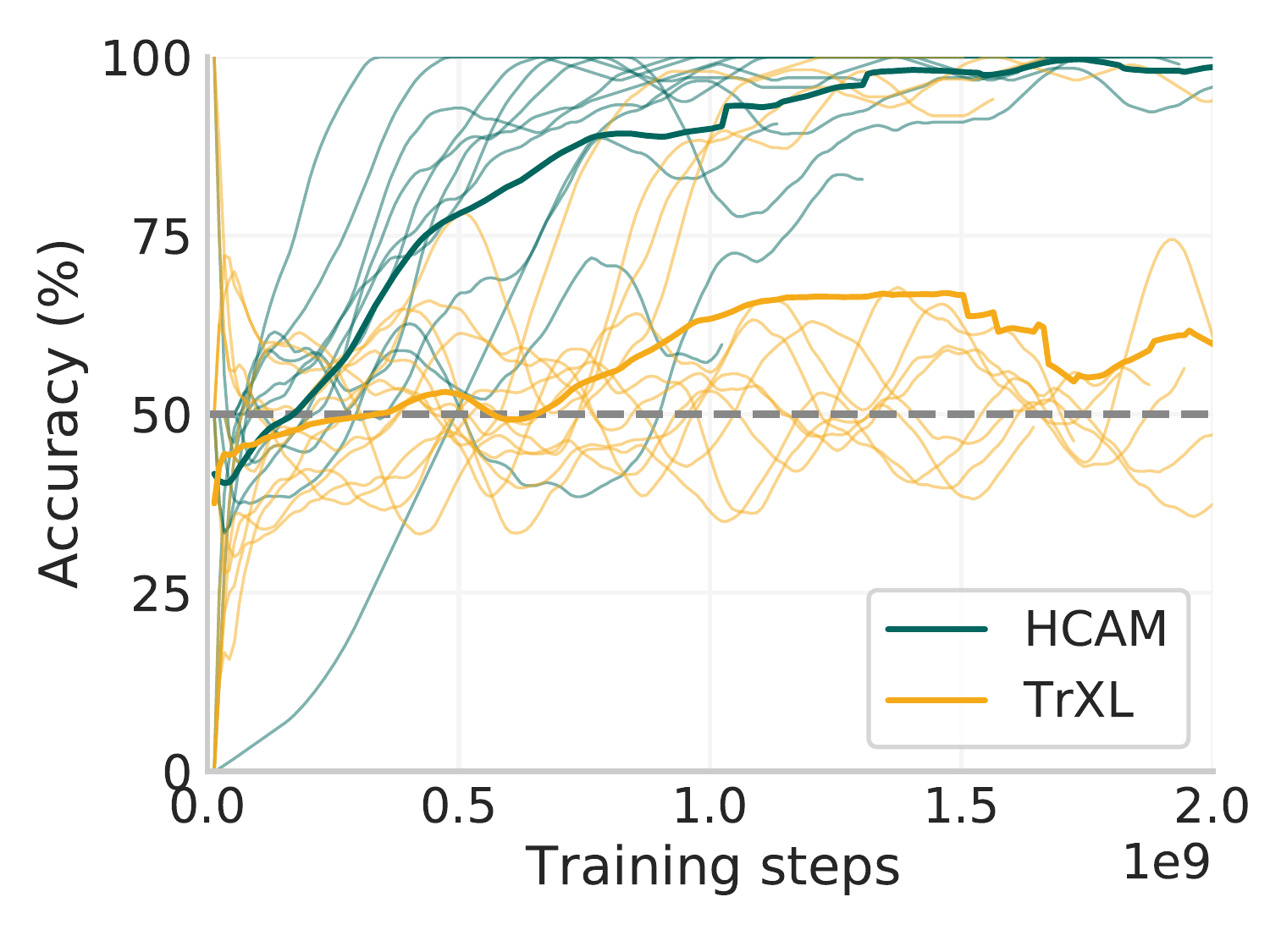}
\caption{Ballet, 2 dances, short delays.}
\label{fig:supp_exp:response:ballet:2}
\end{subfigure}%
\begin{subfigure}{0.33\textwidth}
\centering
\includegraphics[width=\textwidth]{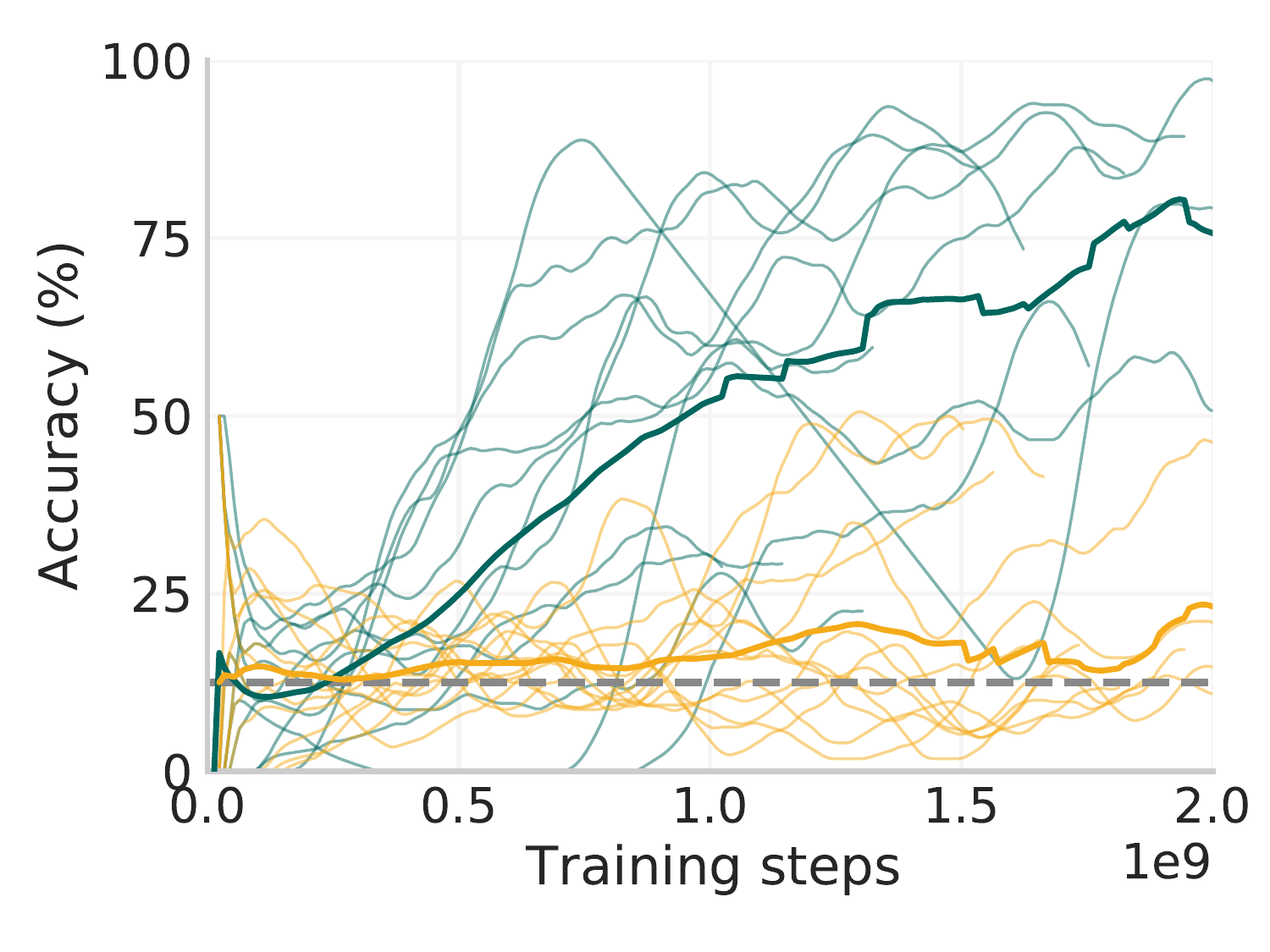}
\caption{Ballet, 8 dances, short delays.}
\label{fig:supp_exp:response:ballet:8short}
\end{subfigure}%
\begin{subfigure}{0.33\textwidth}
\centering
\includegraphics[width=\textwidth]{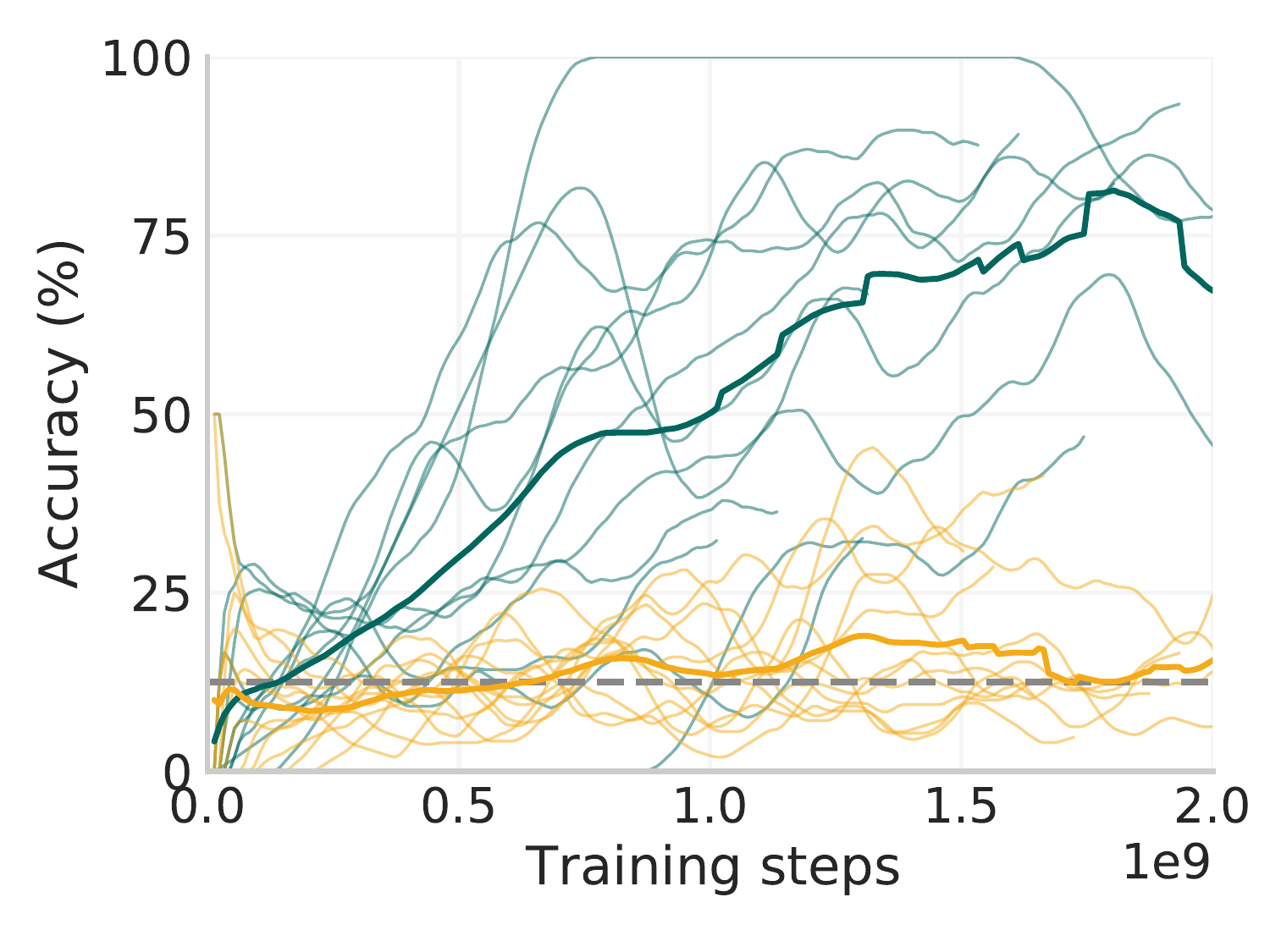}
\caption{Ballet, 8 dances, long delays.}
\label{fig:supp_exp:response:ballet:8long}
\end{subfigure}%
\caption{Sweeping hyperparameters (learning rate and entropy weight) in the Ballet tasks: HCAM is substantially more robust to variation in these hyperparameters, and the conclusions of our main text experiments are unaltered. The thick line plots the mean, while the thin lines plot individual sweep values. (The wide variability in early accuracy values should be disregarded---it is due to smoothing artifacts due to sparse data in this region as evaluation jobs are starting.)}
\label{fig:supp_exp:response:ballet}
\end{figure}

\textbf{Rapid word learning results:} The results are similar to the above. First, HCAM is more robust to varying hyperparameters: In these more challenging tasks, only 17\% of the TrXL jobs achieve high training performance within 5 billion steps, while 50\% of the HCAM jobs achieve high performance on the training tasks. HCAM also generalizes better than TrXL. However, unlike our original experiments, one set of these TrXL jobs does achieve somewhat above-chance performance at one of the the evaluation tasks we considered. We performed a replication with three new random seeds in the best hyperparameters from this sweep for each memory (as in the main results), and in this replication the TrXL did not achieve significantly off chance performance. However, HCAM did achieve significantly off-chance performance, (though not as high as the main text results using the hyperpareters tuned in our original sweeps). Thus, HCAM again appears to be both more robust across hyperparameters, and better when comparing best-hyperparameter configurations.

\begin{figure}[htb]
\begin{subfigure}{0.33\textwidth}
\centering
\includegraphics[width=\textwidth]{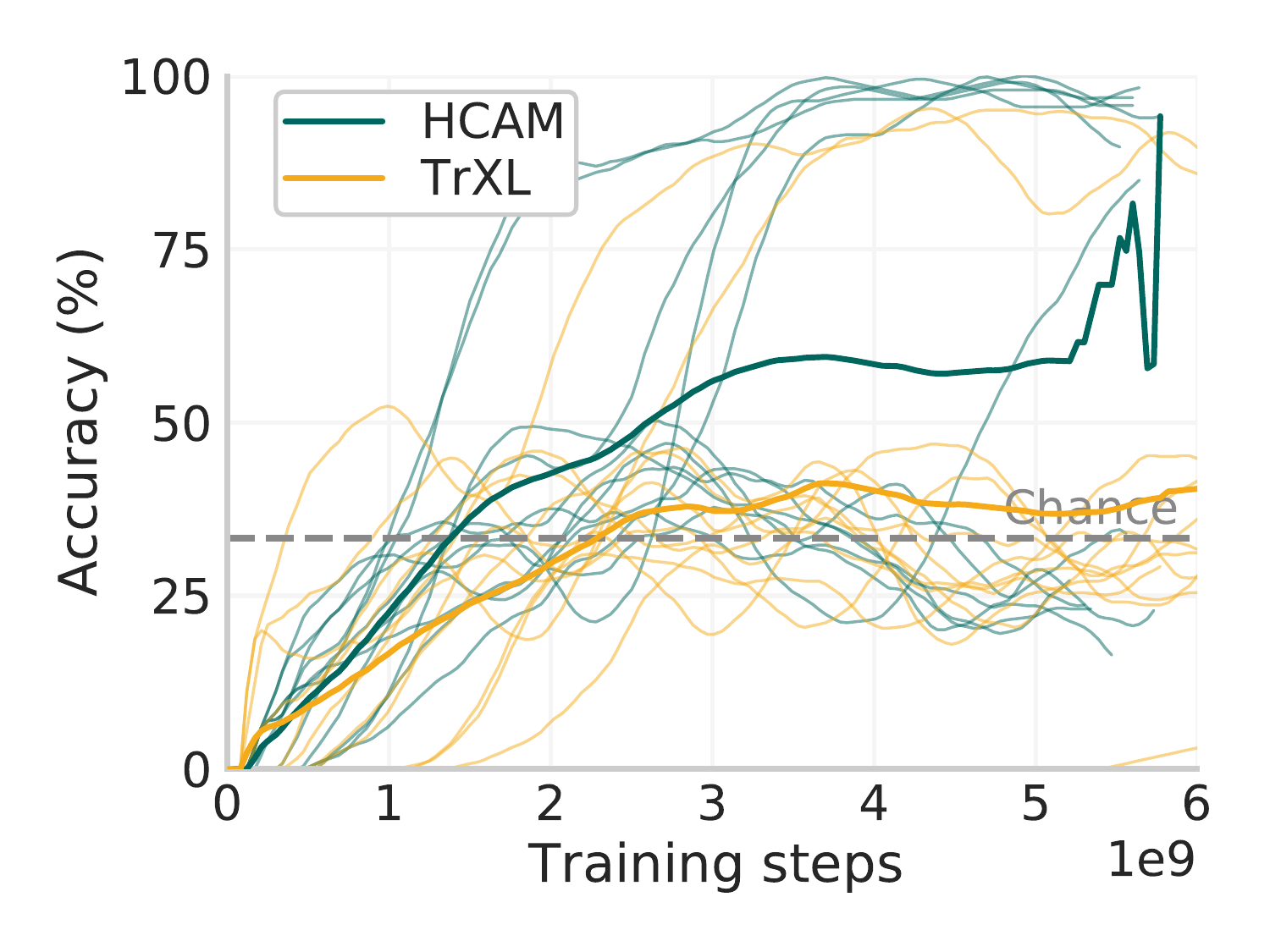}
\captionsetup{width=.8\textwidth}
\caption{Sweep: Train 0-2 distractors, evaluate 2 distractors.}
\label{fig:supp_exp:response:words:2}
\end{subfigure}\\
\begin{subfigure}{0.33\textwidth}
\centering
\includegraphics[width=\textwidth]{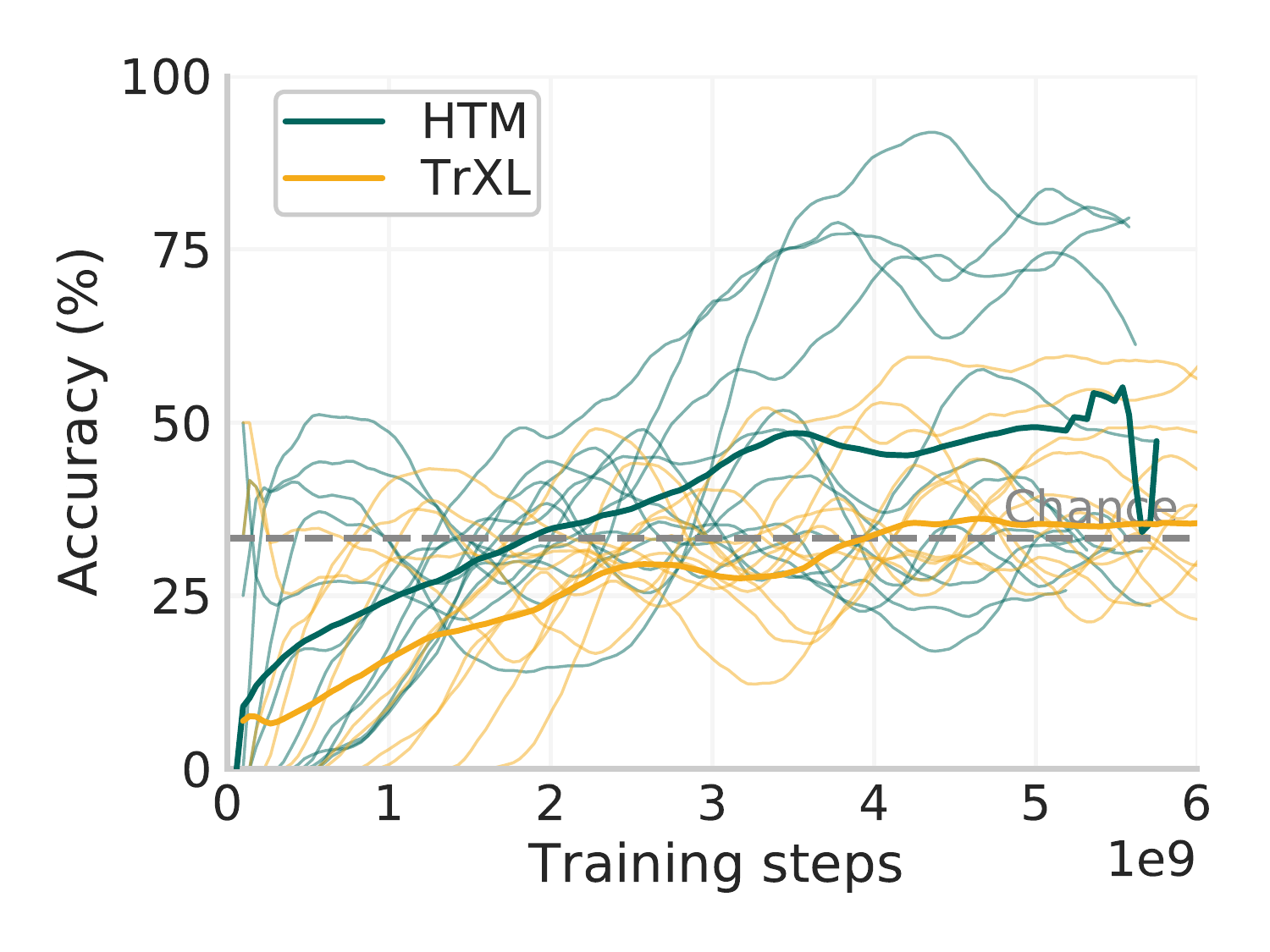}
\captionsetup{width=.8\textwidth}
\caption{Sweep: Train 0-2 distractors, evaluate 20 distractors.}
\label{fig:supp_exp:response:words:20}
\end{subfigure}%
\begin{subfigure}{0.33\textwidth}
\centering
\includegraphics[width=\textwidth]{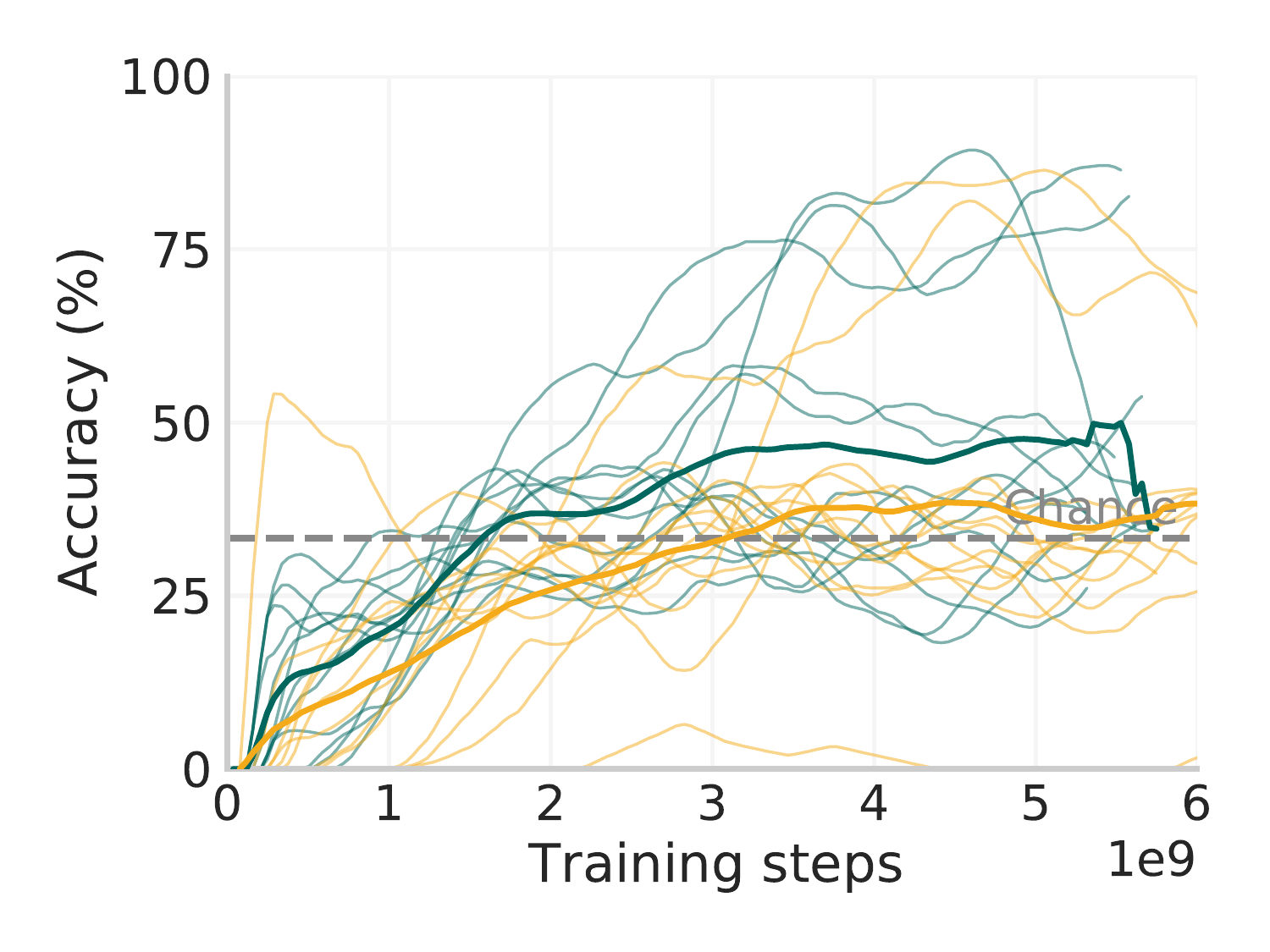}
\captionsetup{width=.8\textwidth}
\caption{Sweep: Train 0-2 distractors, evaluate 4 eps, 1 dist. each.}
\label{fig:supp_exp:response:words:4_1}
\end{subfigure}%
\begin{subfigure}{0.33\textwidth}
\centering
\includegraphics[width=\textwidth]{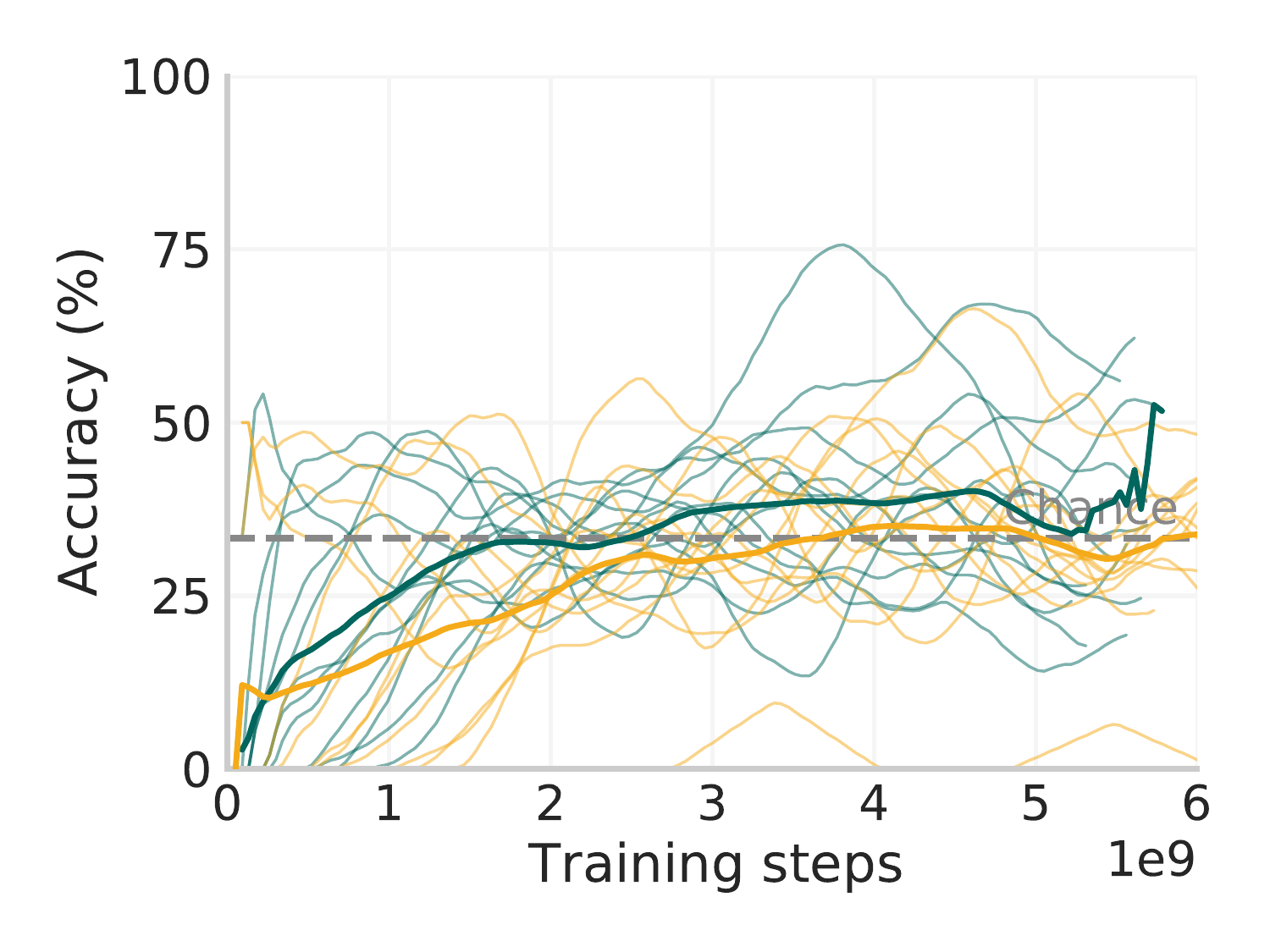}
\captionsetup{width=.8\textwidth}
\caption{Sweep: Train 0-2 distractors, evaluate 3 eps, 5 dist. each.}
\label{fig:supp_exp:response:words:3_5}
\end{subfigure}\\
\begin{subfigure}{0.33\textwidth}
\centering
\includegraphics[width=\textwidth]{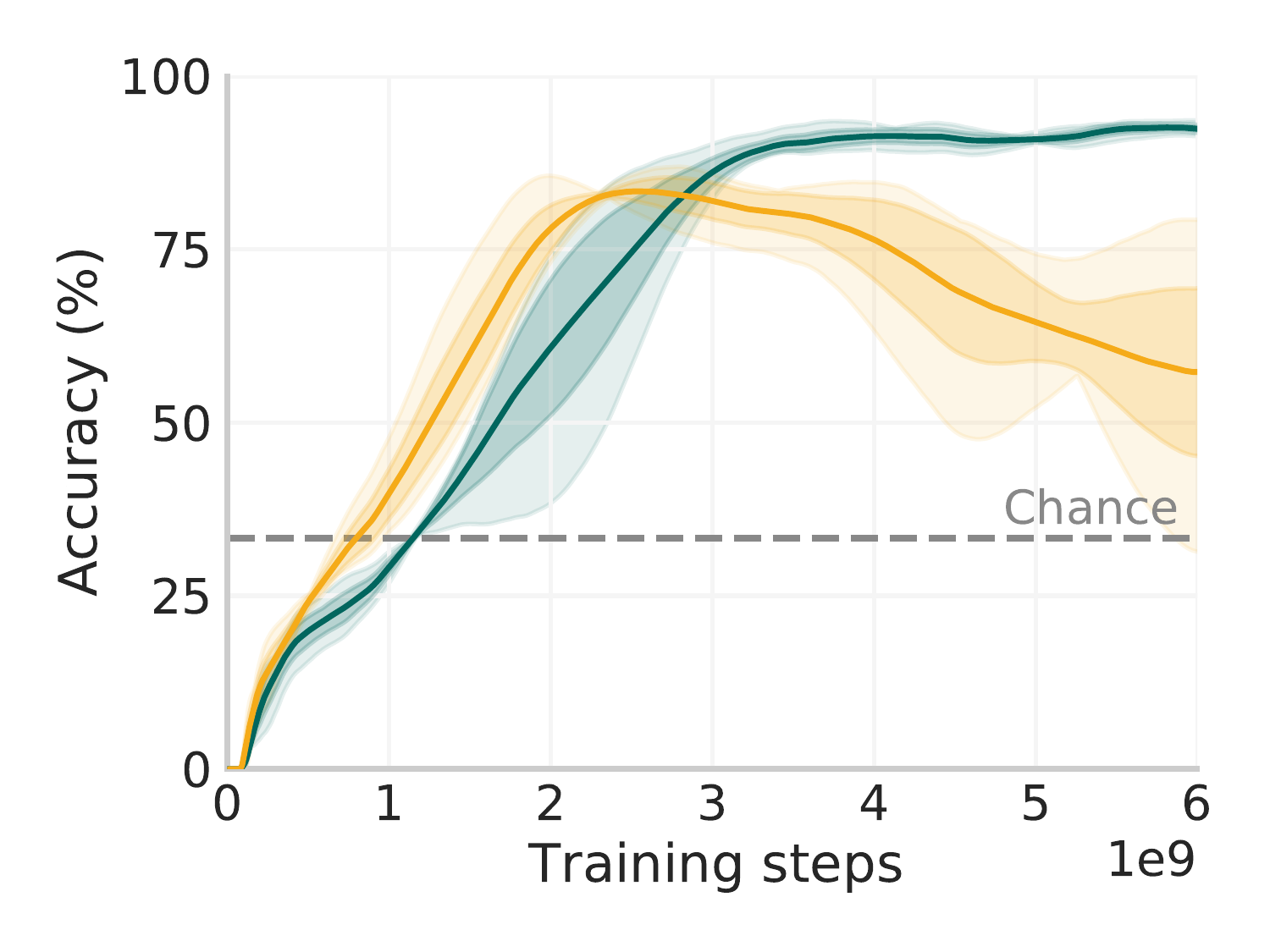}
\captionsetup{width=.8\textwidth}
\caption{Replication: Evaluate 2 distractors.}
\label{fig:supp_exp:response:words:rep2}
\end{subfigure}\\
\begin{subfigure}{0.33\textwidth}
\centering
\includegraphics[width=\textwidth]{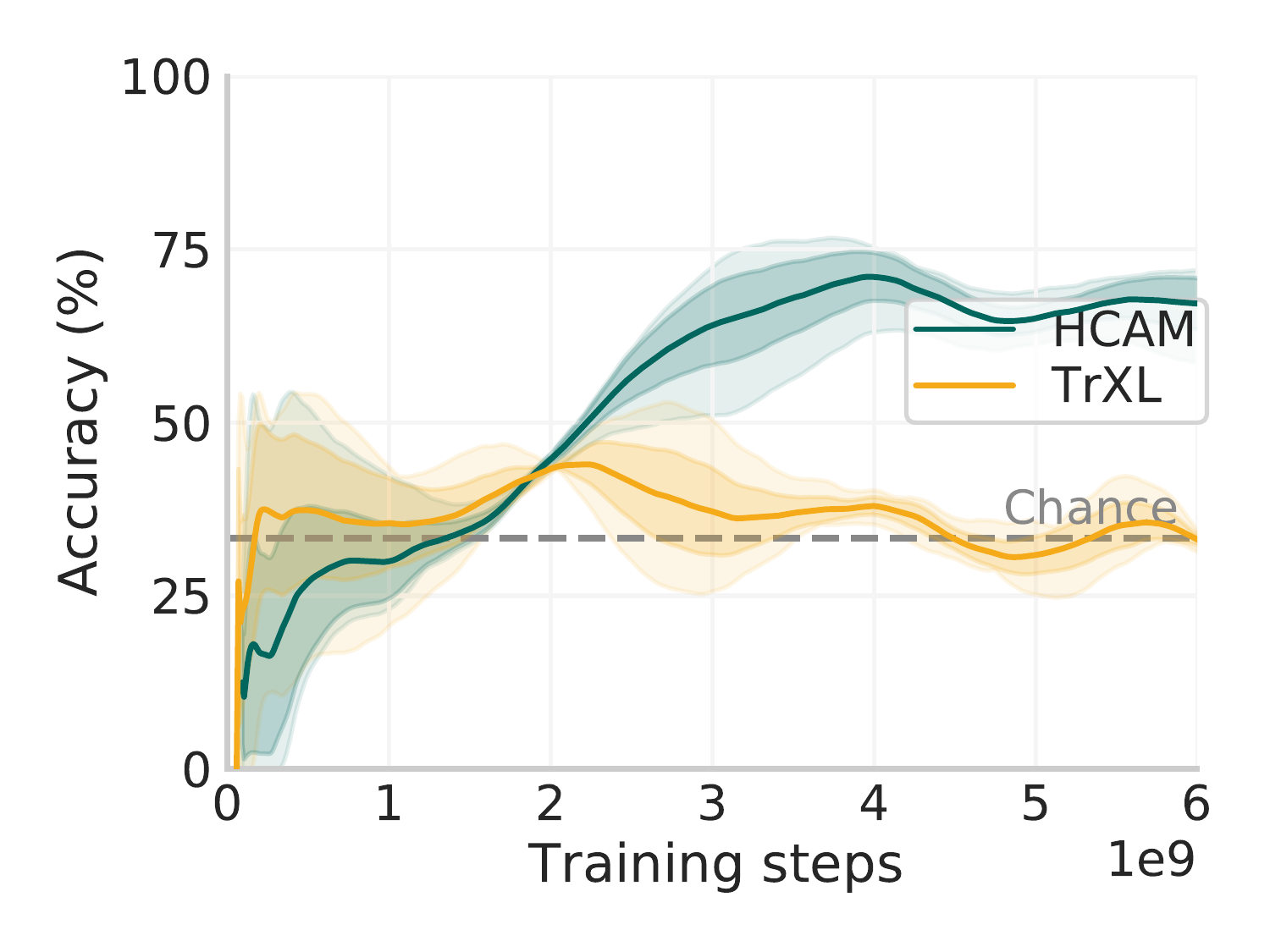}
\captionsetup{width=.8\textwidth}
\caption{Replication: Evaluate 20 distractors.}
\label{fig:supp_exp:response:words:rep20}
\end{subfigure}%
\begin{subfigure}{0.33\textwidth}
\centering
\includegraphics[width=\textwidth]{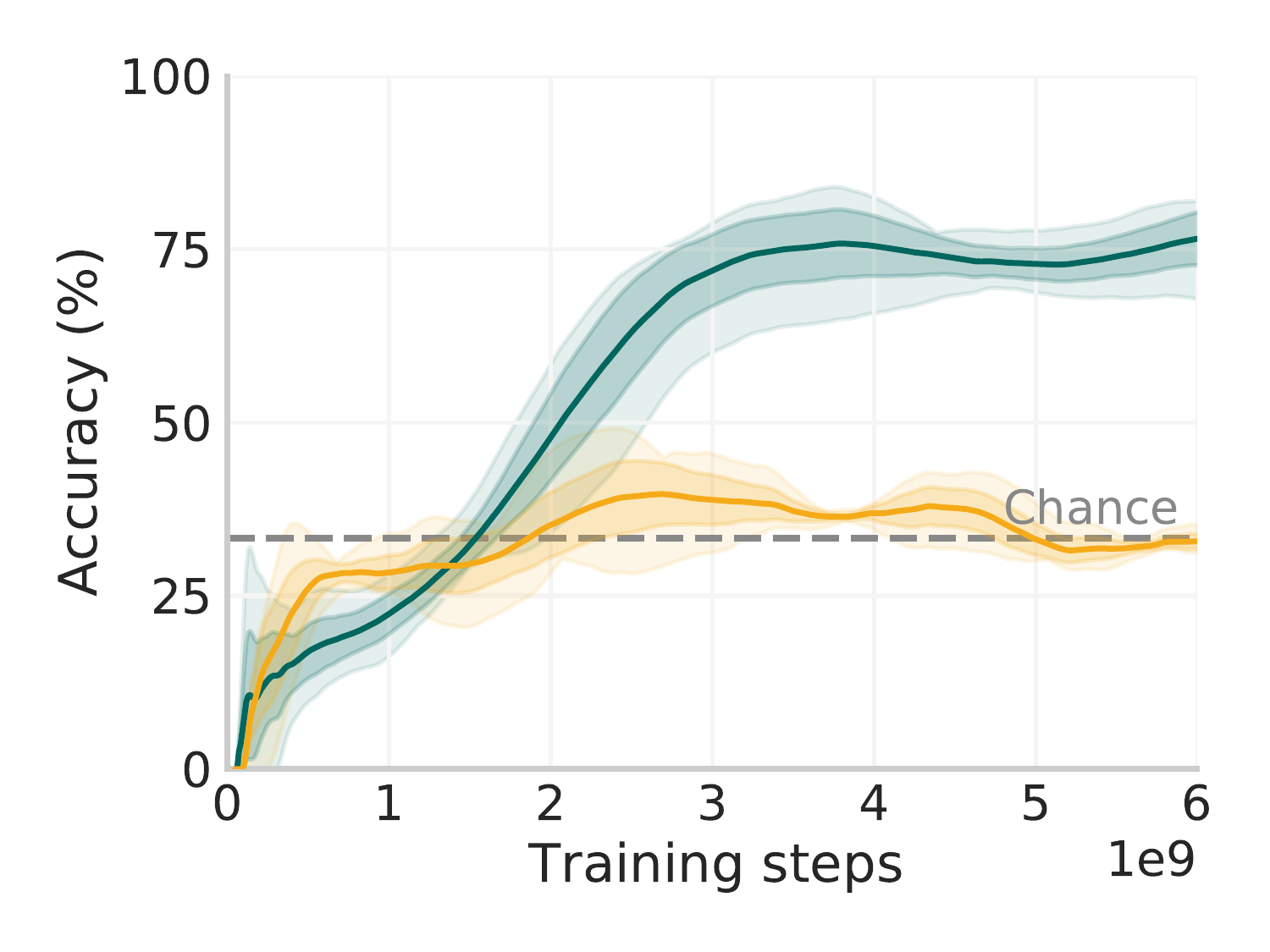}
\captionsetup{width=.8\textwidth}
\caption{Replication: Evaluate 4 eps, 1 dist. each.}
\label{fig:supp_exp:response:words:rep4_1}
\end{subfigure}%
\begin{subfigure}{0.33\textwidth}
\centering
\includegraphics[width=\textwidth]{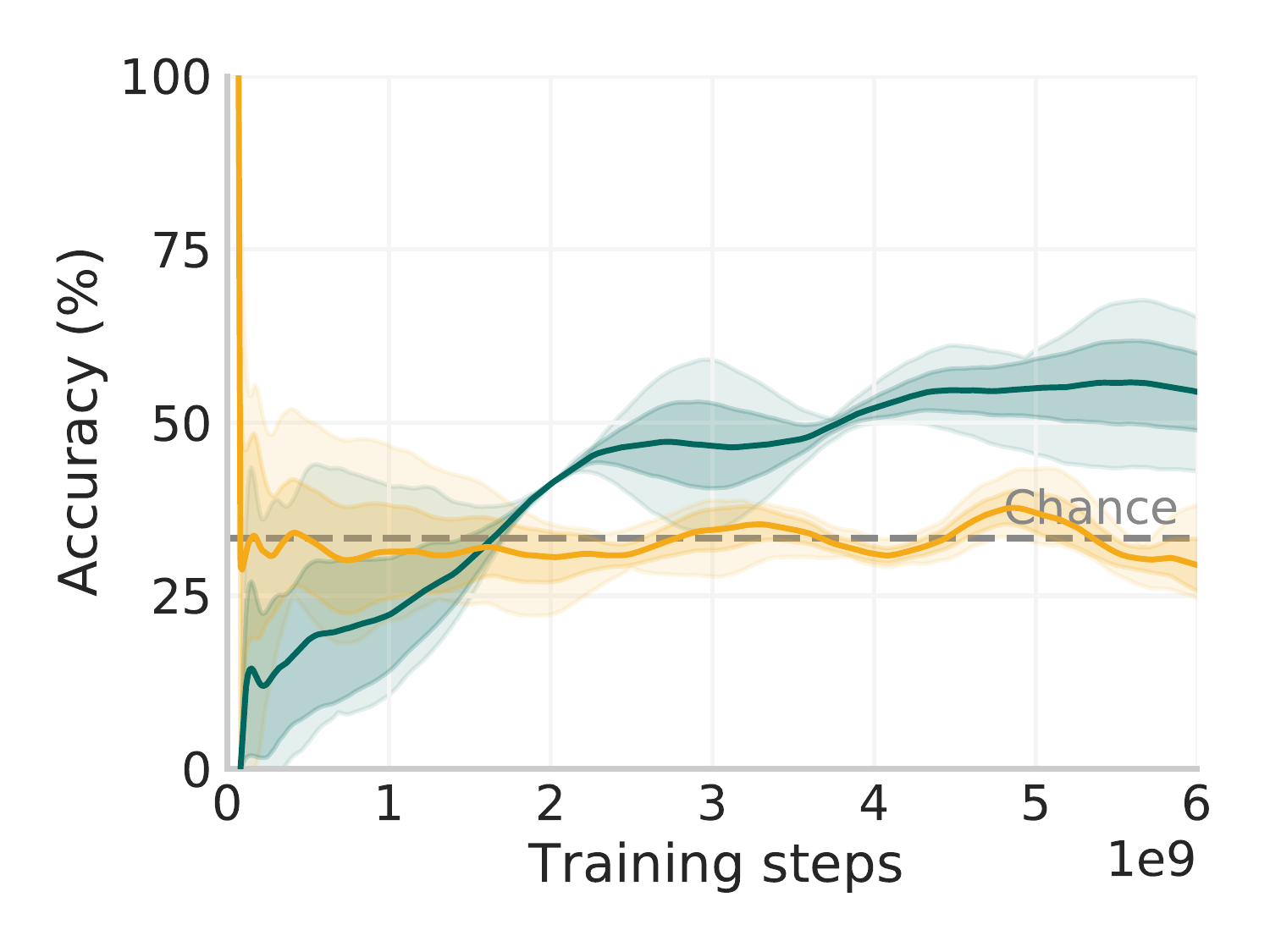}
\captionsetup{width=.8\textwidth}
\caption{Replication: Evaluate 3 eps, 5 dist. each.}
\label{fig:supp_exp:response:words:rep3_5}
\end{subfigure}%
\caption{Sweeping hyperparameters (learning rate and entropy weight) in the Words tasks: HCAM is again more robust to variation in these hyperparameters. (\subref{fig:supp_exp:response:words:2}-\subref{fig:supp_exp:response:words:3_5}) The sweep results. The thick line plots the mean, while the thin lines plot individual sweep values. While HCAM is much more robust overall, as shown in the number of hyperparameter settings that learn the train tasks (\subref{fig:supp_exp:response:words:2}), a few TrXL jobs show above chance generalization on some tasks in the sweep. (\subref{fig:supp_exp:response:words:rep2}-\subref{fig:supp_exp:response:words:rep3_5}) To follow-up on the above result, we ran a replication of the best hyperparameters from the sweep with three new random seeds (as we did for all our main text results). This replication does not show substantially above-chance generalization performance from TrXL and shows some collapse in performance on the train tasks. HCAM's performance remains substantially above chance in a replication of the best values in the sweep, although note that the best results from the sweep are worse than the results with the tuned hyperparameters used in the original experiments. (Note also that these experiments were run with a longer attention window for HCAM than intended, see App. \ref{app:methods:bug_fixed}.)}
\label{fig:supp_exp:response:words}
\end{figure}

\end{document}